\pdfoutput=1

\newcounter{myctr}

\documentclass{ws-ijhr}

\usepackage{hyperref}
\usepackage{tabulary}
\usepackage{booktabs}
\usepackage{subfigure}
\usepackage{xcolor}

\begin{document}

\markboth{C.-L. Fok, G. Johnson, J. D. Yamokoski, A. Mok, and L. Sentis}{ControlIt! - A Software Framework for Whole-Body Operational Space Control}

%
\catchline{}{}{}{}{}
%

\title{ControlIt! - A Software Framework for Whole-Body Operational Space Control }

\author{C.-L. Fok\textsuperscript{*}, G. Johnson\textsuperscript{*}, J. D. Yamokoski\textsuperscript{$\dagger$}, A. Mok\textsuperscript{*}, and L. Sentis\textsuperscript{*}}

\address{\textsuperscript{*}The University of Texas at Austin, \textsuperscript{$\dagger$}NASA Johnson Space Center}

\maketitle

\begin{history}
\received{Day Month Year} %
\revised{Day Month Year}  %
\accepted{Day Month Year} %
\end{history}

\begin{abstract}
Whole Body Operational Space Control (WBOSC) is a pioneering algorithm in the field of human-centered Whole-Body Control (WBC). It enables floating-base highly-redundant robots to achieve unified motion/force control of one or more operational space objectives while adhering to physical constraints. Although there are extensive studies on the algorithms and theory behind WBOSC, limited studies exist on the software architecture and APIs that enable WBOSC to perform and be integrated into a larger system. In this paper we address this by presenting ControlIt!, a new open-source software framework for WBOSC. Unlike previous implementations, ControlIt! is multi-threaded to increase servo frequencies on standard PC hardware. A new parameter binding mechanism enables tight integration between ControlIt! and external processes via an extensible set of transport protocols. 
To support a new robot, only two plugins and a URDF model needs to be provided --- the rest of ControlIt! remains unchanged. New WBC primitives can be added by writing a \texttt{Task} or \texttt{Constraint} plugin. ControlIt!'s capabilities are demonstrated on Dreamer, a 16-DOF torque controlled humanoid upper body robot containing both series elastic and co-actuated joints, and using it to perform a product disassembly task. Using this testbed, we show that ControlIt! can achieve average servo latencies of about 0.5ms when configured with two Cartesian position tasks, two orientation tasks, and a lower priority posture task. This is significantly higher than the 5ms that was achieved using UTA-WBC, the prototype implementation of WBOSC that is both application and platform-specific. Variations in the product's position is handled by updating the goal of the Cartesian position task. 
ControlIt!'s source code is released under an LGPL license and we hope it will be adopted and maintained by the WBC community for the long term as a platform for WBC development and integration.
\end{abstract}

\keywords{Software Framework; Whole Body Control; Whole Body Operational Space Control; Upperbody Humanoid Robot}

\section{Introduction} \label{sec:intro} Whole Body Control (WBC) takes a holistic view of a multi-branched highly
redundant robot like humanoids to achieve general coordinated behaviors. One
of the first WBC algorithms is Whole Body Operational Space Control
(WBOSC)~\cite{Sentis2005,Sentis2007-Thesis,Sentis2010}, which provides the
theoretical foundations for achieving operational space inverse dynamics, task
prioritization, free floating degrees of freedom, contact constraints, and
internal forces. There is now a growing community of researchers in this field
as exemplified by the recent formation of an IEEE technical committee on
WBC~\cite{Website-WBC_TC}. While the foundational theory and algorithms behind
WBC have recently made great strides, less progress exists in software
support, limiting the use of WBC today. In this paper, we remedy this problem
by presenting ControlIt!,\footnote{ControlIt! should not be associated with
MoveIt!~\cite {Website-MoveIt}. ControlIt! is primarily focused on whole body
feedback control whereas MoveIt! is primarily focused on motion planning.
Thus, MoveIt! and ControlIt! typically reside at different levels of a robot
application's software stack. The default feedback controller used by MoveIt!
is
\texttt{ros\_control}~\cite{Website_ROS_Control}. However, MoveIt! could be
configured to work with ControlIt! instead of \texttt{ros\_control} if
needed.} an open source\footnote{The source code for ControlIt! is available
under a LGPLv2.1 license. Instructions for downloading and using it are
available at
\url{https://robotcontrolit.com}} software framework for WBOSC.

\textbf{In this paper, we introduce ControlIt!, a software framework that enables
WBOSC controllers to be instantiated and is designed for systems integration,
extensibility, high performance, and use by both WBC researchers and the
general public.} Instantiating a WBOSC controller consists of defining a
prioritized compound task that defines the operational space objectives and
lower priority goal postures that the controller should achieve, and a
constraint set that specifies the natural physical constraints of the robot.
Systems integration is achieved through a parameter binding mechanism that
enables external processes to access WBOSC parameters through various
transport layers, and a set of introspection tools for gaining insight into
the controller's state at runtime. ControlIt! is extensible through the use of
plugins that enable the addition of new WBC primitives and support for new
robot platforms. High performance is achieved by using state-of-the-art
software libraries and multiple-threads that enable ControlIt! to offer
higher servo frequencies relative to previous WBOSC implementations. By making
ControlIt! open source and maintaining a centralized website
(\url{https://robotcontrolit.com}) with detailed documentation, installation
instructions, and tutorials, ControlIt! can be modified to evaluate new WBC
ideas and supported long term.

The intellectual merit and key contributions of this paper are as follows:

\begin{enumerate}
\item We design a software architecture for supporting general use of WBOSC
and its integration within a larger system via parameter binding and events.
\item We introduce the first API based on WBOSC principles for use across
general applications and robots.
\item We provide an open-source software implementation.
\item We design and implement a high performance multi-threaded architecture
that increases the achievable servo frequency by 10X relative to previous
implementations of WBOSC.
\item We reduce the number of components that need to be modified to develop a
new behavior to the set of {\texttt{RobotInterface}, \texttt{ServoClock},
\texttt{CompoundTask}, \texttt{ConstraintSet}} and decouple these changes from
core ControlIt! code via dynamically loadable plugins.
\item We demonstrate ControlIt!'s utility and performance using a humanoid
robot executing a product disassembly task.
\end{enumerate}

The remainder of this paper is organized as follows.
Section~\ref{sec:related_work} discusses related work. Section~\ref{sec:wbosc}
provides an overview of WBOSC's mathematical foundations.
Section~\ref{sec:architecture} presents ControlIt!'s software architecture
and APIs. Section~\ref{sec:evaluation} presents how ControlIt! was integrated
with Dreamer and used to develop a product disassembly task.
Section~\ref{sec:discussion} contains a discussion on other experiences using
ControlIt! and future research directions. The paper ends with conclusions in
Section~\ref{sec:conclusions}.

\section{Related Work}\label{sec:related_work} As a field, WBC is rapidly evolving. Most algorithms issue torque commands~\cite{Aghili2005,Hyon2007,Nakanishi2007,Mistry2010,Nagasaka2010,Mistry2011,Righetti2011,Righetti2012,Wakita2011,Lee2012,Salini2013_Thesis,Moro2013,Righetti2013,Saab2013,Lengagne2013,Koolen2013-ICRA,Henze2014,Righetti2014,Escande2014-IJRR}. 
They differ in whether they are centralized~\cite{Hyon2009,Sentis2013} or distributed~\cite{Mizuuchi2007,Nakanishi2013}, focus on manipulation~\cite{Dietrich2013}, locomotion~\cite{Ott2011,Englsberger2013,Moro2014}, or behavior sequencing~\cite{Whiteman2010,Hutter2013}, the underlying control models used~\cite{Hirai1998,Kajita2003,Bouyarmane2012}, and whether they've been evaluated in simulation or on hardware~\cite{Ohmichi1985,Hirose1991,Eiji1993,Matsumoto1995,Asfour2000,Katz2006,Albu-Schaeffer2007,Borst2007,Theobold2008,Freitas2009,Beetz2010,Iwata2009,Reiser2009,King2010,Stephens2010,Stilman2010,Meeussen2010,Hart2011_R2,Stephens2011,Moro2011,Tsagarakis2011,Bertrand2014,Herzog2014_IROS,Herzog2014_Arxiv,hutter2012,Hutter2014,Fuchs2009,Semini2011,Tellez2008}. These efforts demonstrate the behaviors enabled by WBC such as the use of compliance, multi-contact postures, robot dynamics, and joint redundancy to balance multiple competing objectives. ControlIt! is currently focused on supporting general use of WBOSC and its capabilities, but may be enhanced to include ideas and capabilities from these recent WBC developments.

An implementation of WBOSC called Stanford-WBC~\cite{Phlippsen2011} was released in 2011.  Stanford-WBC includes mechanisms for parameter reflection, data logging, and script-based configuration, but was a limited implementation of WBOSC that did not support branched robots, mobile robots, or contact constraints. It was used to make Dreamer's right arm wave and shake hands. More recently, UTA-WBC extended Stanford-WBC to support the full WBOSC algorithm, which includes branched robots, free floating degrees of freedom, contact constraints, and a more accurate robot model that includes rotor inertias~\cite{Website-UTA-WBC}. UTA-WBC was used to make a wheeled version of Dreamer containing 13 DOFs maintain balance on rough terrain. While this demonstrated the feasibility of WBOSC using a real humanoid robot, UTA-WBC was a research prototype targeted for a specific robot and specific behavior, i.e., balancing~\cite{Sentis2013}.  The implementation was not designed to work as part of a larger system for general applications. Instead, ControlIt! is a complete software re-design and re-implementation of the WBOSC algorithm with a focus on the software constructs and APIs that facilitate the integration of WBOSC into larger systems.

The differences between UTA-WBC and ControlIt! are shown in Table~\ref{table:UTA-WBC_vs_ControlIt}. Compared with UTA-WBC and Stanford-WBC, ControlIt! is a complete re-implementation that does not build upon but rather replaces the previous implementation. Specifically, ControlIt! contains new and more expressive software abstractions that enable arbitrarily complex WBOSC controllers to be configured, works with newer software libraries, middleware, and simulators, supports extensibility through a plugin-based architecture, is multi-threaded, and is designed to easily integrate with external processes through parameter binding and controller introspection mechanisms. 

\begin{table}[tb]
\centering
\small
\begin{tabulary}{\columnwidth}{l|L|L}
\midrule
\textbf{Property} & \textbf{UTA-WBC} & \textbf{ControlIt!} \\
\midrule
OS & Ubuntu 10.04 & Ubuntu 12.04 and 14.04\\
\midrule
ROS Integration & ROS Fuerte & ROS Hydro and Indigo \\
\midrule
Linear Algebra Library & Eigen 2 & Eigen 3 \\
\midrule
Model Library & Tao & RBDL 2.3.2 \\ 
\midrule
Model Description Format & Proprietary XML & URDF \\
\midrule
Integration (higher levels) & N/A & Parameter binding \\ 
\midrule
Integration (lower levels) & Proprietary & \texttt{RobotInterface} and \texttt{ServoClock} plugins \\
\midrule 
Controller Introspection & Parameter Reflection & Parameter Reflection and ROS Services \\
\midrule 
WBC Initial Configuration & YAML & YAML and ROS parameter server \\ 
\midrule
WBC Reconfiguration & N/A & Enable / disable tasks and constraints, update task priority levels \\ 
\midrule
Key Abstractions & task, constraint, skill & Compound task, constraint set \\
\midrule
Task / Constraint Libraries & Statically coded & Dynamically loadable via ROS pluginlib \\
\midrule
Number of threads & 1 & 3 \\
\midrule
Simulator & Proprietary & Gazebo 5.1 \\
\midrule
Website & \url{https://github.com/lsentis/uta-wbc-dreamer} & \url{https://robotcontrolit.com} \\
\midrule
\end{tabulary}
\caption{A comparison between UTA-WBC and ControlIt!} \label{table:UTA-WBC_vs_ControlIt}
\end{table}

The ability to integrate with external processes is important because applications of branched highly-redundant robots of the type targeted by WBC are typically very sophisticated involving many layers of software both above and below the whole body controller. To handle such complexity, a distributed component-based software architecture is typically used where the application consists of numerous independently-running software processes or threads that communicate over both synchronous and asynchronous channels~\cite{Heineman2001,Szyperski2002}. The importance of distributed component-based software for advanced robotics is illustrated by the number of recently developed middleware frameworks that provide it. They include OpenHRP~\cite{Kanehiro2002,Hirukawa2004}, RT-Middleware~\cite{Ando2005}, Orocos Toolchain~\cite{Website-Orocos-Toolchain}, YARP~\cite{Metta2006_YARP}, ROS~\cite{Website-ROS,Quigley2009_ROS}, CLARAty~\cite{Nesnas2006,Nesnas2007}, aRD~\cite{Hirzinger2006}, Microblx~\cite{Website-Microblx,Klotzbuecher2013}, OpenRDK~\cite{Website-OpenRDK,Calisi2008,Calisi2012}, and ERSP~\cite{Munich2005}. Among these, ControlIt! is currently integrated with ROS and is a ROS node within a ROS network, though usually as a real-time process potentially within another component-based framework (i.e., ControlIt!'s servo thread was an Orocos real-time task during the DRC Trials, and is a RTAI~\cite{Website-RTAI} real-time process in the Dreamer experiments discussed in this paper). In general, ControlIt! can be modified to be a component within any of the other aforementioned component-based robot middleware frameworks. 

ControlIt! is designed to interact with components both below (i.e., closer to the hardware) and above (i.e., closer to the end user or application) it within a robotic system. Components below ControlIt! include robot hardware drivers or resource allocators like \texttt{ros\_control}~\cite{Website_ROS_Control,Tsouroukdissian2014} and Conman~\cite{Website-Conman} that manage how a robot's joints are distributed among multiple controllers within the system. This is necessary since multiple WBC controllers may coexist and a manager is needed to ensure only one is active at a time. In addition, joints in a robots' extremity like those in an end effector usually have separate dedicated controllers. Components that may reside above ControlIt! include task specification frameworks like iTaSC~\cite{Website-iTaSC,Schutter2007,Decre2009,Decre2013}, planners like MoveIt!~\cite{Website-MoveIt}, management tools like Rock~\cite{Website-ROCK}, MARCO~\cite{Brunner1999}, and G$^{\rm en}$oM~\cite{Fleury1997}, behavior sequencing frameworks like Ecto~\cite{Website-Ecto} and RTC~\cite{Hart2014}, and other frameworks for achieving machine autonomy~\cite{Arkin1990,Alami1998,Jenkins2004,Pastor2009,Kim2010,Ott2013,Simmons1998,Kortenkamp1999}. Clearly, the set of components that ControlIt! interacts with is large, dynamic, and application-dependent. This is possible since component-based architectures provide sufficient decoupling to allow these external components to change without requiring ControlIt! to be modified.

\section{Overview of Whole Body Operational Space Control}\label{sec:wbosc} This section provides a brief overview of WBOSC. Details are provided in previous publications~\cite{Sentis2005,Sentis2007-Thesis,Sentis2010,Sentis2013}. Let $n_{joints}$ be the number of actual DOFs in the robot. The robot's joint state is represented by the vector $q_{actual}$ as shown by the following equation. 
\begin{equation}
q_{actual} = <q_1 \ldots q_{n_{joints}}>
\label{eq:q_actual}
\end{equation}

The robot's global pose is represented by a 6-dimensional floating virtual joint that connects the robot's base link to the world, i.e., three rotational and three prismatic virtual joints. It is denoted by vector $q_{base} \in \mathbb{R}^6$. The two partial state vectors, $q_{actual}$ and $q_{base}$, are concatenated into a single state vector $q_{full} = q_{actual} \cup q_{base}$. This combination of real and virtual joints into a single vector is called the \emph{generalized} joint state vector. Let $n_{dofs}$ be the number real and virtual DOFs in the model that is used by WBOSC. Thus, $q_{full} \in \mathbb{R}^{6 + n_{joints}} = \mathbb{R}^{n_{dofs}}$.


The underactuation matrix $U \in \mathbb{R}^{n_{joints} \times n_{dofs}}$ defines the relationship between the actuated joint vector and the full joint state vector as shown by the following equation.
\begin{equation}
q_{actual} = U \cdot q_{full}
\label{eq:underactuation}
\end{equation}

Let $A$ be the robot's generalized joint space inertia matrix, $B$ be the generalized joint space Coriolis and centrifugal force vector, $G$ be the generalized joint space gravity force vector, $J_c$ be the contact Jacobian matrix that maps from generalized joint velocity to the velocity of the constraint space dimensions, $\lambda_c$ be the co-state of the constraint space reaction forces, and $\tau_{command}$ be the desired force/torque joint command vector that is sent to the robot's joint-level controllers. The robot dynamics can be described by a single linear second order differential equation shown by the following equation. 
\begin{equation}
A \binom{\ddot{q}_{base}}{\ddot{q}_{actual}} + B + G + J_c^{\texttt{T}}\lambda_c = \binom{0_{6 \times 1}}{\tau_{command}} 
\label{eq:dynamics_equation}
\end{equation}

Constraints are formulated as follows. Let $\dot{p}_c$ be the velocity of the constrained dimensions, which we approximate as being completely rigid and therefore yielding zero velocity on the contact points, as shown by the following equation.
\begin{equation}
\dot{p}_c = J_c \binom{\dot{q}_{base}}{\dot{q}_{actual}} = 0
\label{eq:constraint}
\end{equation}

Tasks are formulated as follows. Let $\dot{p}_t$ be the desired velocity of the task, $J_t$ be the Jacobian matrix of task $t$ that maps from generalized joint velocity to the velocity of the task space dimensions, and $N_c$ be the generalized null-space of the constraint set. Furthermore, let $J_t^*$ be the contact consistent reduced Jacobian matrix~\cite{Sentis2007-Thesis} of task $t$, i.e., it is consistent with $U$ and $N_c$. The definition of $\dot{p}_t$ is given by the following equation where operator $\overline{arg}$ is the dynamically consistent generalized inverse of $arg$.
\begin{eqnarray}
\dot{p}_{t} & = & J_{t} \binom{\dot{q}_{base}}{\dot{q}_{actual}} = J_{t} \overline{UN_c} \dot{q}_{actual} \nonumber\\
            & = & J_{t}^{*} \dot{q}_{actual} \label{eq:task}
\end{eqnarray}

Let $\Lambda^*_t$ be the contact-consistent prioritized task-space inertia matrix~\cite{Sentis2007-Thesis} for task $t$, $\ddot{p}_{t,ref}$ be the reference, i.e., desired, task-space acceleration for task $t$, $\beta^*_t$ be the contact-consistent task-space Coriolis and centrifugal force vector for task $t$, and $\gamma^*_t$ be the contact-consistent task space gravity force vector for task $t$. The force/torque command of task $t$, denoted $F_t$, is given by the following equation.
\begin{equation}
F_t  = \Lambda^*_t \ddot{p}_{t,ref} + \beta^*_{t} + \gamma^*_{t}
\label{eq:taskCommand}
\end{equation}

To achieve multi-priority control, let $J_{t|prev}^{*}$ be the Jacobian matrix of task $t$ that is consistent with $U$, $N_c$, and all higher priority tasks. The equation for $\tau_{command}$ is the sum of all of the individual task commands multiplied by the corresponding $J_{t|prev}^{*}$ matrix as shown by the following equation.
\begin{equation}
\tau_{command}  = \sum_{t}J_{t|prev}^{*\texttt{T}} F_t 
\label{eq:command}
\end{equation}

Finally, when a robot has more than one point of contact with the environment, there are internal tensions within the robot. By definition, these ``internal forces'' are orthogonal to joint accelerations, i.e., they result in no net movement of the robot. The control structures like the multicontact/grasp matrix that are used to control these internal forces are documented in previous publications~\cite{Sentis2010}. Let $L^*$ be the nullspace of $(UN_c)$ and $\tau_{internal}$ be the reference (i.e., desired) internal forces vector. The contribution of the internal forces can thus be added to Equation~(\ref{eq:command}) as shown by the following equation.
\begin{equation}
\tau_{command}  = \sum_{t}\left(J_{t|prev}^{*\texttt{T}} F_t\right) + L^{*\texttt{T}} \tau_{internal}
\label{eq:command_with_internal_forces}
\end{equation}

\section{ControlIt! Software Architecture}\label{sec:architecture} There are six guiding principles behind ControlIt!'s development: (i)
\textbf{separate concerns} into interface definitions, implementations, and
configuration, (ii) support \textbf{extensibility and platform-independence}
through dynamically loadable plugins, (iii) encourage \textbf{code reuse}
through plugin libraries, (iv) support \textbf{systems integration} through
parameter binding, events, data introspection services, and compatibility with
a modern software ecosystem, (v) be cognizant of \textbf{performance and real-
time considerations}, and (vi) support \textbf{two types of end users}:
developers who use ControlIt! and researchers who modify ControlIt!.

Section~\ref{sec:software} contains a discussion of ControlIt!'s software
architecture, which describes the software components within ControlIt's core.
Many of these components either instantiate plugins or are implemented by
plugins. The use of plugins enables ControlIt! to be extensible in terms of
supporting different robots and applications. Section~\ref{sec:integration}
discusses mechanisms for configuring and integrating ControlIt! into a larger
system. This includes the parameter reflection, binding, and event signaling
mechanisms, and YAML specification files. Finally, a description of
ControlIt!'s multi-threaded architecture is discussed in
section~\ref{sec:multithreaded}.

\subsection{Software Architecture}\label{sec:software}

\begin{figure}[tb]
\centering
\includegraphics[width=.9\columnwidth]{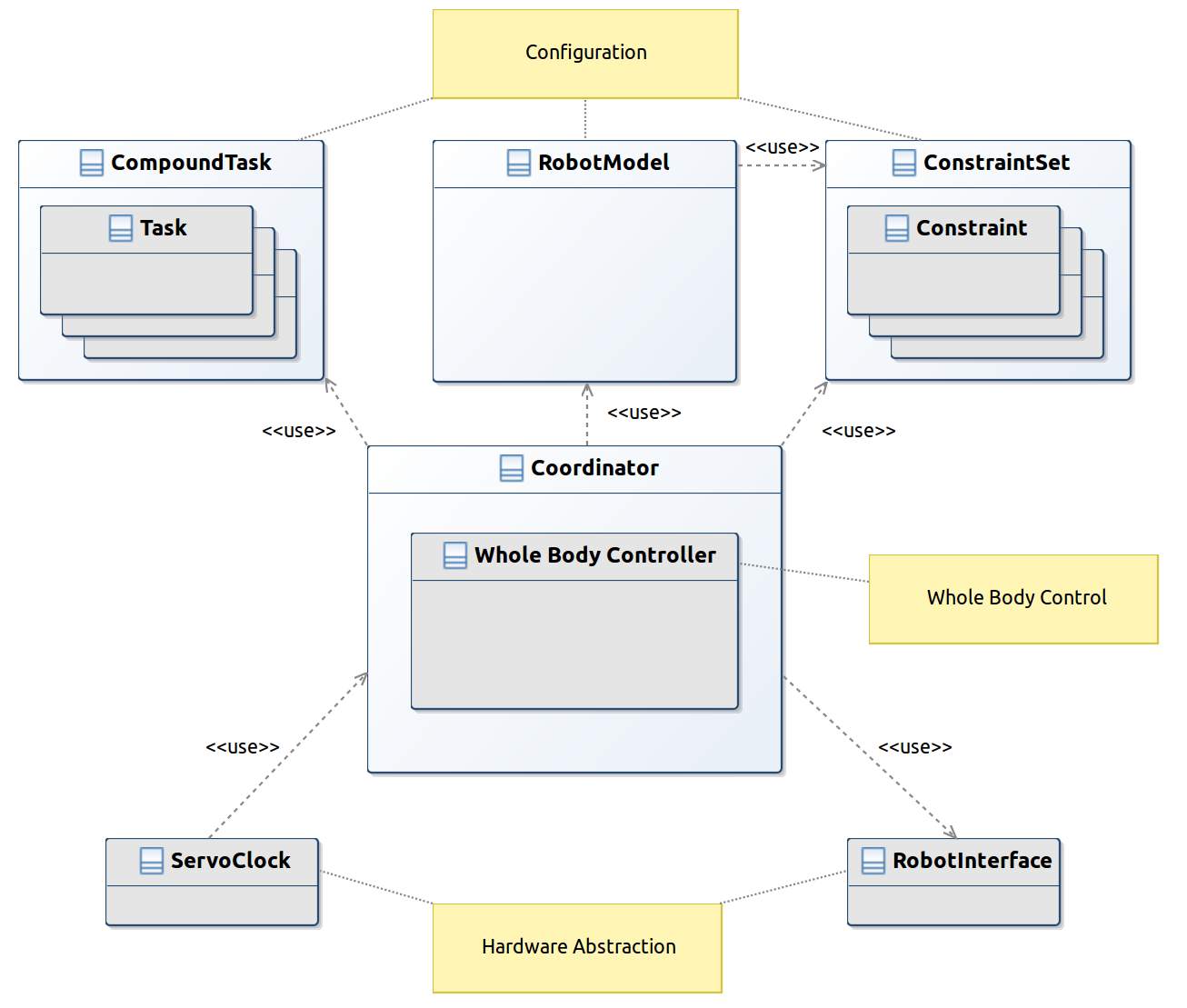}
\vspace{-.1in}
\caption{The primary software abstractions within ControlIt! consist of a
compound task, constraint set, robot model, whole body controller, servo
clock, robot interface, and coordinator. The compound task contains a set of
prioritized tasks. Tasks specify operational space or postural objectives and
contain task-space controllers; multiple tasks may have the same priority
level. Constraints specify natural physical constraints that must be satisfied
at all times and are effectively higher priority than the tasks. The robot
model computes kinematic and dynamic properties of the robot based on the
current joint states. The servo clock and robot interface constitute a
hardware abstraction layer that enables ControlIt! to work on many platforms.
The coordinator is responsible for managing the execution of the whole body
controller. Arrows indicate usage relationships between the software
abstractions. Abstractions that are dynamically extensible via plugins are
colored gray.}
\label{fig:Abstractions}
\end{figure}

The software abstractions that enable ControlIt! to instantiate and integrate
general WBOSC controllers are shown in Figure~\ref{fig:Abstractions}.  The
abstractions that are extendable via dynamically loadable plugins are colored
gray. They include tasks, constraints, the whole body controller, the servo
clock, and the robot interface. Non-extensible components include the compound
task, robot model, constraint set, and coordinator. The coordinator implements
the servo loop and uses all of the other abstractions except for the servo
clock, which implements the servo thread and controls when the coordinator
executes the next cycle of the servo loop. The software abstractions can be
divided into three general categories: configuration, whole body control, and
hardware abstraction.

\begin{figure}[tb]
\centering
\includegraphics[width=.9\columnwidth]{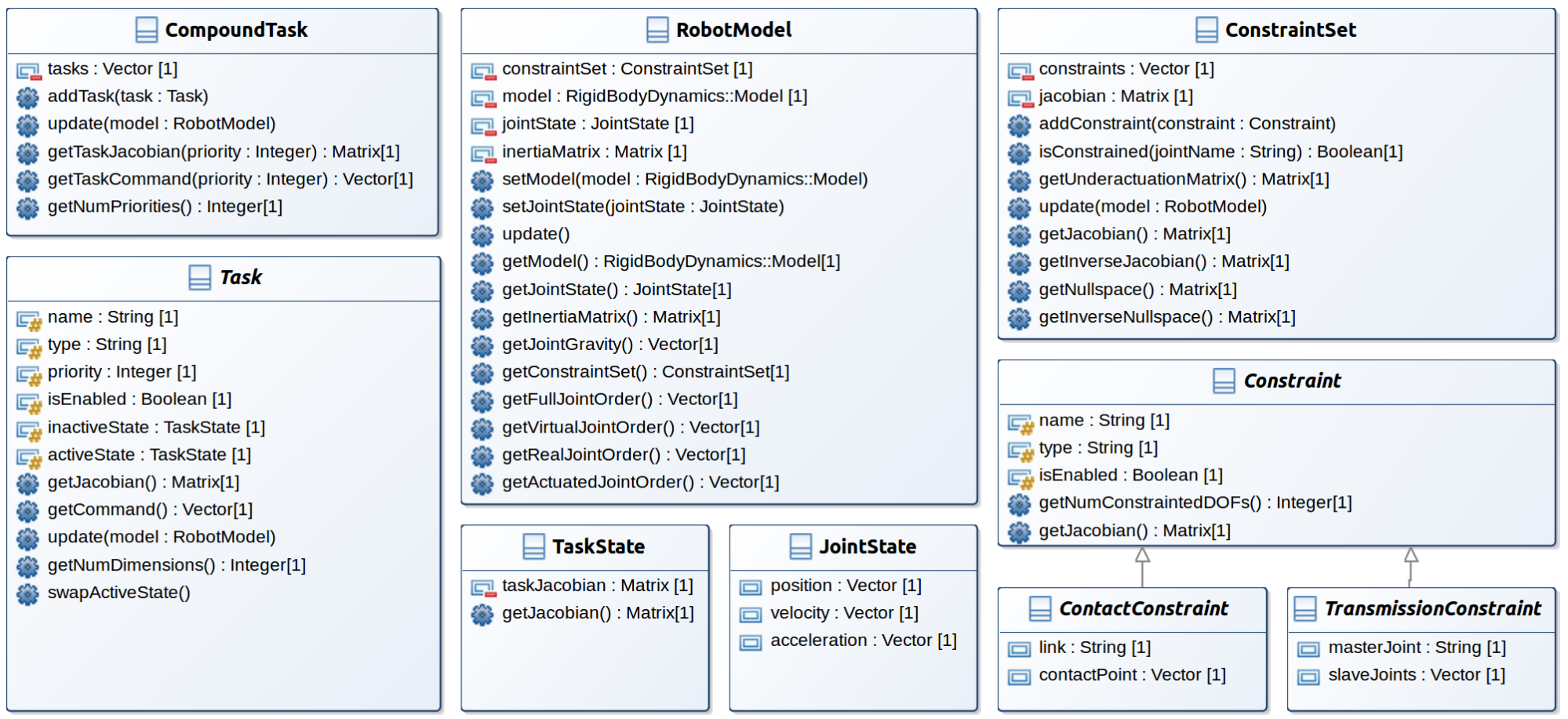}
\vspace{-.1in}
\caption{This UML diagram specifies the APIs of ControlIt!'s configuration
software abstractions. They are used to specify the objectives and constraints
of the whole body controller.}
\label{fig:ConfigurationAPIs}
\end{figure}

\textbf{Configuration.} Configuration software abstractions include the robot
model, compound task, and constraint set. Their APIs and attributes are shown
in Figure~\ref{fig:ConfigurationAPIs}. The robot model determines the
kinematic and dynamic properties of the robot and builds upon the model
provided by the Rigid Body Dynamics Library (RBDL)~\cite{Website-RBDL}, which
includes algorithms for computing forward and inverse kinematics and dynamics
and frame transformations. The kinematic and dynamic values provided by the
model are only estimates and may be incorrect, necessitating the use of a
whole body feedback controller. The robot model API includes methods for
saving and obtaining the joint state and getting properties of the robot like
the joint space inertia matrix and gravity compensation vector. There are also
methods for obtaining the joint order within the whole body controller. A
reference to the constraint set is kept within the robot model to determine
which joints are virtual (i.e., the 6-DOF free floating joints that specify a
mobile robot's position and orientation within the world frame), real, and
actuated.

The compound task and constraint set contain lists of tasks and constraints,
respectively. Tasks and constraints are abstract; concrete implementations are
added to ControlIt! through plugins. Both have names and types for easy
identification and can be enabled or disabled based on context. A task
represents an operational or postural objective for the whole body controller
to achieve. Concrete task implementations contain goal parameters that, in
combination with the robot model, produce an error. The error is used by a
controller inside the task to generate a task-space effort command\footnote{We
use the word ``effort'' to denote generalized force, i.e., force or torque.},
which is accessible through the \texttt{getCommand()} method and may be in
units of force or torque. In addition to the command, a task also provides a
Jacobian that maps from task space to joint space. The compound task combines
the commands and Jacobians of the enabled tasks and relays this information to
the whole body controller. Specifically, for each priority level, the compound
task vertically concatenates the Jacobians and commands belonging to the tasks
at the priority level. The WBOSC algorithm uses these concatenated Jacobian
and command matrices to support task prioritization and multiple tasks at the
same priority level.

\begin{figure}[tb]
\centering
\includegraphics[width=.9\columnwidth]{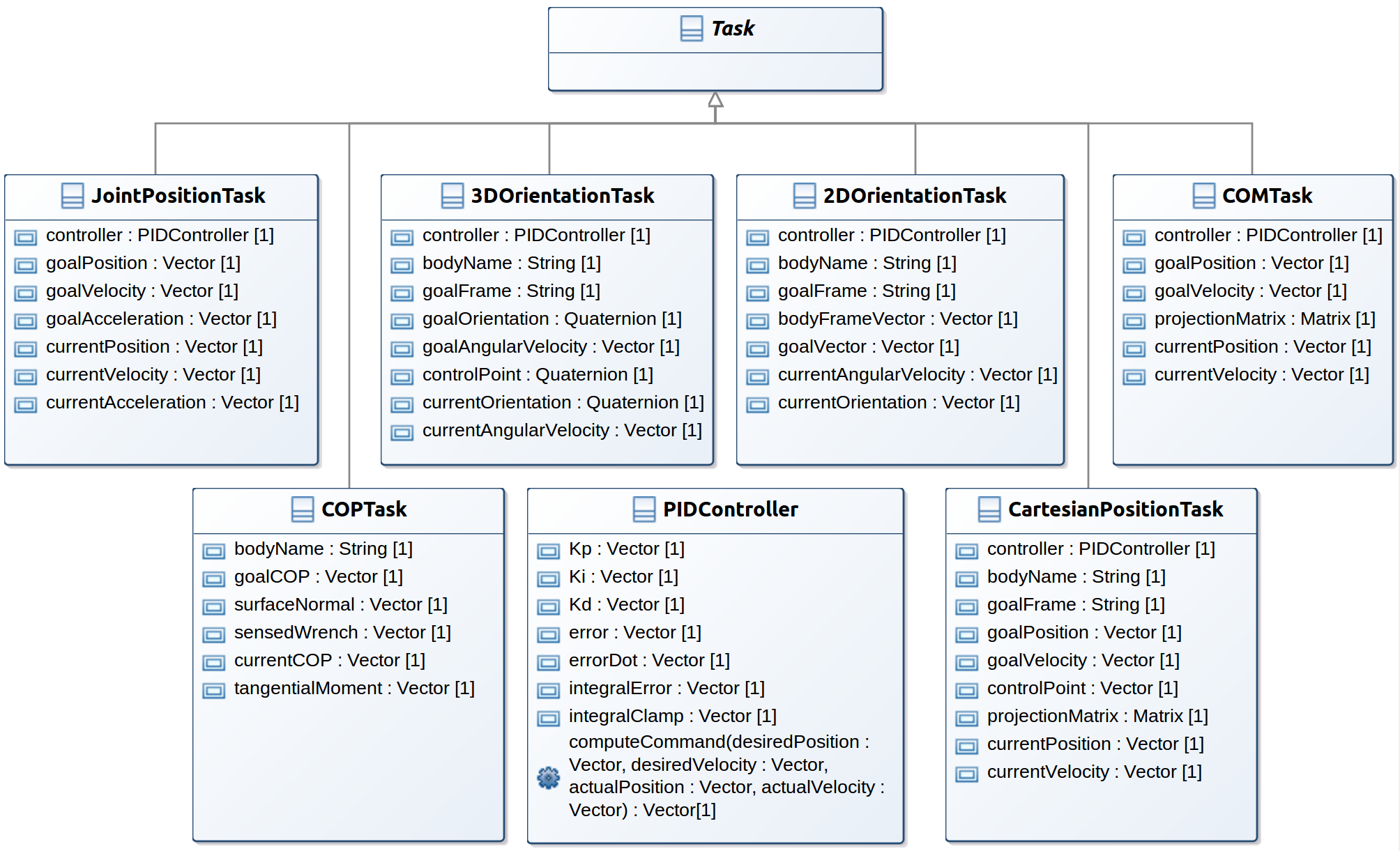}
\vspace{-.1in}
\caption{This UML class diagram shows the tasks in ControlIt!'s task library
and the PID controller that they use. Combinations of these tasks specify the
operational space and postural objectives of the whole body controller and
collectively form the compound task. Concrete tasks are implemented as
dynamically loadable plugins. ControlIt! can be easily extended with new tasks
via the plugin mechanism.}
\label{fig:TaskLibrary}
\end{figure}

\textbf{Task Library.} To encourage code reuse and enable support for basic
applications, ControlIt! comes with a task library containing commonly used-
tasks. The tasks within this library are shown in
Figure~\ref{fig:TaskLibrary}. There are currently six tasks in the library:
joint position, 2D / 3D Orientation, center of mass, Cartesian position, and
center of pressure. In the future, more tasks can be added to the library by
introducing additional plugins. Of these, the joint position, orientation,
and Cartesian position tasks have been successfully tested in hardware. The
rest have only been tested in simulation. Note that all of the tasks make use
of a PIDController. This feedback controller generates the task-space command
based on the current error and gains. Alternative types of controllers like
sliding mode control may be provided in the future.

The joint position task directly specifies the goal positions, velocities, and
accelerations of every joint in the robot. It typically defines the desired
``posture'' of the robot, which is not an operational space objective but
accounts for situations where there is sufficient redundancy within the robot
to result in non-deterministic behavior when no posture is defined.
Specifically, a posture task is necessary when the null space of all higher
priority tasks and constraints is not nil, and the best practice is to always
include one as the lowest priority task in the compound task. The
joint position task has an input parameter called \texttt{goalAcceleration} to enable smooth
transitions between joint positions. The goal acceleration is a desired
acceleration that is added as a feedforward command to the control law. The
\texttt{currentAcceleration} output parameter is a copy of the \texttt{goalAcceleration} 
parameter and is used for debugging purposes.

The 2D and 3D orientation tasks are used to control the orientation of a link
on the robot. They differ in terms of how the orientations are specified.
Whereas the 2D orientation is specified by a vector in the frame of the body
being oriented, the 3D orientation is specified using a quaternion. The
purpose of providing a 2D orientation task even though a 3D orientation could
be used is to reduce computational overhead when only two degrees of
orientation control is required. For example, a 2D orientation task is used to
control the heading of Trikey, a 3 wheeled holonomic mobile robot, as shown in
Figure~\ref{fig:TrikeyHeading}, whereas a 3D orientation task is used to
control the orientation of Dreamer's end effectors, as shown in
Figure~\ref{fig:DreamerEndEffectorControl}(b). Visualizations of these two task-level controllers are given in~\ref{sec:introspection}. The 2D orientation
task does not include a \texttt{goalAngularVelocity} input parameter because
its current implementation assumes the goal velocity is always zero. This
assumption can be easily removed in the future by modifying the control law to
include a non-zero goal velocity.

The Center Of Mass (COM) task controls the location of the robot's COM, which
is derived from the robot model. It is useful when balancing since it can
ensure that the robot's configuration always results in the COM being above
the convex polygon surrounding the supports holding the robot up. The Center
Of Pressure (COP) task controls the center of pressure of a link that is in
contact with the ground. It is particularly useful for biped robots containing
feet since it can help ensure that the COP of a foot remains within the
boundaries of the foot thereby preventing the foot from rolling. The Cartesian
position task controls the operational space location of a point on the robot.
Typically, this means the location of an end effector in a frame that is
specified by the user and is by default the world frame. For example, it is
used to position Dreamer's end effectors in front of Dreamer as shown in
Figure~\ref{fig:DreamerEndEffectorControl}. As indicated by the figure,
multiple Cartesian position tasks may exist within a compound task, as long as
they control different points on the robot.

As previously mentioned, the aforementioned tasks are those that are currently
included with ControlIt!. They are implemented as plugins that are dynamically
loaded on-demand during the controller configuration process. Additional tasks
may be added in the future. For example, an external force task may be added
that controls a robot to assert a certain amount of force against an external
obstacle. In addition, an internal force task may be added to control the
internal tensions between multiple contact points. A prototype of such a task
was successfully used during NASA JSC DRC critical design review\footnote{As a
Track A DRC team, NASA JSC was required to undergo a critical design review by
DARPA officials in June 2013, which was in the middle of the period leading up
to the DRC Trials in December 2013. The results of the review determined
whether the team would continue to receive funding and proceed to compete in
the DRC Trials as a Track A team. NASA JSC was one of six Track A teams to
pass this critical design review.} to make Valkyrie to walk in simulation, as
shown in~\ref{sec:introspection}, but is not included in the current task
library due to the need for additional testing and refinement. For the walking
behavior, ControlIt!'s compound task included a COM Task, internal tensions
task, posture task, and, for each foot, a COP, Cartesian position, and
orientation task.

\textbf{Constraints.} A constraint specifies natural physical limits of the
robot. There are two types of constraints: \texttt{ContactConstraint} and
\texttt{TransmissionConstraint}. Contact constraints specify places where a
robot touches the environment. Transmission constraints specify dependences
between joints, which occur when, for example, joints are co-actuated. The
parent \texttt{Constraint} class includes methods for obtaining the number of
DOFs that are constrained and the Jacobian of the constraint. Contact
constraints have a \texttt{getJoint()} method that specifies the parent joint
of the link that is constrained. Transmission constraints have a master joint
that is actuated and a set of slave joints that are co-actuated with the
master joint. Unlike tasks, constraints do not have commands since they simply
specify the nullspace within which all tasks must operate. Like the compound
task, the constraint set computes a Jacobian that is the vertical
concatenation of all the constraint Jacobians. In addition, it provides an
update method that computes both the null space projector and
$\overline{UN_c}$ (defined in Equation~(\ref{eq:task})), accessors for these
matrices, and methods for determining whether a particular joint is
constrained. The whole body controller uses this information to ensure all of
the constraints are met. While it is true that contact constraints are
mathematically similar to tasks without an error term, we wanted to
distinguish between the two since they serve significantly different purposes:
tasks denote a user's control objectives while constraints denote a robot's
physical limits. We did not want to confuse the API by using the same software
abstraction for both purposes. Furthermore, by separating tasks and constraints,
the API will be easier to extend to support optimization based controllers with 
inequality constraints.

\begin{figure}[tb]
\centering
\includegraphics[width=.9\columnwidth]{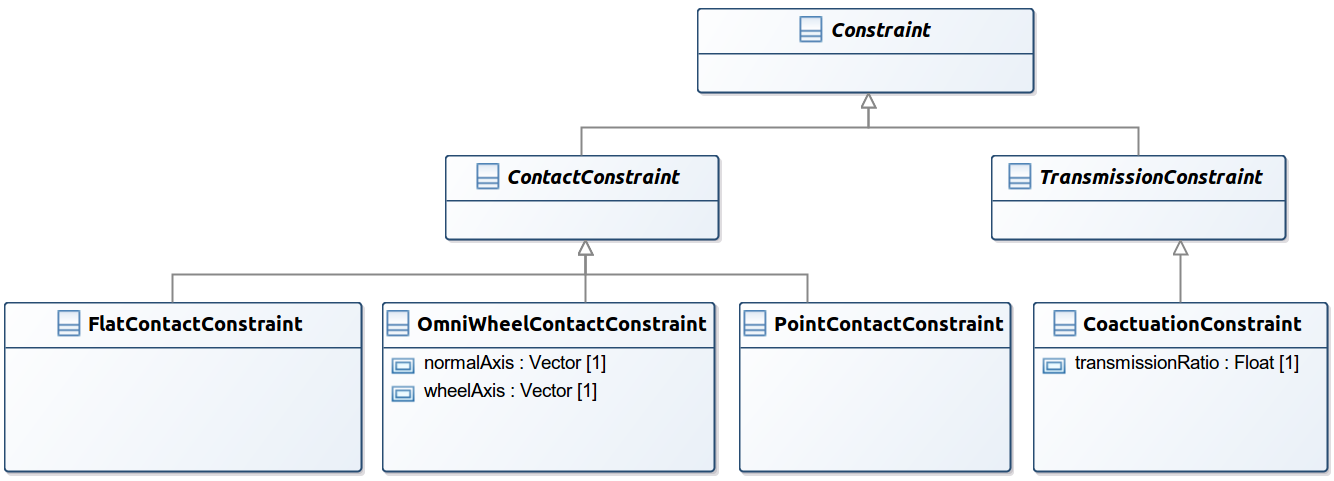}
\vspace{-.1in}
\caption{This UML class diagram shows the constraints in ControlIt!'s
constraint library. Combinations of these constraints specify natural physical
limits of the robot and constitute the constraint set. Concrete constraints
are implemented as dynamically loadable plugins. Additional constraints can be
easily added via the plugin mechanism.}
\label{fig:ConstraintLibrary}
\end{figure}

\textbf{Constraint Library.} Constraints included in ControlIt!'s constraint
library are shown in Figure~\ref{fig:ConstraintLibrary}. Contact constraints
include the flat contact constraint, omni wheel contact constraint, and point
contact constraint. The flat contact constraint restricts both link
translation and rotation. The omni wheel contact constraint restricts one
rotational DOF and one translational DOF based on the current orientation of
the wheel. Point contact constraint restricts just link translation. One
transmission constraint called \texttt{CoactuationConstraint} is provided that
enables ControlIt! to handle robots with two co-actuated joints, like the
torso pitch joints in Dreamer. It includes a transmission ratio specification
to handle situations where the relationship between the master joint and slave
joint is not one-to-one. Currently only the two-joint co-actuation case is
supported, though a more generalized constraint that supports more than two
co-actuated joints could be trivially added in the future.  Specifically,
another child class of \texttt{TransmissionConstraint} can be added as a plugin to
support the co-actuation of more than two joints by adding more rows to the
constraint's Jacobian.  Like the task library, the constraint library can
easily be extended with new constraints via the plugin mechanism used by
ControlIt!.

\begin{figure}[tb]
\centering
\includegraphics[width=.9\columnwidth]{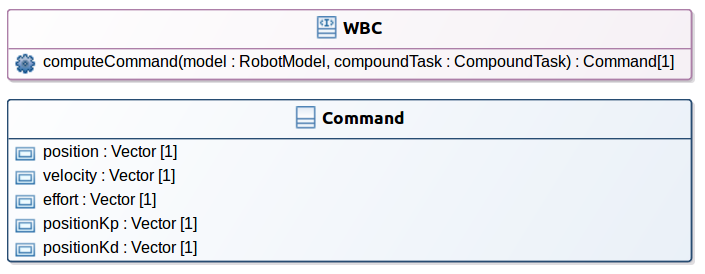}
\vspace{-.1in}
\caption{The WBC software abstractions within ControlIt! consist of an
interface called WBC and a class called Command. The WBC interface defines a
single method called computeCommand that takes two input parameters, the robot
model, which includes the constraint set, and the compound task. It returns a
Command object. The command includes position, velocity, effort, and position
controller gains. Depending on the type of joint controller used, one or more
of the member variables inside the command may not be used. For example, a
pure force or torque-controlled robot will only use the effort specification
within the command.}
\label{fig:WBC_API}
\end{figure}

\textbf{Whole body control.} The class diagrams for the whole body control
software abstractions are shown in Figure~\ref{fig:WBC_API}. There are two
classes: WBC and Command. WBC is an interface that contains a single
computeCommand() method. This method takes as input the robot model, which
includes the constraint set, and the compound task. It performs the WBC
computations that generate a command for each joint under its control and
returns it within a Command object. The Command object specifies the desired
position, velocity, effort, and position controller gains. Note that not all
of these variables need to be used. For example, a robot that is purely effort
controlled will only use the effort command. The optional fields within the
command are included to support robots with joints that are position or
impedance controlled.

\begin{figure}[tb]
\centering
\includegraphics[width=.9\columnwidth]{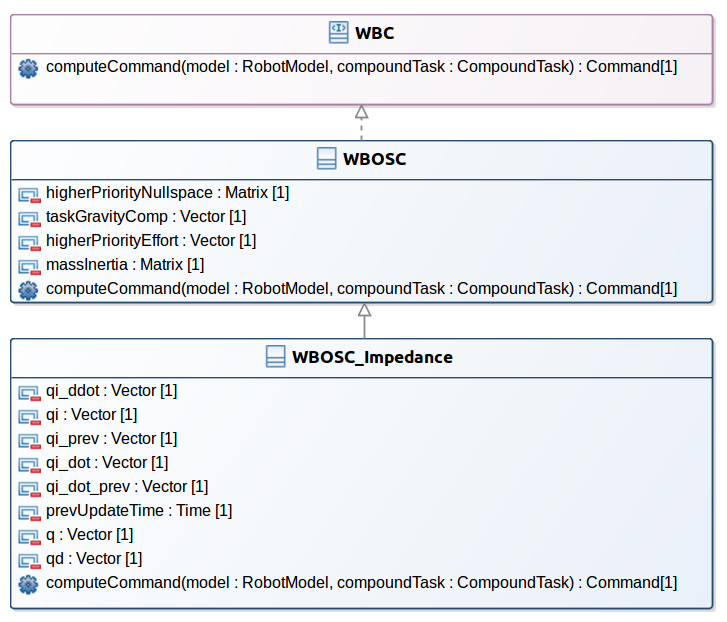}
\vspace{-.1in}
\caption{ControlIt! currently includes two plugins in its WBC plugin library.
They consist of WBOSC and WBOSC\_Impedance. WBOSC implements the actual WBOSC
algorithm that takes a holistic view of the robot and achieves multiple
prioritized task objectives using nullspace projection. It outputs a pure
effort command and is use with effort-controlled robots like Dreamer. The
second plugin, WBOSC\_Impedance, extends WBOSC with an internal robot model
that specifies the desired joint positions and velocities based on the torque
commands generated by WBOSC. This is useful to support robots with joint
impedance controllers, an example of which is NASA JSC's Valkryie.}
\label{fig:WBC_Plugins}
\end{figure}

The whole body controller within ControlIt! is dynamically loaded as a plugin
using ROS pluginlib~\cite{Website-ROS-Pluginlib}. Two plugins are currently
available as shown in Figure~\ref{fig:WBC_Plugins}. They include WBOSC and
WBOSC\_Impedance. The WBOSC plugin implements the WBOSC algorithm. It computes
the nullspace of the constraint set and projects the task commands through
this nullspace. Task commands are iteratively included into the final command
based on priority. The commands of tasks at a particular priority level are
projected through the nullspaces of all higher priority tasks and the
constraint set. This ensures that all constraints are met and that higher
priority tasks override lower priority tasks. The output of WBOSC is an effort
command that can be sent to effort controlled robots like Dreamer. The member
variables within the WBOSC plugin ensure that memory is pre-allocated, which
reduces execution time jitter and thus increases real-time predictability.

To support impedance-controlled robots, ControlIt! also comes with the
WBOSC\_Impedance plugin. Unlike effort-controlled robots, impedance-controlled
robots take more than just effort commands. Specifically, in addition to
effort, impedance controllers also take desired position and velocity
commands, and optionally position controller gains when controller gain
scheduling is desired. The benefit of using impedance control is the ability
to attain higher levels of impedance. This is possible since the position and
velocity control loop can be closed by the embedded joint controller, which
typically has a higher servo frequency and lower communication latency than
the WBC controller. The WBOSC\_Impedance plugin extends the WBOSC plugin with
an internal model that converts the effort commands generated by the WBOSC
algorithm into expected joint positions and velocities. The member variables
within the WBOSC\_Impedance plugin that start with ``qi\_'' hold the internal
model's joint states. The prevUpdateTime member variable records when this
internal model was last updated. Each time computeCommand is called,
WBOSC\_Impedance computes the desired effort command using WBOSC. It then uses
this effort command along with the robot model to determine the desired
accelerations of each joint. WBOSC\_Impedance then updates the internal model
based on these acceleration values, the time since the last update, the
previous state of the internal model, and the actual position and velocity of
the joints. The derived joint positions, velocities, and efforts are saved
within a Command object, which is returned. This control strategy was used on
the upper body of NASA JSC's Valkyrie robot to perform several DRC
manipulation tasks as previously mentioned.

\begin{figure}[tb]
\centering
\includegraphics[width=.9\columnwidth]{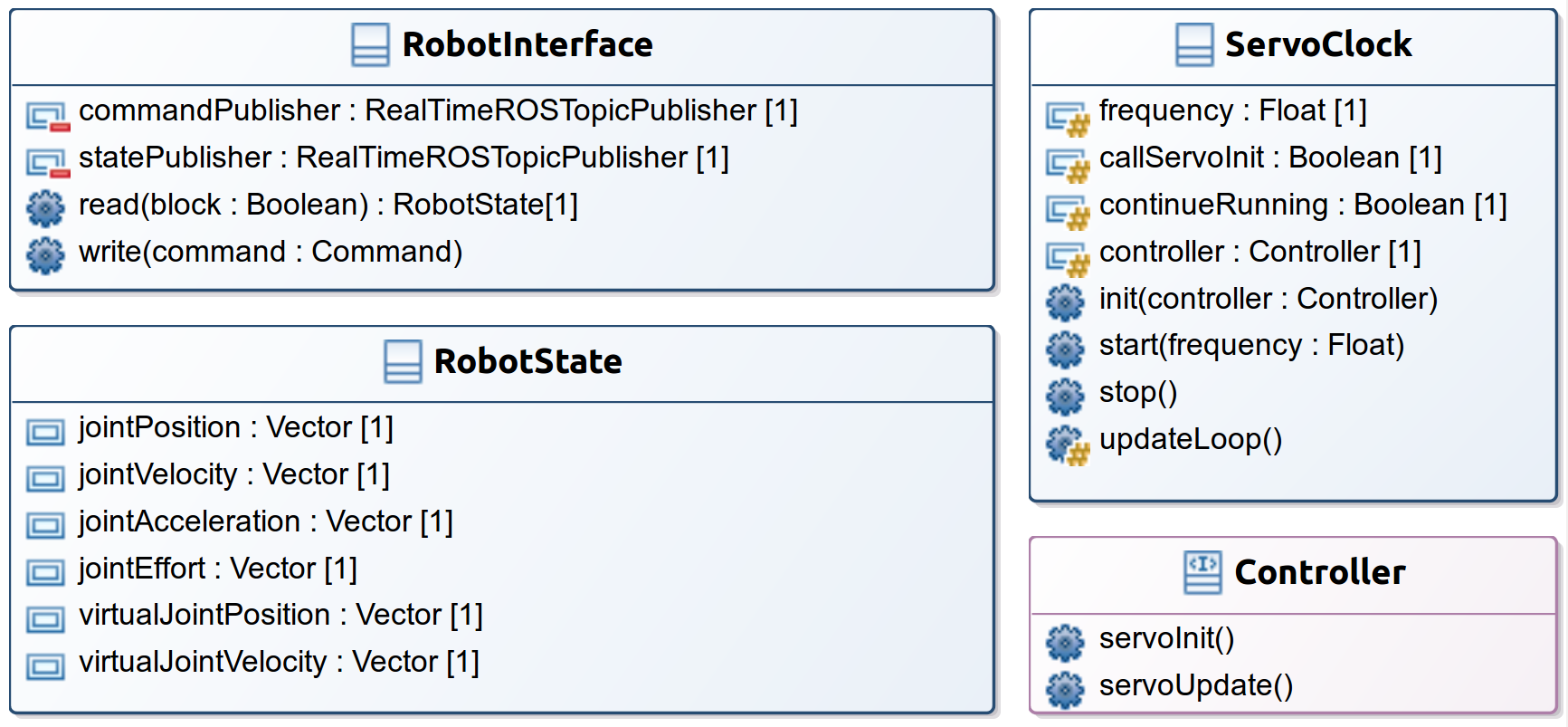}
\vspace{-.1in}
\caption{ControlIt! employs a hardware abstraction layer that consists of a
RobotInterface and a Clock. The RobotInterface has two methods: \texttt{read}
and \texttt{write}. The read method returns a \texttt{RobotState} object that
includes details about the robot joint positions, velocities, accelerations,
and efforts. The write method takes as input a Command object and issues the
command to the robot joints.}
\label{fig:HardwareInterface}
\end{figure}

\textbf{Hardware abstraction.} To enable support for a wide variety of robot
platforms, ControlIt! includes a hardware abstraction layer consisting of two
abstract classes, the \texttt{RobotInterface} and the \texttt{ServoClock}, as
shown in Figure~\ref{fig:HardwareInterface}. Concrete implementations are
provided through dynamically loadable plugins. \texttt{RobotInterface} is
responsible for obtaining the robot's joint state and sending the command from
the whole body controller to the robot. For diagnostic purposes, it also
publishes the state and command information onto ROS topics using a real-time
ROS topic publisher, which uses a thread-pool to offload the publishing process
from the servo thread. \texttt{ServoClock} instantiates the servo thread and
contains a reference to a Controller, which is implemented by the Coordinator.
\texttt{ServoClock} is responsible for initializing the controller by calling
\texttt{servoInit()} and then periodically executing the servo loop by calling
the \texttt{servoUpdate()} method. Initialization using the actual servo
thread is needed  to handle situations where certain initialization tasks can
only be done by the servo thread. This occurs, for example, when the servo
thread is part of a real-time context meaning only it can initialize certain
real-time resources.

\begin{figure}[tb]
\centering
\includegraphics[width=.9\columnwidth]{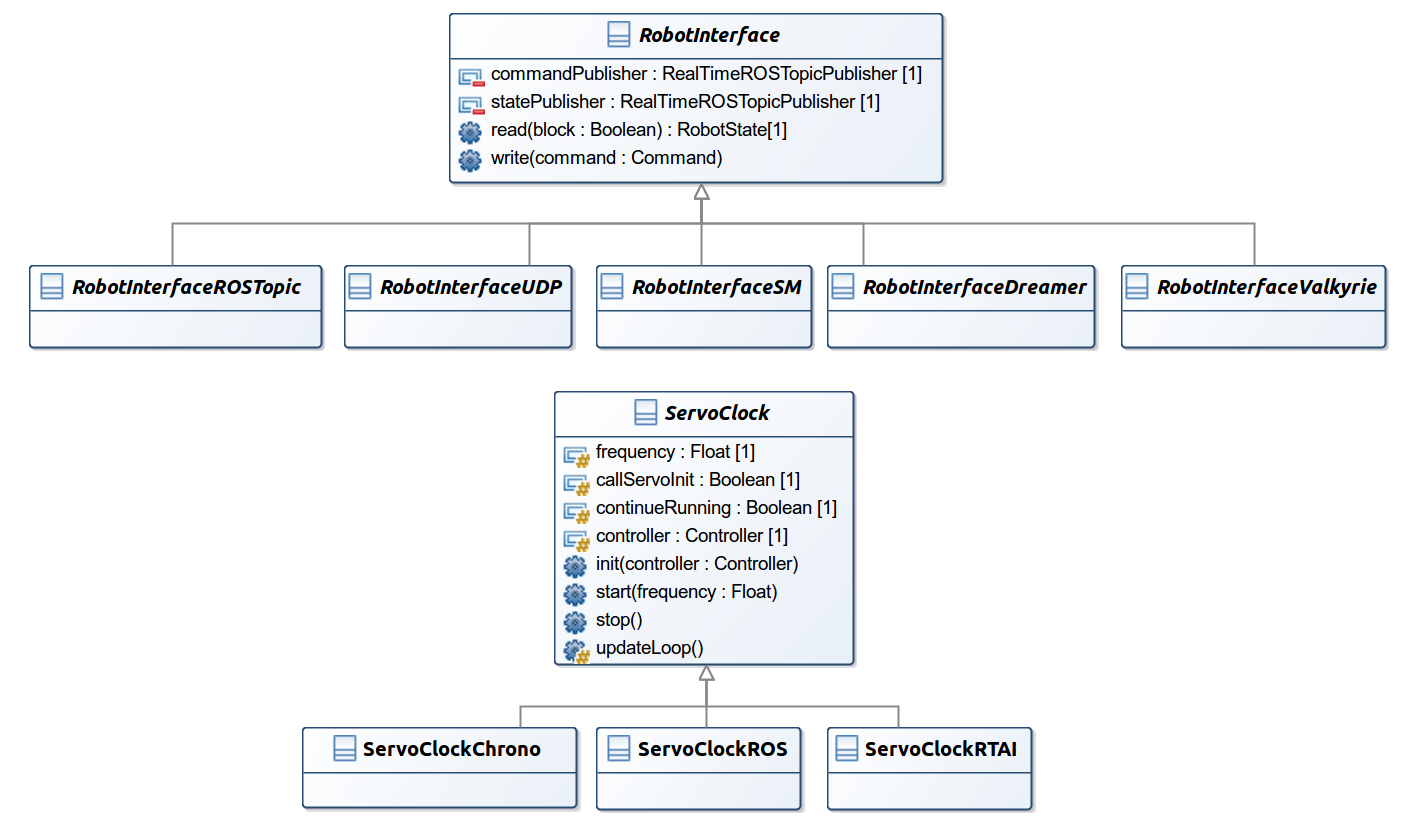}
\vspace{-.1in}
\caption{The robot interface plugins that are currently available include
support for the following transport protocols: ROS Topic, UDP, and shared
memory. There are also specialized robot interfaces for Dreamer and Valkyrie.
The servo clocks provided include support for \texttt{std::chrono}, ROS time,
and RTAI time.}
\label{fig:HardwareInterfacePlugins}
\end{figure}

ControlIt! includes libraries of \texttt{RobotInterface} and the
\texttt{ServoClock} plugins as shown in
Figure~\ref{fig:HardwareInterfacePlugins}. RobotInterface plugins include
general ones that communicate with a robot via three different transport
layers: ROS topics (RobotInterfaceROSTopic), UDP datagrams (RobotIntefaceUDP),
and shared memory (RobotInterfaceSM). These are meant for general use --
ControlIt! includes generic Gazebo plugins and abstract classes that
facilitate the creation of software adapters for allowing simulated and real
robots to communicate with ControlIt! using these three transport layers.
Among the three transport layers, shared memory has the lowest latency and is
most reliable in terms of message loss. It uses the ROS shared memory
interface package~\cite{Website-ROS-SHM}, which is based on boost's
interprocess communication library.

In addition to general \texttt{RobotInterface} plugins, ControlIt! also
includes two robot-specific plugins, one for Dreamer
(\texttt{RobotInterfaceDreamer}), and one for Valkyrie
(\texttt{RobotInterfaceValkyrie}). \texttt{RobotInterfaceDreamer} interfaces
with a RTAI real-time shared memory segment that is created by the robot's
software interface called the M3 Server. It also implements separate PID
controllers for robot joints that are not controlled by WBC. They include the
finger joints in the right hand, the left gripper joint, the neck joints, and
the head joints. In the current implementation, these joints are fixed  from
WBC's perspective. \texttt{RobotInterfaceValkyrie} interfaces with shared
memory segment created by Valkyrie's software interface. This involves
integration with a controller manager provided by
ros\_control~\cite{Website_ROS_Control} to gain access to robot resources.

ControlIt! includes several \texttt{ServoClock} plugins to enable flexibility
in the way the servo thread is instantiated and configured to be periodic. The
current \texttt{ServoClock} plugin library includes plugins for supporting
servo threads based on a ROS timer, a C++ \texttt{std::chrono} timer, or an
RTAI timer. Support for additional methods can be included in the future as
additional plugins.

\subsection{Configuration and Integration}\label{sec:integration}

Support for configuration and integration is important because as a software framework
ControlIt! is expected to be (1) used in many different applications and
hardware platforms that require different whole body controllers and (2) just
one component in a complex application consisting of many components.  In
addition, ControlIt!'s configuration and integration capabilities directly
impacts the software's usability, which must be high to achieve the goal of
widespread use. ControlIt! supports integration through four mechanisms: (1)
\textbf{parameter reflection}, which exposes controller parameters to other
objects within ControlIt! and is used by the other two mechanisms, (2)
\textbf{parameter binding}, which enables the parameters to be connected to
external processes through an extensible set of transport layers, (3)
\textbf{events}, which enable parameter changes to trigger the execution of
external processes without the use of polling, and (4) \textbf{services},
which enable external processes to query information about the controller.
ControlIt! supports configuration through scripts that enable users to specify
the structure of the compound task and constraint set, the type of whole body
controller and hardware interface to use, the initial values of the
parameters, the parameter bindings, and the events. These scripts are
interpreted during ControlIt!'s initialization to automatically instantiate
the desired whole body controller and integrate it into the rest of the
system. Details of ControlIt!'s support for configuration and integration are
now discussed.

\begin{figure}[tb]
\centering
\includegraphics[width=.9\columnwidth]{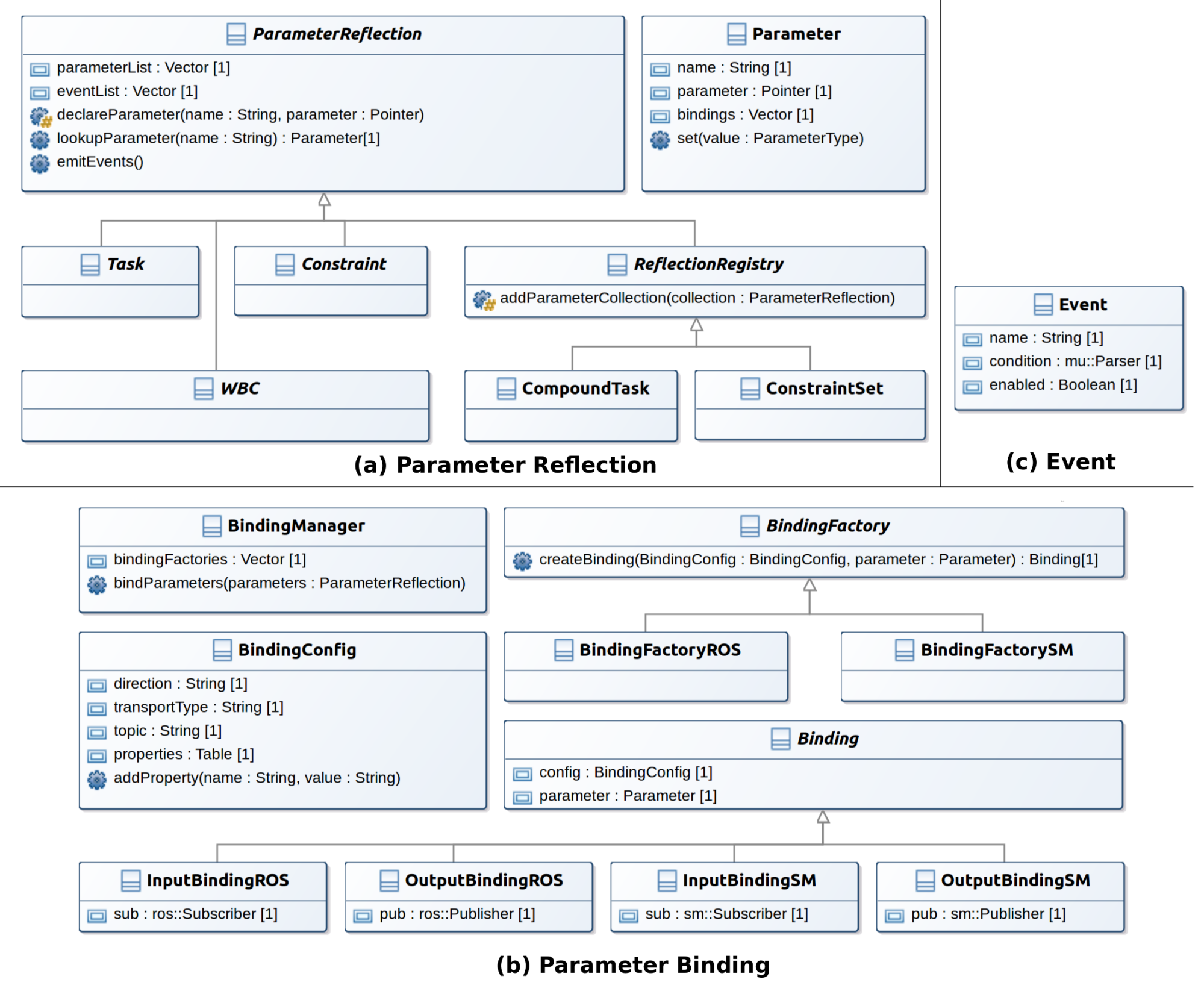}
\vspace{-.1in}
\caption{ControlIt! includes three mechanisms for integration: parameter
reflection, parameter binding, and events. Sub-figure (a) shows the parameter
reflection mechanism that enables parameters to be exposed to other objects
within ControlIt! including the parameter binding and event mechanisms. Sub-
figure (b) shows the parameter binding mechanism that enables parameters to be
bound to an extensible set of transport layers, which enables them to be
accessed by external processes. Sub-figure (c) shows an event definition.
Events are stored within ParameterReflection objects and are emitted at the
end of the servo loop. They enable external processes to be notified when a
logical expression over a set of parameters transitions from being false to
true and eliminates the need for external processes to poll for state changes
within ControlIt!.}
\label{fig:Integration}
\end{figure}

\textbf{Parameter Reflection.} Parameter reflection was originally introduced
in Stanford-WBC. It defines a ParameterReflection parent class through which
child class member variables can be exposed to other objects within
ControlIt!. The API and class hierarchy of the ParameterReflection class is
shown in Figure~\ref{fig:Integration} (a). Parameter reflection enables
internal control parameters to be exposed to other objects within ControlIt!.
It consists of an abstract parent called ParameterReflection that provides
methods for declaring and looking up parameters. When a parameter is declared,
it is encapsulated within a Parameter object, which contains a name, pointer
to the actual variable, a list of bindings, and a method to set the
parameter's value. Subclasses of ParameterReflection are able to declare their
member variables as parameters and thus make them compatible with ControlIt's
parameter binding and event mechanism.

\textbf{Parameter Binding.} Parameter binding enables the integration of
ControlIt! with other processes in the system by connecting parameters to an
extensible set of transport layers. Its API and class hierarchy is shown in
Figure~\ref{fig:Integration} (b). The classes that constitute the parameter
binding mechanism consist of a BindingManager that maintains a set of
BindingFactory objects that actually create the bindings, and a
\texttt{BindingConfig} object that specifies properties of a binding. The
required properties include the binding direction (either input or output),
the transport type, which is a string that must match the name of a Binding
provided by a BindingFactory plugin, and a topic to which the parameter is
bound. The \texttt{BindingConfig} also contains an extensible list of name-
value properties that is transport protocol specific. For example, transport-
specific parameters for ROS topic output bindings include the publish rate,
the queue size, and whether the latest value published should be latched.

During the initialization process, \texttt{BindingConfig} objects are stored
as parameters within a ParameterReflection object, which is passed to the
BindingManager. The BindingManager searches through its BindingFactory
objects, which are dynamically loaded via plugins, for factories that are able
to create the desired binding. The current bindings in ControlIt's binding
library include input and output bindings for ROS topics and shared memory
topics. More can be easily added in the future via the plugin architecture.
The newly created Binding objects are stored in the parameter's Parameter
object. When a parameter's value is set via \texttt{Parameter.set()}, the new
value is transmitted through output bindings to which the parameter is bound.
This enables changes in ControlIt! parameters to be published onto various
transport layers and topics notifying external processes of the latest values
of the parameters. Similarly, when an external process publishes a value onto
a transport layer and topic to which a parameter is bound via an input
binding, the parameter's value is updated to be the published value. This
enables, for example, external processes to dynamically change a task's
references or controller gains, which is necessary for integration.

\textbf{Events.} Events contain a logical expression over parameters that are
interpreted via muParser~\cite{Website-muParser}, an open-source math parser
library. Its API is shown in Figure~\ref{fig:Integration} (c). Events are
stored in the ParameterReflection parent class. The servo thread calls
ParameterReflection.emitEvents() at the end of every servo cycle. The names of
events whose condition expression evaluates to true are published on ROS topic
\texttt{/[controller name]/events}. Events contain a boolean variable called
``enabled'' that is used to prevent an event from continuously firing when the
condition expression remains true since this would likely flood the events ROS
topic. Instead, events maintain a fire-once semantic meaning they only fire
when the condition expression changes from false to true.

\textbf{Service-based controller introspection capabilities.} To further
assist ControlIt! integration, into a larger system, ControlIt! also includes
a set of service-based introspection capabilities. Unlike ROS topics, which
are asynchronous unidirectional, ROS services are bi-directional and
synchronous.  ControlIt! uses this capability to enable external processes to
query certain controller properties as it is running. For example, two often-
used services include \texttt{/[controller
name]/diagnostics/getTaskParameters}, which returns a list of all tasks in the
compound task, the parameters, and their parameter values, and
\texttt{/[controller name]/diagnostics/getRealJointIndices}, which returns the
ordering of all real joints in the robot. This is useful to determine the
joint order when updating the reference positions of a posture task or
interpreting the meaning of the posture task's error vector. A full list of
ControlIt's service-based controller introspection capabilities is provided
in~\ref{sec:introspection}.

\textbf{Script-based configuration and initialization.} As previously
mentioned, ControlIt! supports script-based configuration specification and
initialization enabling integration into different applications and platforms
without being recompiled. This is necessary given the plethora of properties
that must be defined and the wide range of anticipated applications and
hardware platforms. To instantiate a whole body controller using ControlIt!,
the user must specify many things including the compound task, constraint set,
whole body controller, robot interface, servo clock, initial parameter values,
parameter bindings, and events. In addition, there are numerous controller
parameters as defined in Appendix~\ref{sec:parameters}. ControlIt! enables
users to define the primary WBC configuration and integration abstractions
including tasks, constraints, compound tasks, constraint set, parameter
bindings, and events via a YAML file whose syntax is given in
Appendix~\ref{sec:configuration}. The remaining parameters are defined through
the ROS parameter server, which can also be initialized via another YAML file
that is loaded via a ROS launch file~\cite{Website-ROS-Launch}. ROS launch is
actually a powerful tool for loading parameters and instantiating processes.
ControlIt! leverages this capability to enable users to initialize and execute
a ControlIt! whole body controller using a single command.

\subsection{Multi-threaded Architecture}\label{sec:multithreaded}

\begin{figure}[tb]
\centering
\includegraphics[width=.9\columnwidth]{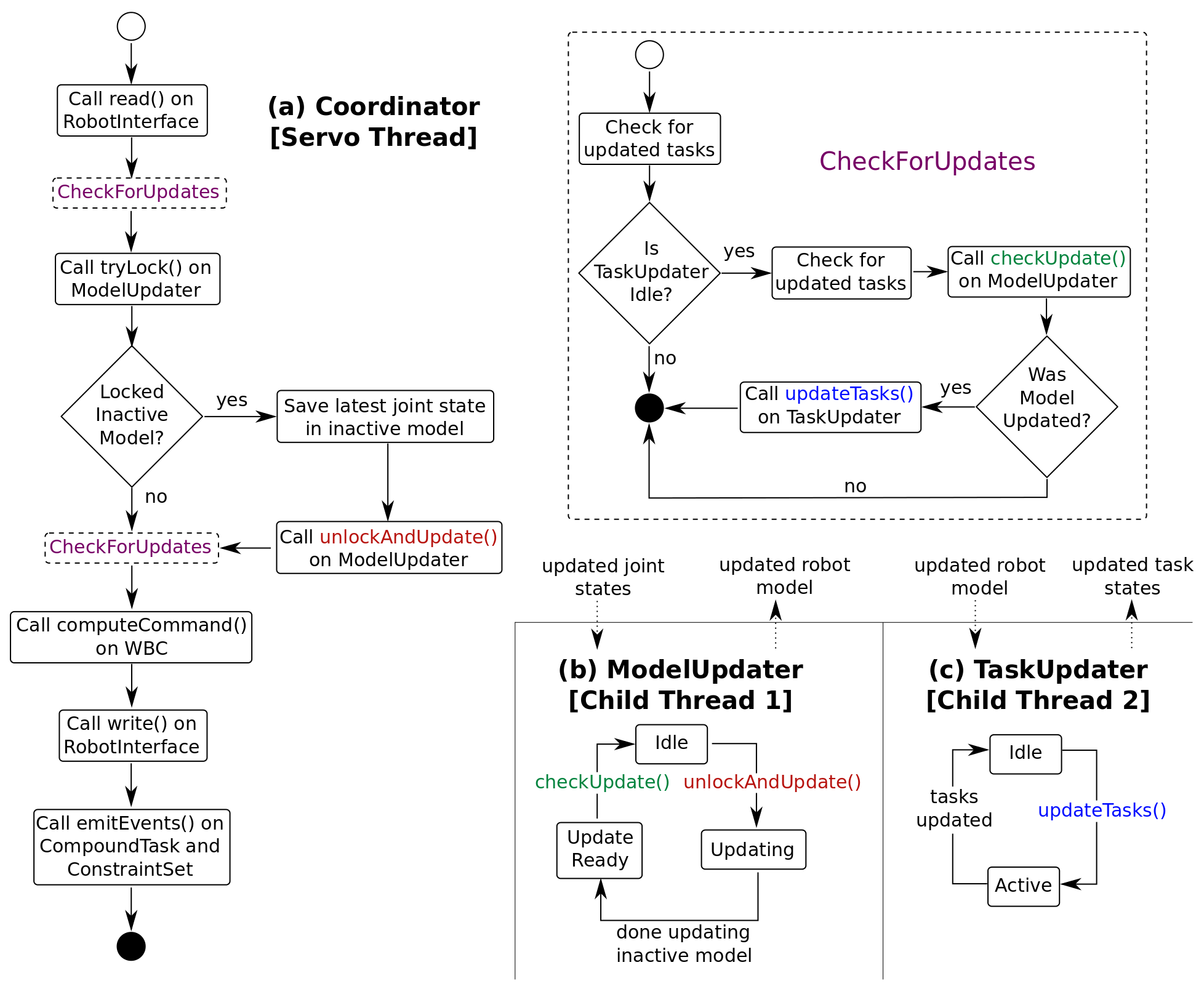}
\vspace{-.1in}
\caption{To achieve higher servo frequencies, ControlIt! employs a multi-
threaded architecture consisting of three threads: (a) \texttt{Servo}, (b)
\texttt{ModelUpdater}, and (c) \texttt{TaskUpdater}. Servo is a real-time
thread whereas \texttt{ModelUpdater} and \texttt{TaskUpdater} are non-real-
time threads. The names are self-descriptive. This figure shows the behavior
and interactions of these threads. At a high level, Servo gives
\texttt{ModelUpdater} the latest joint states and receives an updated robot
model. It also gives \texttt{TaskUpdater} an updated robot model and receives
updated state for each task, which includes the task Jacobians. To prevent
Servo from blocking due to contention between it and the other threads, which
is necessary for real-time operation, ControlIt! maintains two copies of the
robot model and two copies of the state for each task -- an ``active'' one and
an ``inactive'' one.  Active versions are used solely by Servo. Inactive
versions are updated by the child threads. To get updates from the child
threads, Servo swaps the active and inactive versions when it can be done in a
non-blocking and safe manner. It does this by calling the non-blocking
\texttt{tryLock()} operation on the mutex protecting the inactive version of
the robot model and only performing the swap when it successfully obtains the
lock. The swapping of task state is kept non-blocking and safe through FSM
design -- a task will only indicate it has updated state after the
\texttt{TaskUpdater} thread is done updating it. To prevent contention between
the child threads, the inactive and active robot models can only be swapped
when \texttt{TaskUpdater} is idle. To further reduce unnecessary computations,
\texttt{TaskUpdater} only executes after the robot model is swapped.}
\label{fig:MultiThreadedArch}
\end{figure}

Higher servo frequencies can be achieved by decreasing the amount of
computation in the servo loop. The amount of computation can be reduced
because robots typically move very little during one servo period, which is
usually 1ms. Thus, state that depends on the robot configuration like the
robot model and task Jacobians often do not need to be updated every servo
cycle. ControlIt! takes advantage of this possibility by offloading the
updating of the robot model and the task states, which include the task
Jacobians, into child threads. Specifically, ControlIt! uses three threads as
shown in Figure~\ref{fig:MultiThreadedArch}. They include (1) a Servo thread
that executes the actual servo loop, (2) a \texttt{ModelUpdater} thread that
is responsible for updating the robot model, which includes the kinematics,
inertia matrix, gravity compensation vector, the constraint set, and the
virtual linkage model, and (3) a \texttt{TaskUpdater} thread that is
responsible for updating the states of each task in the compound task, which
includes the task Jacobians. The Servo thread is instantiated by the
ServoClock and can thus be real-time when, for example,
\texttt{ServoClockRTAI} is used. \texttt{ModelUpdater} and TaskUpdater are
child threads that do not operate in a real-time manner. From a high-level
perspective, Servo provides \texttt{ModelUpdater} with the latest joint
states. The \texttt{ModelUpdater} uses this information to update the robot
model in parallel with the Servo thread, and provides the updated robot model
to the Servo when complete. Whenever the robot model is updated, the Servo
thread provides the updated model to the TaskUpdater thread, which updates the
task states. These updated task states are then provided to the Servo thread.
Details on how this process is achieved in a manner that is non-blocking and
safe are now discussed.

Two key requirements of the multi-threaded architecture are (1) the Servo
thread must not block and (2) there must not be any race conditions between
threads. The first requirement implies that the servo thread cannot call the
blocking lock() method on the mutexes protecting the shared states between it
and the child threads.  Instead, it can only call the non-blocking
\texttt{try\_lock()} method, which returns immediately if the lock is not
obtainable. ControlIt!'s multi-threaded architecture is thus structured to
only require calls to \texttt{try\_lock()} within the Servo thread. To prevent
race conditions between threads, two copies of the robot model and task state
are maintained: an ``active'' copy that is used by the Servo thread, and an
``inactive'' one that is updated by the non-servo threads.  Updates from the
child threads are provided to the Servo thread by swapping the active and
inactive states. This swapping is done by the Servo thread in a non-blocking
and opportunistic manner.

Figures~\ref{fig:MultiThreadedArch} (a) and (b) show how the Servo thread
passes the latest joint state information to the ModelUpdater thread and
trigger it to execute. After obtaining the latest joint states by calling
RobotInterface.read() and checking for updates from the child threads by
executing the CheckForUpdates FSM, the Servo thread attempts to obtain the
lock on the mutex protecting the inactive robot model by calling
\texttt{ModelUpdater.tryLock()}. If it is able to obtain the lock on the
mutex, it saves the latest joint states in the inactive robot model and then
triggers the ModelUpdater thread to execute by calling
\texttt{ModelUpdater.unlockAndUpdate()}. As the name of this method implies,
the Servo thread also releases the lock on the inactive model thereby allowing
the ModelUpdater thread to access and update the inactive robot model. If the
Servo thread fails to obtain the lock on the inactive model, the ModelUpdater
thread must be busy updating the inactive model. In this situation, the Servo
thread continues without updating the inactive model.

To prevent race conditions between the Servo thread and the child thread,
updates from child threads are opportunistically pulled by the Servo thread.
This is because the child threads operate on inactive versions of the robot
model and task states, and only the Servo thread can swap the active and
inactive versions. There are two points in the servo loop where the Servo
thread obtains updates from the child threads. This is shown by the two
``CheckForUpdates'' states in left side of Figure~\ref{fig:MultiThreadedArch}
(a). They occur immediately after obtaining the latest joint states by calling
RobotInterface.read(), and immediately after triggering the ModelUpdater
thread to run or failing to obtain the lock on the inactive robot model. More
checks for updates could be interspersed throughout the servo loop but we
found these two placements to be sufficient.

The operations of the CheckForUpdates state are shown in the upper-right
corner Figure~\ref{fig:MultiThreadedArch}. The Servo thread first obtains task
state updates and then checks whether the \texttt{TaskUpdater} thread is idle.
If it is idle, the Servo thread again checks for updated task states. This is
to account for the following degenerate thread interleaving during the
previous check for updated task states that would result in the permanent loss
of updated task state:

\begin{enumerate}
\item The Servo thread begins to check some of the tasks for updated states.
\item TaskUpdater thread updates all of the tasks including those that were
just checked by the Servo thread and returns to idle state. Note that this is
possible even if the Servo thread is real-time and has higher priority since
the TaskUpdater may be executing on a different CPU core.
\item The Servo thread completes checking the remainder of the tasks for
updates.
\end{enumerate}

In the above scenario, the tasks that were checked in step 1 would have
updated states that would be lost without the Servo-thread re-checking for
them after it confirms that the \texttt{TaskUpdater} is idle. In a worst-case
scenario, the \texttt{TaskUpdater} thread may update all of the tasks after
the Servo thread checks for updates but before it checks whether the
\texttt{TaskUpdater} is idle, resulting in the loss of updated state from
every task. The loss of updated task state is not acceptable despite the
presence of future update rounds since it is theoretically possible for the
updated states of the same tasks to be continuously lost during every update
round. While seemingly improbable, this ``task update starvation'' problem was
actually observed and thus discovered while testing ControlIt! on Valkyrie.

After verifying that the \texttt{TaskUpdater} thread is idle and ensuring all
of the updated task states were obtained, the Servo thread next checks for an
updated robot model by calling \texttt{ModelUpdater.checkUpdate()}. This
method switches to the updated robot model if one is available. If the model
was updated, the \texttt{Servo} thread then calls
\texttt{TaskUpdater.updateTasks()} passing it the updated robot model. This
method is non-blocking since the TaskUpdater must be idle. It triggers the
\texttt{TaskUpdater} to update the states of each task in the compound task.
Note that if the robot model was not updated, the Servo thread does not call
\texttt{TaskUpdater.updateTasks()} since task state updates are based on
changes in the robot model.

The current implementation does not consider the possibility that the active
robot model or task states become excessively stale. This can occur if the
robot moves so quickly that the model changes significantly since the last
time it was updated. ControlIt's multi-threaded architecture can be easily
modified to monitor difference between the current robot state and the robot
state that was used to update the currently-active robot model and task
states. If the difference exceeds a certain threshold, the Servo loop can
update the active model itself to prevent excessive staleness. We currently do
not implement this because our evaluations did not indicate the need for it.

Sometimes a multi-threaded architecture is not necessary when the robot has a
limited number of joints, the control computer is particularly fast, and the
compound task is structured to reduce computational complexity (e.g., by using
simpler tasks or limiting the number of tasks that share the same priority
level). In this case, ControlIt!'s multi-threaded architecture can be disabled
by setting two ROS parameters, \texttt{single\_threaded\_model} and
\texttt{single\_threaded\_tasks}, to be true prior to starting ControlIt!. Details of these parameters are given in Table~\ref{table:parameters2}, which is in~\ref{sec:parameters}.
When these parameters are set to true, the servo loop updates the model and
task states each cycle of the servo loop.

Regardless of whether a multi-threaded architecture is used, the servo loop
must be executed in a real-time manner. To help facilitate this, no dynamic
memory allocation can occur once the servo loop starts. The initialization
process consists of instantiating all objects using their constructors and
then calling \texttt{init()} methods on all of the objects. All necessary
memory is allocated during either the construction or initialization phases.
To ensure no memory is being dynamically allocated in the linear algebra
operations that are extensively used in WBOSC, we tested the code by defining
the \texttt{EIGEN\_RUNTIME\_NO\_MALLOC} preprocessor macro prior to including
the Eigen headers.

\section{Evaluation}\label{sec:evaluation} We integrate ControlIt! with Dreamer, a dual-arm  humanoid upperbody made by
Meka Robotics, which was purchased by Google in December 2013. Dreamer's arms
and torso contains series elastic actuators and high fidelity torque control.
The robot is modeled as a (16 + 6 = 22) DOF robot where 16 are the physical
joints and the remaining 6 DOFs represent the floating DOFs.\footnote{WBOSC by default always assumes a floating base. In the case when the robot is fixed in place, it is represented in WBOSC as a constraint. This enables ControlIt! to be more generic in terms of supporting both mobile and fixed robots.}

\subsection{Product Disassembly Application} \label{sec:disassembly}

\begin{figure}[tb]
\centering
\includegraphics[width=.9\columnwidth]{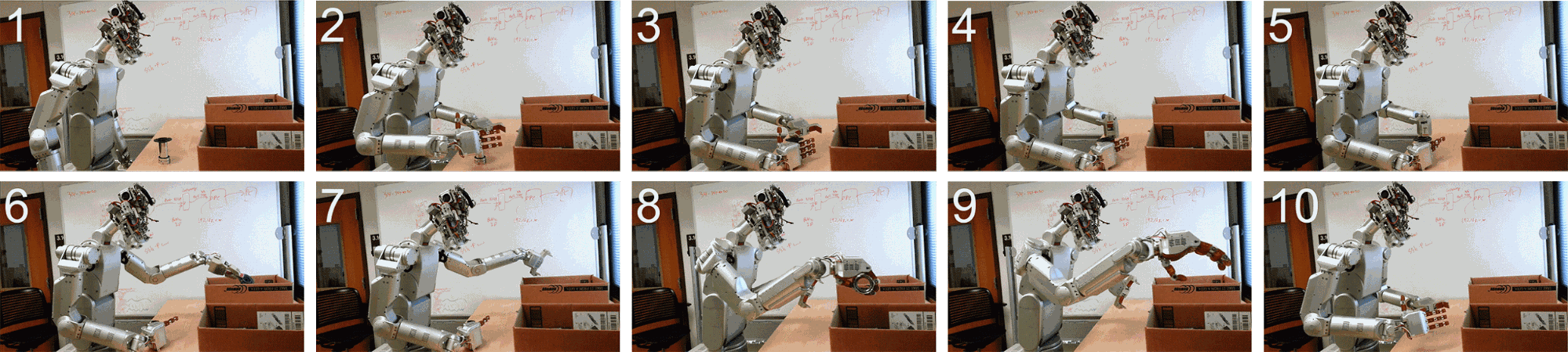}
\vspace{-.1in}
\caption{This sequence of snapshots show the movements of Dreamer performing a
product disassembly task. Initially a metal pipe with a rubber valve is in
front of Dreamer. To disassemble the product, Dreamer  grabs the pipe with her
right hand while using her left gripper to remove the valve. The pipe and
valve are then placed into separate containers for storage. This demonstrates
the integration of ControlIt! with a robot and an application, and the fact
that the task and constraint libraries are sufficiently expressive to
accomplish this task.}
\label{fig:ProductDisassembly}
\end{figure}

Using ControlIt!, we developed an application that makes
Dreamer disassemble a product. A sequence of snapshots showing Dreamer
performing the task using ControlIt! is given in
Figure~\ref{fig:ProductDisassembly}. The task is to take apart an assembly
consisting of a metal pipe with a rubber valve installed at one end. To remove
the valve, Dreamer is programed to grab and hold the metal pipe with her right
hand while using her left gripper to detach the valve. Once separated, Dreamer
places the two pieces into separate storage containers.

Two compound task configurations were used to achieve the product disassembly task:

\begin{enumerate}
\item single priority level containing a joint position task
\item dual priority level containing two higher priority Cartesian position
tasks and two 2D orientation tasks (one for each wrist) and a lower priority
posture task.
\end{enumerate}

The benefits of the second configuration are shown by demonstrating
how changing just three controller parameters, i.e., the Cartesian position of
the product, enables the controller to adapt to changes in the product's
location while continuously minimizing the squared error of the posture task.
This is in the spirit of WBC where changes in a low-dimensional space (three
Cartesian dimensions) results in desirable changes in a larger dimensional
space (e.g., the number of DOFs in the robot).

\begin{figure}[t]
\centering
\includegraphics[width=.9\columnwidth]{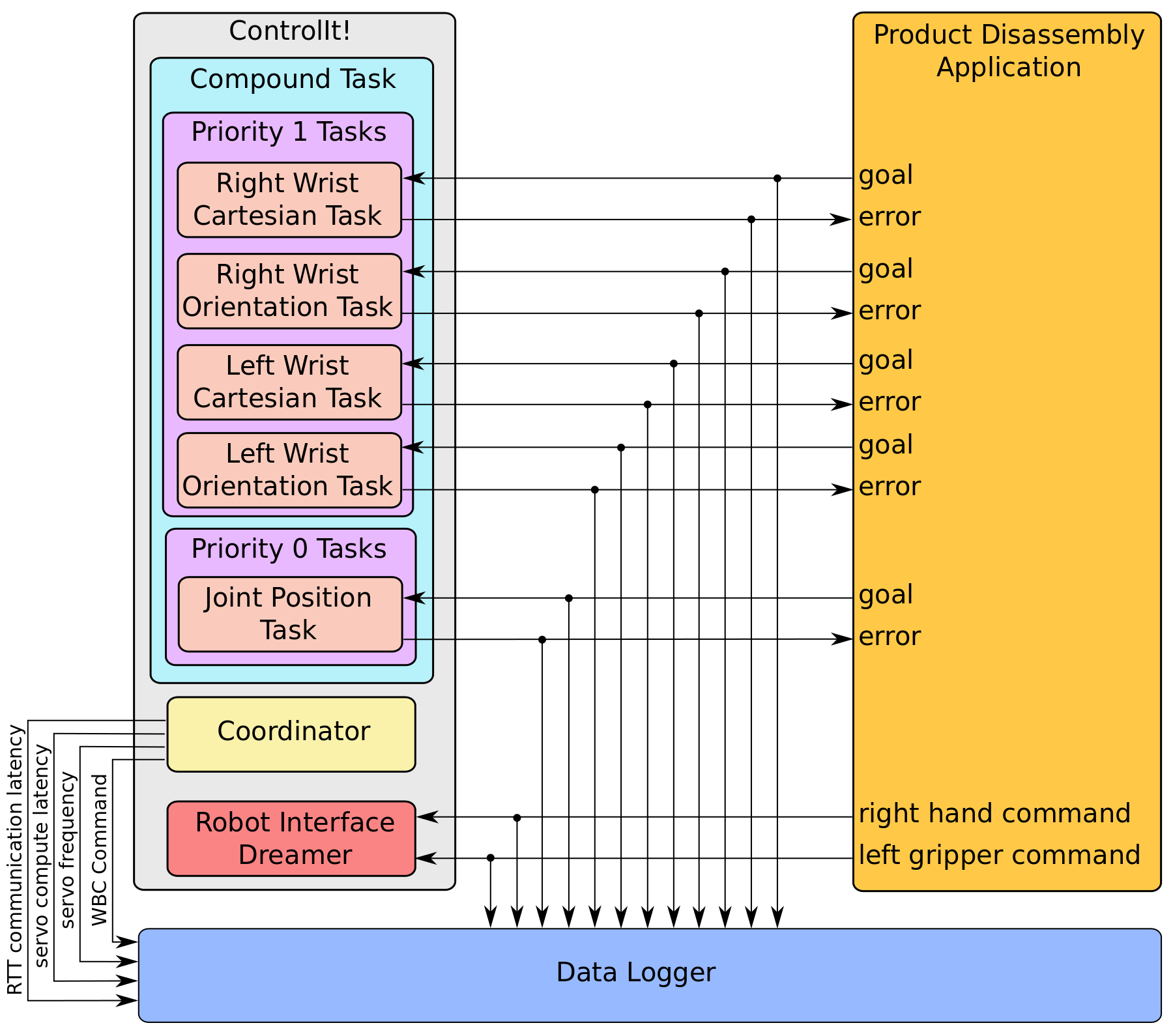}
\vspace{-.1in}
\caption{ControlIt! is integrated into a larger system consisting of three
major components: ControlIt!, the application, and a data logger. Each of
these components run as a separate process but communicate over ROS topics,
which are represented by the arrows. The ROS topics are bound the variables
within ControlIt!. The WBOSC configuration consists of two priority levels
within the compound task is shown. Higher priority numbers correspond to
higher priority tasks. The other components within ControlIt!
are not shown since they do not have any bound parameters in this
application.}
\label{fig:ApplicationArchitecture}
\end{figure}

Developing the product disassembly application required writing new \texttt{RobotInterface} and
\texttt{ServoClock} plugins that enable ControlIt! to work with Dreamer. This
is because Dreamer comes with the M3 software that is designed specifically
for robots built by Meka. The M3 software includes the M3 Server, which
instantiates an RTAI shared memory region through which ControlIt! can
transmit torque commands and receive joint state information. In addition, the
M3 Server also implements the transmissions that translate between joint space
and actuator space and the protocol for setting the modes and gains of the
joint controllers executing on the robot's DSPs. Other useful tools provided
by the M3 software include applications for tuning and calibrating
individual joints. The ControlIt! robot interface we developed for Dreamer is
called \texttt{RobotInterfaceDreamer}.  It uses the shared memory region
created by the M3 Server to connect the WBOSC controller to the robot, and
implements separate simpler controllers for the joints that are not controlled
by WBOSC. These joints include the finger joints in the right hand, the left
gripper joint, the neck joints, and the head joints (eyes and ears). In the
current implementation, these joints are fixed in place from WBOSC's
perspective. While this is not true, they are located at the robot's
extremities and are attached to relatively small masses; the feedback portion
of the WBOSC controller is able to sufficiently account for these inaccuracies
as demonstrated by the successful execution of the application.

Because Dreamer's M3 software is designed to work with RTAI we created an
RTAI-enabled servo clock called \texttt{ServoClockRTAI}, which instantiates a
RTAI real-time thread for executing the servo loop within ControlIt!. Whereas
\texttt{RobotInterfaceDreamer} is specific to Dreamer, \texttt{ServoClockRTAI}
can be re-used on any robot that is RTAI-compatible to get real-time execution
semantics.

\begin{figure}[tb]
\centering

\subfigure[]{\includegraphics[width=.9\columnwidth]{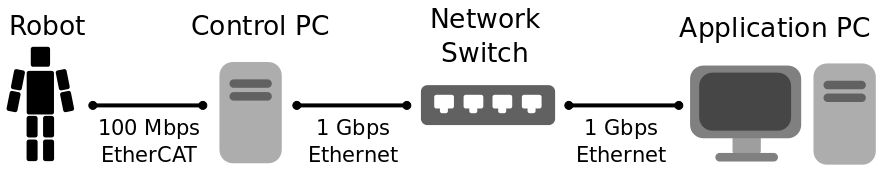}}
\qquad
\subfigure[]{
\footnotesize
\begin{tabulary}{\columnwidth}{L|L|L}
\midrule
\textbf{Property} & \textbf{Control PC} & \textbf{Application PC} \\
\midrule
CPU & Intel Core i7-4771 @ 3.56GHz  & Intel Core i7-4771 @ 3.56GHz\\
\midrule
Motherboard & Zotac H87 & JetWay JNF9J-Q87\\
\midrule
OS & Ubuntu 12.04 server, 32-bit, kernel 2.6.32.20,  RTAI 3.9, EtherCAT 1.5.1 & Ubuntu 14.04 desktop, 64-bit, Kernel 3.13.0-44\\
\midrule
Middleware and Applications & ROS Hydro, ControlIt!, M3 Server & ROS Indigo, demo applications, Gazebo\\
\midrule
\end{tabulary}
}
\vspace{-.1in}
\caption{The system consists of a humanoid robot that's connected to a control
PC over a 100Mbps EtherCAT network. The control PC runs ControlIt! and is
connected to an application PC over a two-hop 1Gbps Ethernet network. The
application PC runs the application, which remotely interacts with ControlIt!
via ROS topics. Details of the hardware and software on the control and
application PCs are given in the table. Note that the control PC runs an older
operating system and older middleware than the application PC despite having
similar hardware. This is because configuring the control PC for real-time
operation while remaining compatible with the robot hardware is difficult.
Allowing applications to run on a separate PC enables them to operate in a
more up-to- date software environment and reduces the likelihood of
interference between the applications and the controller.}
\label{fig:SystemArchitecture}
\end{figure}

Since Dreamer contains a 2-DOF torso and two 7-DOF arms, we use a compound
task containing a Cartesian position and orientation task for each of the two
end effectors, and a lower priority joint position task for defining the
desired posture. The constraint set contains two constraints: a
\texttt{FlatContactConstraint} for fixing the robot's base to the world and a
\texttt{CoactuationConstraint} for the upper torso pitch joint that is mechanically
connected to the lower torso pitch joint by a 1:1 transmission. This results in the
positions and velocities of the two joints to always be the same. The Jacobian of 
the \texttt{CoactuationConstraint} consists of one row and a column for each DOF 
in the robot's model. The column representing the slave joint contains a 1 and 
the column representing the master joint contains the negative of the transmission 
ratio. Details of these types of constraints were discussed in~\cite{Sentis2013}.

Finally, the goal state and error of every task in the compound task are bound
to ROS topics so they can be accessed by the application. A data logger based
on ROSBag~\cite{Website-ROS-Bag} is used to record experimental data.
Figure~\ref{fig:ApplicationArchitecture} shows how the various components are
connected. Kinesthetic teaching is used to obtain the trajectories for
performing the task, which consists of manually moving the robot along the
desired trajectories while taking snapshots of the robot's configuration.
Cubic spline is used to interpolate intermediate points between snapshots.
Note that the application is open-loop in that the robot does not sense where
the metal pipe and valve assembly is located. We manually reposition the metal
pipe and valve assembly at approximately the same location prior to executing
the application.

Before the application can be successfully executed, calibration and gain
tuning must be done for every joint and controller in the system. We
calibrated and tuned one joint at a time starting from those in the robot's
extremities (e.g., wrist yaw joints) and moving inward to joints with
increasing numbers of child joints. Once all of the joints were calibrated and
torque controller gains tuned, we proceeded to tune the task-level gains in
the following order: joint position task, Cartesian position tasks, and
finally orientation tasks. The gains used are given in~\ref{sec:gains}. Note
that these gains are dependent on ControlIt's servo frequency, which we set to
be 1kHz, and the end-to-end communication latency between the whole body
controller and the joint torque controllers, which is about 7ms.

The system architecture is shown in Figure~\ref{fig:SystemArchitecture}. It
consists of the robot, the control PC, and the application PC. The robot
communicates with the control PC over a 100Mbps EtherCAT link. The control PC
communicates with an application PC via a 2-hop 1Gbps Ethernet network. The
control PC runs ControlIt! on an older but real-time patched version of Linux
relative to the application PC. This is because upgrading the operating system
on the control PC while maintaining compatibility with RTAI and necessary
drivers like EtherCAT and ensuring acceptable real-time performance is a
difficult and time consuming process that requires extensive testing. The
product disassembly application could run directly on the Control PC, but we
chose to run in on a different application PC to emphasize the ability to
integrate ControlIt! with remote processes and to allow the application to
make use of a newer operating system, middleware, and libraries. In addition,
running the application on a separate PC reduces the likelihood that the
application would interfere with the whole body controller especially if the
application includes a complex GPU-accelerated GUI.

\begin{figure}[t]
\centering
\includegraphics[width=.9\columnwidth]{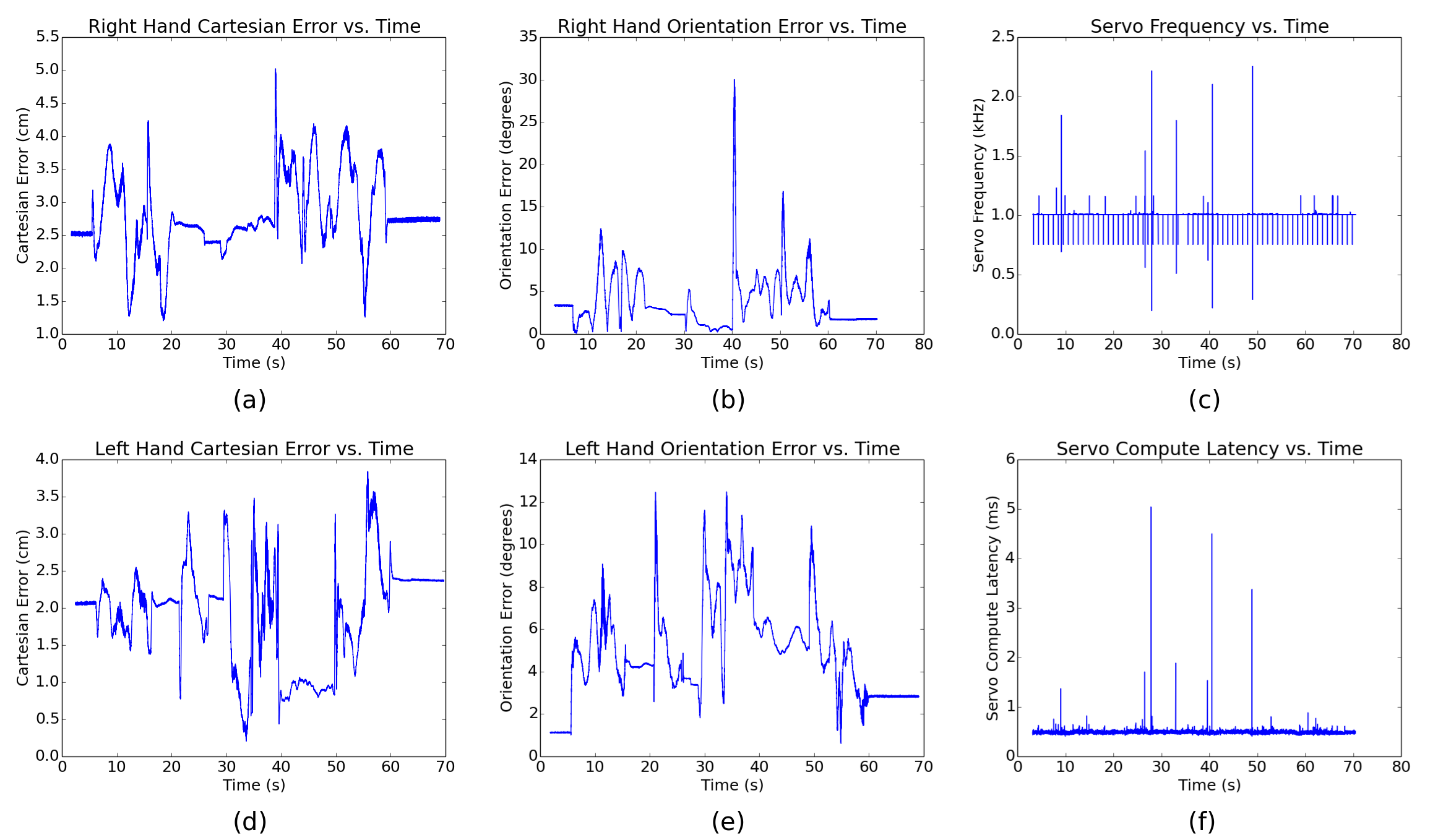}
\vspace{-.1in}
\caption{Performance data collected from one execution of the product
disassembly application.}
\label{fig:AppResults}
\end{figure}

The application PC includes the dynamics simulator
Gazebo~\cite{Website_Gazebo}. When developing the product disassembly
application, we always tested the application in simulation prior to on real-
hardware, reducing the number of potentially-catastrophic problems encountered
on hardware. For example, on the real hardware, if the application crashes
while the arms are above the table, the arms may slam into the table with
enough force to result in damage to the robot and perhaps the table. Testing
the application in simulation enabled us to evaluate application stability. We
implemented the application in Python (see~\ref{sec:application_code} for an
example code fragment), which further increases the importance of simulation
testing since there's no compilation stage to identify potential problems.
Note that the application could have been written in any programming language
supported by ROS~\cite {Website-ROS-Client}. Because ControlIt! has a hardware
abstraction layer consisting of a RobotInterface plugin and a ServoClock
plugin, switching between testing the application in simulation versus on the
real hardware is simple and does not require any changes to the code.

After tuning the controllers, we were able to repeatedly execute the
application in a reliable manner. Figure~\ref{fig:AppResults} shows
performance data collected from one of the many executions of the application.
The data was collected from ROS topics to which internal controller parameters
were bound. Average statistics are given in
Table~\ref{table:AppAverageStatistics}. The results show average servo
computational latencies of about 0.5ms, which is the amount of time the servo
thread takes to compute one cycle of the servo loop and is an order of
magnitude faster than the 5ms achieved by UTA-WBC.
Table~\ref{table:ServoLatencyBreakdown} shows the results of an experiment
that obtains a detailed breakdown of the latencies within the servo loop by
instrumenting the servo loop with timers. The values are the average over 1000
executions of the servo loop. The vast majority of the servo loop's
computational latency is from executing the WBOSC algorithm to get the next
command. Multi-threading significantly decreases the latency of updating the
model and slightly decreases the latency of computing the command. The
slightly higher average total latency in the multi-threaded case in
Table~\ref{table:ServoLatencyBreakdown} relative to the servo computational
latency in Table~\ref{table:AppAverageStatistics} is most likely due to the
additional instrumentation that was addded to the servo loop to obtain the detailed
latency breakdown information.

\begin{table}[t]
\centering
\footnotesize
\begin{tabulary}{\columnwidth}{l|L|L|L}
\midrule
\textbf{Statistic} & \textbf{Sample Size} & \textbf{Average} & \textbf{Units} \\
\midrule
Right Hand Cartesian Error & 49,137 & 2.79 $\pm$ 0.56 & cm\\
\midrule
Right Hand Orientation Error & 55,735 & 3.72 $\pm$ 3.12 & degrees\\
\midrule
Left Hand Cartesian Error & 43,026 & 1.91 $\pm$ 0.67 & cm\\
\midrule
Left Hand Orientation Error & 50,381 & 4.86 $\pm$ 2.23 & degrees\\
\midrule
Servo Frequency & 67,225 & 1005.43 $\pm$ 15.68 & Hz\\
\midrule
Servo Compute Latency & 64,118 & 0.487 $\pm$ 0.0335 & ms\\
\midrule
\end{tabulary}
\caption{Average statistics of the performance data from one execution of the
product disassembly task using the 22-DOF Dreamer model. The average range is the standard deviation of the
data set.  The results indicate that average Cartesian position error of the
end effectors are about 2-3cm and average orientation is about 3-5 degrees.
The servo frequency is slightly above the desired 1kHz and there is jitter
despite running within an RTAI real-time context. The servo compute latency
indicates that on average it only takes about 0.5ms to perform all
computations in one cycle of the servo loop, which is significantly faster
than the 5ms required by UTA-WBC.} \label{table:AppAverageStatistics}
\end{table}

The results in Table~\ref{table:AppAverageStatistics} also show
Cartesian positioning errors of up to 5cm and orientation errors of up to 30
degrees, though the errors are much less on average. Note that the Cartesian
position and orientation errors are both model-based meaning they are derived
from the joint states and the robot model and not from external sensors like a
motion capture system. Thus, the accuracy of these error values depend on the
accuracy of the robot's model and should not be considered absolute. However,
they do represent the errors that the whole body controller sees and attempts
to eliminate but cannot because the feedback gains cannot be made sufficiently
high to remove the errors.

\begin{table}[t]
\centering
\footnotesize
\begin{tabulary}{\columnwidth}{l|L|L}
\midrule
\textbf{Step in Servo Loop} & \textbf{Multi-Threaded Latency} & \textbf{Single-Threaded Latency}\\
\midrule
Read Joint State & 0.020 $\pm$ 0.0020 & 0.020 $\pm$ 0.0026 \\
\midrule
Publish Odometry & 0.014 $\pm$ 0.0041 & 0.0147 $\pm$ 0.00526 \\
\midrule
Update Model & 0.0075 $\pm$ 0.00256 & 0.272 $\pm$ 0.00235 \\
\midrule
Compute Command & 0.470 $\pm$ 0.0128 & 0.497 $\pm$ 0.0120 \\
\midrule
Emit Events & 0.0036 $\pm$ 0.00028 & 0.0041 $\pm$ 0.00027 \\
\midrule
Write & 0.0116 $\pm$ 0.00075 & 0.0125 $\pm$ 0.00119 \\
\midrule
\textbf{Total} & \textbf{0.528 $\pm$ 0.0144} & \textbf{0.820 $\pm$ 0.0145} \\
\midrule
\end{tabulary}
\caption{A breakdown of the latencies incurred within one cycle of the servo
loop for both the single and multi-threaded scenarios using a 22 DOF robot model. All values are in
milliseconds and are the average and standard deviation over one thousand
samples. Most of the latency is spent computing the command, which includes
executing the WBOSC algorithm. The benefits of multi-threading are apparent in
the latency of updating the model.}
\label{table:ServoLatencyBreakdown}
\end{table}

\begin{figure}[tb]
\centering
\subfigure[]{\includegraphics[width=.45\columnwidth]{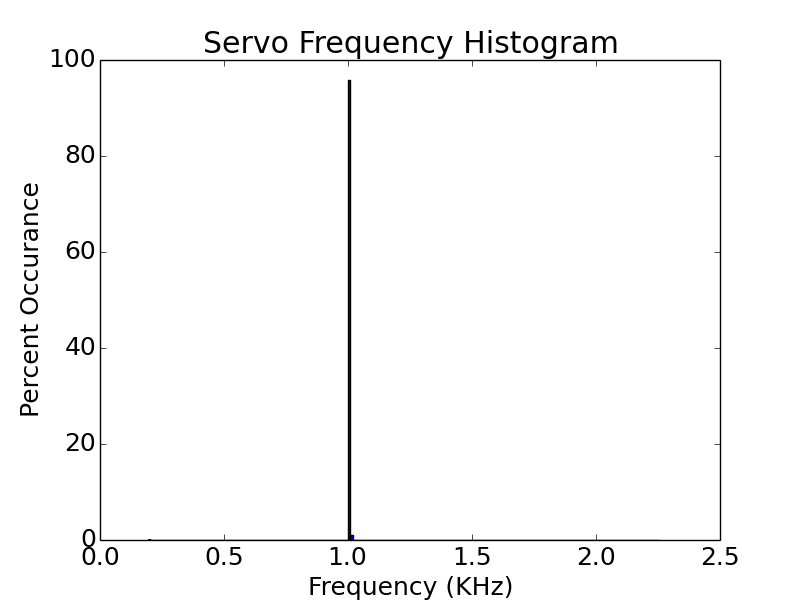}}
\qquad
\subfigure[]{\includegraphics[width=.45\columnwidth]{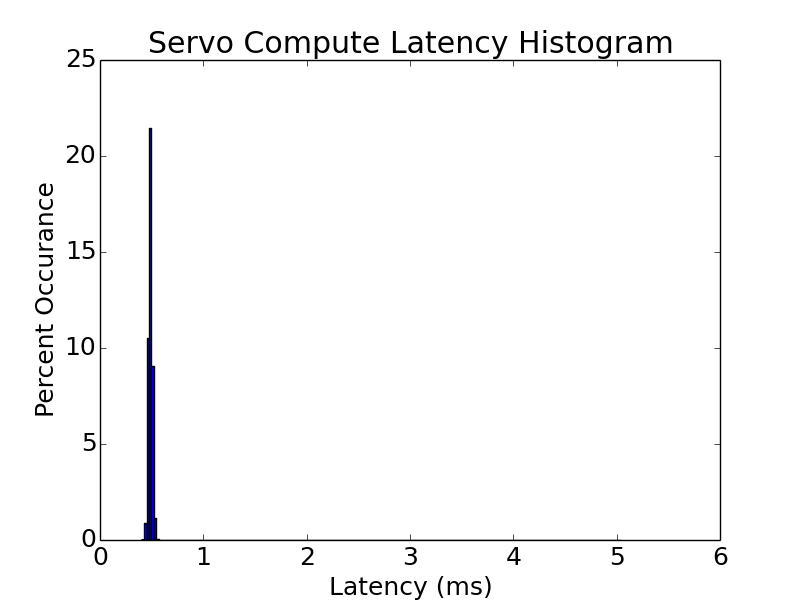}}
\vspace{-.1in}
\caption{Histograms of the servo frequency and computational latency measured
during one execution of the product disassembly application. The vast majority
of the measurements were at the desired 1KHz frequency and expected 0.5ms
computational latency.}
\label{fig:Histograms}
\end{figure}

Figures~\ref{fig:AppResults}(c) and~\ref{fig:AppResults}(f) indicate a problem
with achieving real-time semantics on the control PC since the servo frequency
and computational latency occasionally suffers excessively low and high
spikes. The lowest servo frequency measured in this sample execution is only
195.3Hz, the maximum is 2.254kHz, and the average is 1.01 $\pm$ 0.016kHz.
Coincident with the large spikes in the servo frequency are large spikes in
the servo compute latency. This indicates that something in the operating
system or underlying hardware occasionally prevented ControlIt!'s real-time
servo thread from executing as expected. Despite the violations in real-time
semantics and errors in Cartesian position and orientation, the ControlIt! is
still able to make Dreamer reliably perform the task. This is probably because
the spikes are rare as shown by the histograms of the same data as shown in
Figure~\ref{fig:Histograms}.

\subsection{Latency Benchmarks}

The results in Table~\ref{table:AppAverageStatistics} indicate that the servo
loop spends about 0.487 $\pm$ 0.0335 ms computing the next command. This is
for a specific compound task with two priority levels and 2D orientation tasks
and with multi-threading enabled. We now vary the compound task configuration
in terms of both number of priority levels (which affects the number of tasks
per priority level) and types of orientation task used. We also evaluate both
multi-threaded and single-threaded execution of ControlIt!.

All tests involve five tasks: a Cartesian position task for each of the two
end effectors, an orientation task for each of the two end effectors, and a
posture task. Two types of orientation tasks are used: 2D and 3D. When 2D
orientation tasks are used, only 5 DOFs of each end effector are controlled by
the orientation and position tasks; the sixth DOF is controlled by a lower
priority posture task. When 3D orientation tasks are used, all 6 DOFs of each
end effector are controlled by the orientation and position tasks.

Three configurations of the compound task are evaluated. The first
configuration uses two priority levels and assigns all four Cartesian position
and orientation tasks to be at the higher priority level. The posture task is located at the lower priority level. The second configuration
uses three priority levels and assigns the Cartesian position tasks to be at
the highest priority level and the orientation tasks to be in the middle priority level. This is possible since the orientation tasks operate within the
nullspace of the Cartesian position tasks. Like the first configuration, the posture task is located at the lowest priority level. The third configuration uses 5 priority levels. The two Cartesian position tasks are placed in the top two priority levels. The two orientation tasks are placed in the next two priority levels. Finally, the posture task is located in the lowest priority level.

\begin{table}[t]
\centering
\footnotesize
\begin{tabulary}{\columnwidth}{L|L|L|L}
\midrule
\textbf{Priority Levels / Task Allocation} & \textbf{Orientation Task} & \textbf{Threading} & \textbf{Latency (ms)~~} \\
\midrule
\midrule
2 priority levels  & 2D & multi & 0.528 $\pm$ 0.0144\\
\midrule
~4 tasks at higher priority  &    & single & 0.820 $\pm$ 0.0145\\
\midrule
~1 task at lower priority & 3D & multi & 0.999 $\pm$ 0.0261\\
\midrule
  &    & single & 1.289 $\pm$ 0.0218\\
\midrule
\midrule
3 priority levels & 2D & multi & 0.494 $\pm$ 0.0161\\
\midrule
~2 tasks at highest priority  &    & single & 0.764 $\pm$ 0.0217\\
\midrule
~2 tasks at middle priority  & 3D & multi & 0.788 $\pm$ 0.0212\\
\midrule
~1 task at lowest priority  &    & single & 1.068 $\pm$ 0.0207\\
\midrule
\midrule
5 priority levels & 2D & multi & 0.477 $\pm$ 0.0155\\
\midrule
~1 task at each level  &    & single & 0.744 $\pm$ 0.0386\\
\midrule
 & 3D & multi & 0.603 $\pm$ 0.0166\\
\midrule
 &    & single & 0.882 $\pm$ 0.0168\\
\midrule
\end{tabulary}
\caption{The servo loop's computational latency when configured with several
different compound tasks and running in both multi-threaded and single-threaded mode using a 22-DOF model. All latencies are the average over 1000 consecutive
measurements and the intervals are the standard deviations.
The results show that the servo loop's computational latency can be
significantly decreased using by using multi-threading and placing fewer tasks at each priority level.} \label{table:LatencyVsCompoundTask}
\end{table}

The results are shown in Table~\ref{table:LatencyVsCompoundTask}. The use of
multi-threading significantly decreases computational latency by about 0.2-0.3
ms. Interestingly, distributing the tasks across more priority levels
decreases computational latency. In this case, placing the orientation tasks
and Cartesian position tasks at different priority levels results in a
significant decrease in servo computational latency. This is because the
Jacobians and commands of all tasks within the same priority level are
concatenated into a large matrix and, in this case, performing operations on
large matrices takes more time than performing multiple operations and
nullspace projections using smaller matrices.

Note that ControlIt! can maintain a 1kHz servo frequency in many of the
compound task configurations even when running in single-threaded mode.
Specifically, when 2D orientation tasks are used, 1kHz servo frequencies are
achieved in all compound task configurations. When 3D orientation tasks are
used, 1kHz servo frequencies can be achieved when the five tasks are spread
across five priority levels. The $0.882 \pm 0.0168$ ms that's achieved in this
case is similar to the $0.9 \pm 0.045$ ms that's achieved using an optimized
quadratic programming WBC algorithm~\cite{Herzog2014_IROS}.

\begin{figure}[tbh]
\centering
\includegraphics[width=.9\columnwidth]{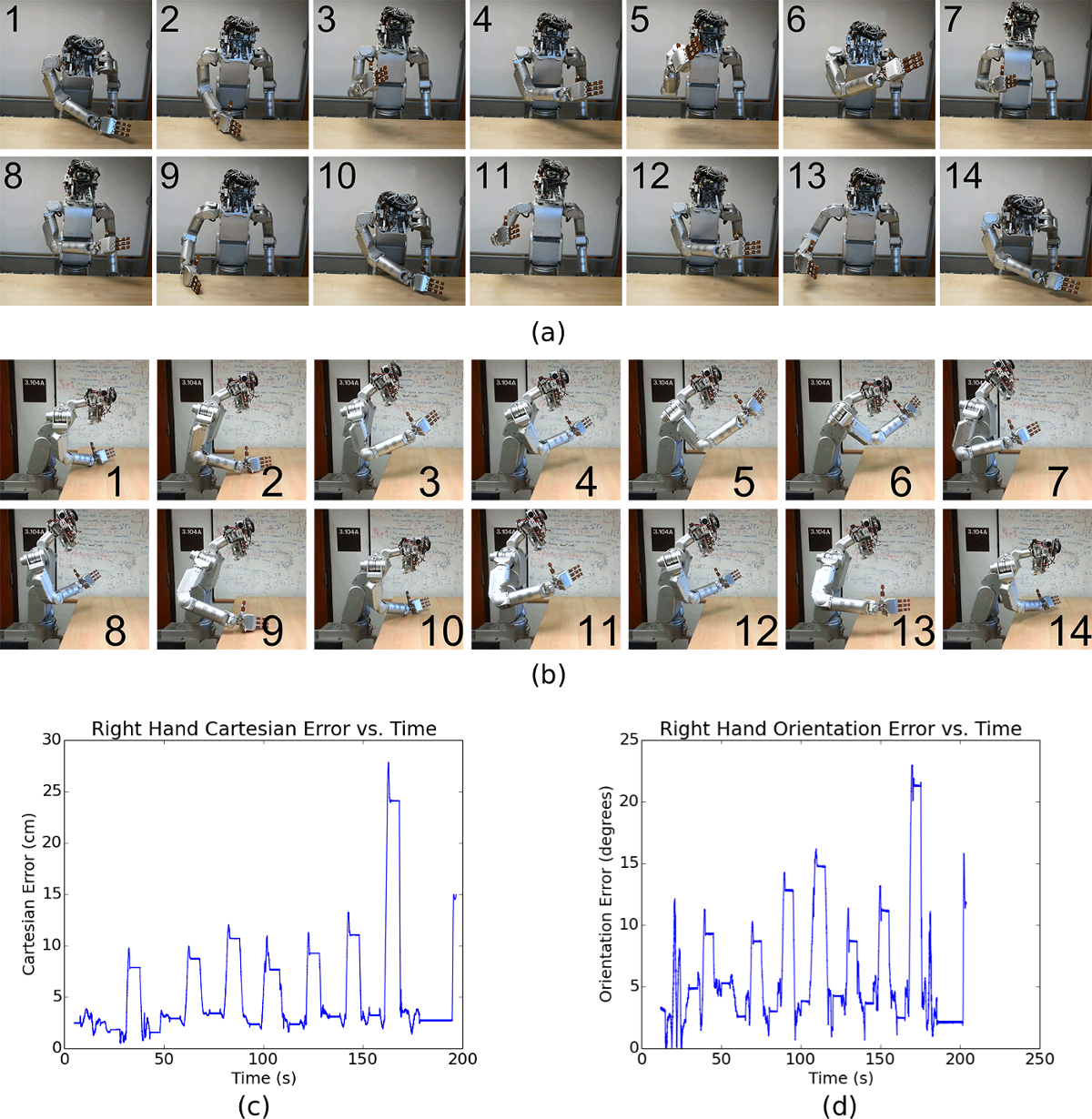}
\vspace{-.1in}
\caption{This figure shows two different perspectives of the same execution of
Dreamer changing the Cartesian position of her right hand while keeping the
lower priority joint position task unchanged. It demonstrates WBOSC's ability
to handle changes in the goal Cartesian position while predictably handling
robot redundancies. The error plots show periodically elevated errors when the
goal Cartesian position is moved beyond the robot's workspace. The errors are
square-shaped because of a 5-second pause inserted between successive
Cartesian trajectories. The controller remains stable despite this problem.}
\label{fig:FlexibleCartesianPositioning}
\end{figure}

\subsection{Flexible End Effector Repositioning} \label{sec:eeRepos}

As previously mentioned, the product disassembly application operates open-
loop and requires the product to be placed at approximately the same location
at the beginning of each execution of the application. For the application to
be more robust, additional sensors need to be integrated that can determine
the actual location of the product and communicate this information to the
application. Such a sensor could be easily integrated since the application is
a ROS node meaning it can simply subscribe to the ROS topic onto which the
sensor publishes the actual location of the product. Once the application
knows where the product is located, it can generate the Cartesian space
trajectories to allow the end effectors to disassemble the product.

To demonstrate the ability for ControlIt! to make Dreamer follow different
Cartesian space trajectories based on a sensed Cartesian goal coordinate, we
created an application that makes Dreamer's right hand move to random
Cartesian positions while keeping the lower priority joint position task
unchanged. The results are shown in
Figure~\ref{fig:FlexibleCartesianPositioning}.  Note that the right hand is
able to move into a wide range of Cartesian positions and that the whole body
of the robot moves to help achieve the goal of the right hand's Cartesian
position task. The elevated error values that periodically appear in Figures
FlexibleCartesianPositioning (c) and (d) are due to the goal Cartesian
position being moved beyond the robot's workspace. Note that despite this
problem the controller remains stable. This demonstrates ControlIt's ability
to be integrated into different applications and WBOSC's ability to handle
robot redundancies in a predictable and reliable manner.

\section{Discussion}\label{sec:discussion} In this section, we provide a brief history of ControlIt's development
followed by future research directions.

\subsection{History of ControlIt!'s Development}

Prior to integration with Dreamer, ControlIt! was initially developed for NASA
JSC's Valkyrie humanoid robot (now called R5)~\cite{Radford2014}. Software and
hardware development commenced simultaneously in October 2012. Since hardware
development took nearly a year, the first year of developing and testing
ControlIt! involved using a simulated version of Valkyrie in
Gazebo~\cite{Website_Gazebo}. During this phase, ControlIt! was initially used
to control individual parts of the robot, e.g., each individual limb, the
lower body, the upper body, and finally the whole robot. By the summer of
2013, ControlIt! was used to control 32-DOFs of Valkyrie in simulation (6 DOFs
per leg, 7 DOFs per arm, 3 DOFs in the waist, and 3 DOFs in the neck).
Compound tasks consisting of up to 15 tasks were employed. They include
Cartesian position and orientation tasks for the wrists, feet, and the head,
an orientation task for the chest, a center of mass task and posture task for
the whole robot, and center of pressure tasks for the feet. Contact
constraints for the hands and feet were configured, though not always enabled,
depending on whether contact with the environment was being made. Management
of all of these tasks and constraints were done using a higher-level
application called Robot Task Commander (RTC)~\cite{Hart2014}, which provided
a graphical user interface for operators to instantiate and configure
controllers based on ControlIt!, integrate these controllers with planners and
other processes via ROS topics (locomotion was done using a phase space
planner~\cite{Kim2015_Arxiv}), and sequence their execution within a finite
state machine. Integration of ControlIt! with Valkyrie in simulation was
successful. We were able to do most of the DRC tasks including valve
turning, door opening, power tool manipulation, ladder and stair climbing,
water hose manipulation, and vehicle ingress. This enabled us to pass the DRC
critical design review in June 2013 and continue to participate in the DRC
Trials as a Track A team.

\begin{figure}[]
\centering
\includegraphics[width=.9\columnwidth]{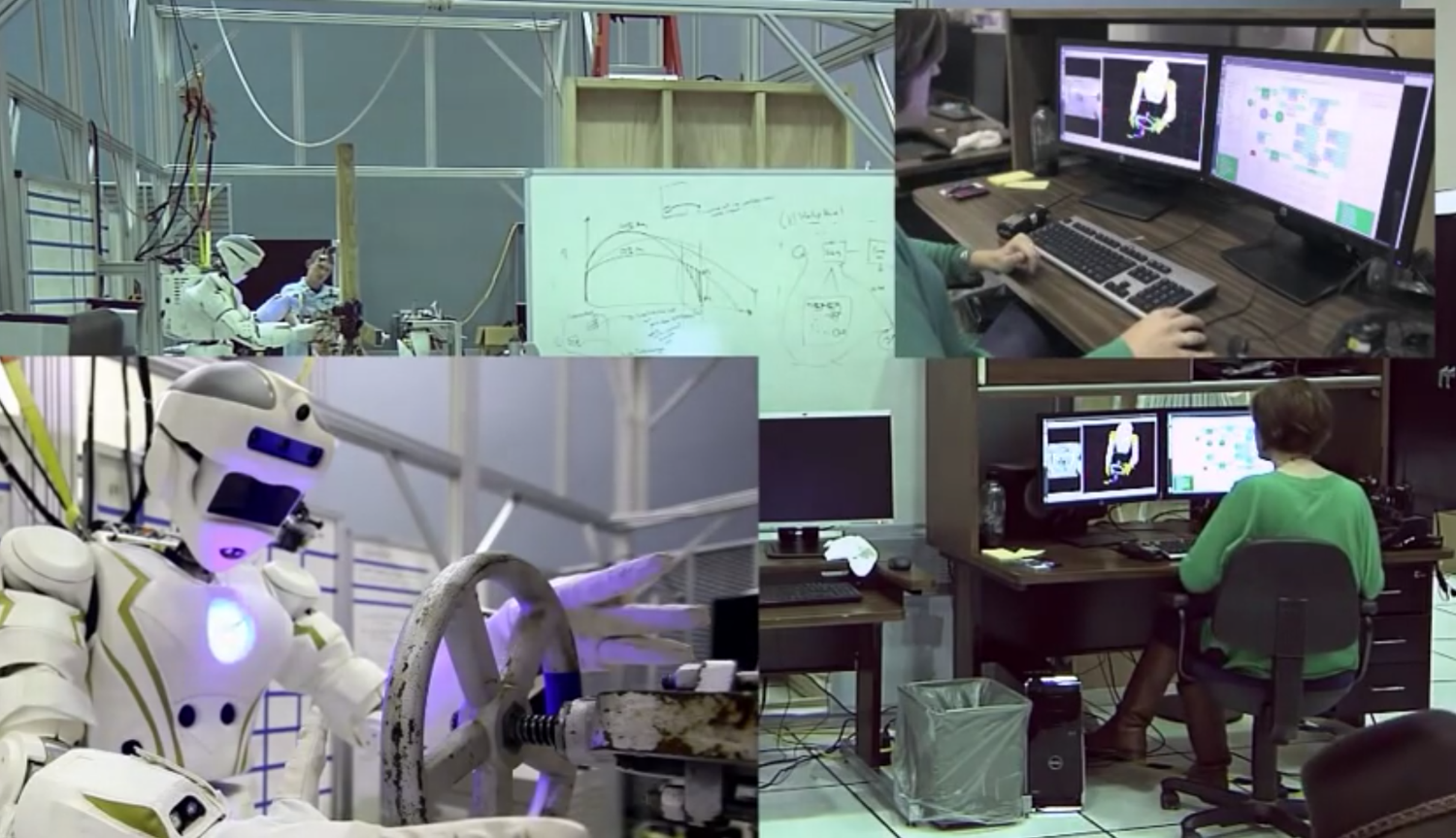}
\vspace{-.1in}
\caption{This figure shows Valkyrie's upper body being controlled by an early
version of ControlIt!. Using a compound task consisting of Cartesian position
and orientation tasks for each hand, and a flat contact constraint for the
torso, a human operator uses Valkyrie to turn an industrial valve.  Parameter
binding is used to integrate ControlIt! with the operator's command and
visualization applications.}
\label{fig:ControlitValkyrieIntegration}
\end{figure}

By the end of Summer 2013, Valkyrie's hardware development was nearing completion. 
At this point we began integrating ControlIt! with actual
Valkyrie hardware. After using ControlIt! to control parts of the robots
individually, we attempted to control all 32 DOFs but ran into problems where
gains could not be increased high enough to sufficiently reduce errors due to
modeling inaccuracies. The robot could stand under joint position control but 
it was not sufficiently stiff to locomote and certain joints like the knees 
and ankles would frequently overheat. We later hypothesized that one problem 
was likely due to the communication latencies between ControlIt! and the joint-level controllers
being too high. We have since developed a strategy called embedded damping to
help maintain stability despite the high communication
latency~\cite{Zhao2015}. Since we could not control all 32 DOFs in time for
the DRC Trials in December 2013, we resorted to use ControlIt! on Valkyrie's
upper body to perform several DARPA Robotics Challenge tasks including opening
a door, using a power tool, manipulating a hose, and turning a valve.
Laboratory tests of ControlIt! being used to make Valkyrie turn a valve and integrated with
the RTC-based operator interface is shown in
Figure~\ref{fig:ControlitValkyrieIntegration}.

It is important to note that the currently-demonstrable capabilities of WBOSC on
real hardware is a subset of the capabilities we've been able to achieve in
simulation. For example, while preparing for the DRC critical design review in
June 2013, we were able to use ControlIt! to make a simulated Valkyrie walk
using a phase-space locomotion planner and a compound task that controls the
center of pressures of the feet, the center of mass location, and the internal
tensions between the feet. We will continue to strive to demonstrate these
capabilities using ControlIt! on real hardware. Recent results showing an
application-specific implementation of WBOSC controlling Hume, a point-foot
biped, and making it walk in two dimensions is promising~\cite{Kim2015_Arxiv}.

\subsection{Future Research Directions}

As an open-source framework that supports whole body controllers, we hope that
ControlIt! will be adopted by the research community and serve as a common
platform for developing, testing, and comparing whole body controllers. As a
standalone system that works in both simulation and on real hardware,
ControlIt! opens numerous avenues of research. For example, ControlIt!
currently allows tasks and constraints to be enabled and disabled and to
change priority levels at run-time. We tested this on hardware by using a
joint position task to get the robot into a ready state and then switching on
higher priority Cartesian position and orientation tasks to perform a
manipulation application. The transition resulted in a discontinuity in the
torque signal going to the robot, which is not a problem for an upper body
manipulation task, but will likely be a problem for legged locomotion.

We are currently considering two ways to enable smooth WBOSC configuration
changes. The first method is to gradually introduce the effects of a new
task configuration. In this option, the task acceleration or force command is
gradually increased to reach its actual value. The second method consists of
projecting the difference between a current compound task's torque command and
the next one in task space and adjusting for the difference in a feed-forward
manner. This feed-forward adjustment can be gradually eliminated to ensure
smooth transition between tasks. We recently used this technique on Hume, a
biped robot, to smoothly transition between contact and non-contact states of
the feet~\cite{Kim2015_Arxiv}.


While ControlIt! is designed to support multiple WBC algorithms
via plugins, we currently only have two WBC plugins and both are
based on WBOSC. Other successful WBC algorithms incorporate quadratic programming~\cite{Escande2014-IJRR,Stephens2011,Herzog2014_IROS,Johnson2015_IHMC}.
Unlike WBOSC that analytically solves the WBC problem, quadratic programming
is an optimization method that more naturally supports inequality constraints.
While quadratic programming is computationally intensive, recent progress on
methods to simplify quadratic programming-based whole body controllers have
enabled them to execute in less than 1ms on robots with two fewer 
joints than Dreamer~\cite{Herzog2014_IROS}. As future work, it would be
interesting to determine (1) whether quadratic programming-based whole body
controllers could be implemented as a plugin within ControlIt!'s architecture
and (2) the pros and cons of WBOSC relative to quadratic programming-based
whole body controllers. Note that others have developed formulations similar
to WBOSC that include support for inequality constraints and solve them using
quadratic programming~\cite{Salini2013_Thesis}. The integration of on-line
optimization techniques to allow the incorporation of inequality constraints
is an area of future work and may require modifying the current constraint API
to include a specification of whether the constraint is negative or positive.

To the best of our knowledge, there are no other multi-threaded open source
implementations of WBOSC or other forms of whole body controllers. We are
currently unable to prove that our multi-threaded design consisting of a real-time
servo thread with two child threads is optimal. Other choices certainly
exist. For example, the two child threads could be combined into a single
child thread that updates both the model and the tasks. Going in the opposite
direction, a separate child thread could be instantiated for each task where
there is one thread per task. Performing a more detailed analysis on the ideal
multi-threaded architecture is a future research direction.

One consequence of adopting a multi-threaded strategy is the robot model is no
longer updated synchronously with the servo thread and thus can become stale.
We currently do not use any metric to determine when the model has become
excessively stale. A child thread simply updates the model as quickly as
possible. For our product disassembly task, the child thread was able to
update the model fast enough to enable WBOSC to reliably complete the task. An
interesting research direction is to formally investigate how stale a model
can be before it negatively impacts robot performance. The answer will likely
depend on the robot's current configuration.

A given constraint can have an infinite number of null space projectors. The
one we use in ControlIt! is the Dynamically Consistent Null Space
Projector~\cite{Khatib1987}. The nullspace projector is currently derived within the
constraint set. Given the existence of alternative null space projectors, a
potential improvement to ControlIt! would be to make the constraint set
extensible via plugins. The default plugin will use the current Dynamically
Consistent Null Space Projector. However, the user can easily override this by
providing a constraint set plugin that provides another null space
projector.

The results in Section~\ref{sec:disassembly} show that the control PC occasionally has latency
spikes that violate the desired servo frequency. Learning why the latency
spikes occur is useful since eliminating them will likely increase system
reliability or at least predictability. However, we have yet to notice the
latency spikes causing any problem during our extensive use of Dreamer. It's
worth noting that Dreamer is a COTS robot and its control PC was configured by
the robot's manufacturer. Given that the control PC was pre-configured for
us, from our perspective, it is somewhat of a ``black box''. If the need
arises (i.e., the latency spikes actually prevent us from executing a
particular task), we will investigate the latency spikes using a two-pronged
approach. First, we will instrument the Linux kernel with debug messages that
help track down when the latency spikes occur. Second, we will remove all
unnecessary kernel modules and disable all unnecessary hardware until the
latency spikes no longer occur. We will then slowly add hardware and software
modules re-testing for latency spikes after each addition. Once the latency
spikes return, we know which hardware or software module caused it and can
investigate it further.


In this paper, we did not explicitly account for singularities but they did
not pose a problem in our tests even when the arms are fully stretched out as
described in Section 5.3. This is probably due to our choice of the tolerances
for computing pseudo-inverses within the controller. However, we have not
performed a detailed study on adequate tolerances nor on handling singularity
thus far.

Other future research areas include how to add adaptive control capabilities
that continuously improve the robot model based on observed robot behavior,
which should enable the resulting WBOSC commands to have an increasingly high
feed-forward component and lower feedback component, and the integration of
ControlIt! with external sensors to enable, for example, visual servoing.


\section{Conclusions}\label{sec:conclusions} With the increasing availability of sophisticated multi-branched highly-redundant robots targeted for general applications, whole body controllers will likely become an essential component in advanced human-centered robotics. ControlIt! is an open-source software framework that defines a software architecture and set of APIs for instantiating and configuring whole body controllers, integrating them into larger systems and different robot platforms, and enabling high performance via multi-threading. While it is currently focused on facilitating the integration of controllers based on WBOSC, the software architecture is highly extensible to support additional WBC algorithms and control primitives. 

This paper provided a software framework that enables the quick instantiation and configuration of WBOSC behaviors for practical applications such as a product disassembly task using a 22-DOF humanoid upperbody robot. The experiments demonstrated high performance with servo computational latencies of about 0.5ms.

In summary, WBC is a rich and vibrant though fragmented research area today with numerous algorithms and implementations that are not cross-compatible and thus difficult to compare in hardware. We present ControlIt! as a software framework for supporting the development and study of whole body operational space controllers and their integration into useful advanced robotic applications.

\section*{Acknowledgements}

We would like to thank the entire 2013 NASA JSC DRC team for helping with the integration, testing, and usage of ControlIt! on Valkyrie. This work is funded in part by the NSF NRI, Texas Emerging Technology Fund, and an anonymous donor. We would also like to thank Nicholas Paine for helping with creating figures in the evaluation section.

\appendix

\section{ControlIt! Dependencies} \label{sec:dependencies}


\begin{table}[htb]
\centering
\begin{tabulary}{\columnwidth}{l|L|L}
\midrule
\textbf{Dependency} & \textbf{Version} & \textbf{Purpose} \\
\midrule
g++ & 4.8.2 or 4.6.3 & Compiler for C++11 programming language\\
\midrule
Eigen & 3.0.5 & Linear algebra operations\\
\midrule
RBDL & 2.3.2 &Robot modeling, forward and inverse kinematics and dynamics \\
\midrule
URDF & 1.11.6 & Parsing robot model descriptions \\
\midrule
ROS & Hydro or Indigo & Component-based software architecture, useful libraries like pluginlib, runtime support like a parameter server and roslaunch bootstrapping capabilities \\ 
\midrule
RTAI & 3.9 & Real-time execution semantics (only required when using Dreamer or other RTAI-compatible robot) \\
\midrule
Gazebo & 5.1.0 & Test controller in simulation prior to on real hardware \\ 
\midrule
\end{tabulary}
\caption{ControlIt! dependencies.} \label{table:dependencies}
\end{table}

\section{ControlIt! Parameters} \label{sec:parameters}

Tables~\ref{table:parameters1}-\ref{table:parameters2} contains additional ControlIt! parameters that can be loaded onto the ROS parameter server. They must be namespaced by the controller's name.

\begin{table}[htb]
\centering
\small
\begin{tabulary}{\columnwidth}{l|L}
\midrule
\textbf{Name} & \textbf{Description} \\
\midrule
coupled\_joint\_groups & Specifies which groups of joints should be coupled. Effectively modifies the model to decouple group of joints from each other. This is useful for debugging purposes or to account for modeling inaccuracies. It is an array of array of strings.\\
\midrule
enforce\_effort\_limits & Whether to enforce joint effort limits. These limits are specified in the robot description. If true, effort commands exceeding the limits will be truncated at the limit and a warning message will be produced. It is an array of Boolean values.\\
\midrule
enforce\_position\_limits & Whether to enforce joint position limits. These limits are specified in the robot description. If true, position commands exceeding the limits will be truncated at the limit and a warning message will be produced. It is an array of Boolean values.\\
\midrule
enforce\_velocity\_limits & Whether to enforce joint velocity limits. These limits are specified in the robot description. If true, velocity commands exceeding the limits will be truncated at the limit and a warning message will be produced. It is an array of Boolean values.\\
\midrule
gravity\_compensation\_mask & Specifies which joints should not be gravity compensated. This is useful when certain joints have so much friction that gravity compensation is not necessary. It is an array of joint name strings.\\
\midrule
log\_level & The log level, which can be DEBUG, INFO, WARN, ERROR, or FATAL. This controls how much log information is generated during run-time. It is a string value.\\
\midrule
\end{tabulary}
\caption{ControlIt! parameters (1 of 2).} \label{table:parameters1}
\end{table}

\begin{table}[htb]
\centering
\small
\begin{tabulary}{\columnwidth}{l|L}
\midrule	
\textbf{Name} & \textbf{Description} \\
\midrule
log\_fields & Specifies the optional fields that are in a log message's prefix. Possible values include:\\
 & package - the ROS package containing the message\\
 & file - file containing the message\\
 & line - the line number of the message.\\
 & function - the method producing the message\\
 & pid - the process ID of the thread producing the message\\
 & It is an arry of strings.\\
\midrule
max\_effort\_command & Specifies the maximum effort that should be commanded for each joint. A warning is produced if this is violated. It is an arry of integers.\\
\midrule
parameter\_binding\_factories & The names of the plugins containing the parameter binding factories to use. It is an array of strings.\\
\midrule
robot\_description & Contains the URDF description of the robot. This is used to initialize ControlIt's floating model. It is a string value.\\
\midrule
robot\_interface\_type & The name of the robot interface plugin to use. It is a string.\\
\midrule
servo\_clock\_type & The name of the servo clock plugin to use. It is a string value.\\
\midrule
servo\_frequency & The desired servo loop frequency in Hz. Warnings will be published if this frequency is not achieved. It is an integer value.\\
\midrule
single\_threaded\_model & Whether to use the servo thread to update the model. It is a Boolean value.\\
\midrule
single\_threaded\_tasks & Whether to use the servo thread to update the task states. It is a Boolean value.\\
\midrule
whole\_body\_controller\_type & The name of the WBC plugin to use. It is a string value.\\
\midrule
world\_gravity & Specifies the gravity acceleration along the X, Y, and Z axis of the world frame. Defaults to $\langle 0, 0, -9.81 \rangle$. This is useful for debugging or when working in worlds where the gravity does not pull in negative Z axis direction. It is an integer array.\\
\midrule
\end{tabulary}
\caption{ControlIt! parameters (2 of 2).} \label{table:parameters2}
\end{table}

\section{ControlIt! Introspection Capabilities} \label{sec:introspection}

This appendix describes ControlIt!'s introspection capabilities, which enable
users to gain insight into the internal states of the controller.

\begin{figure}[b]
\centering
\includegraphics[width=.9\columnwidth]{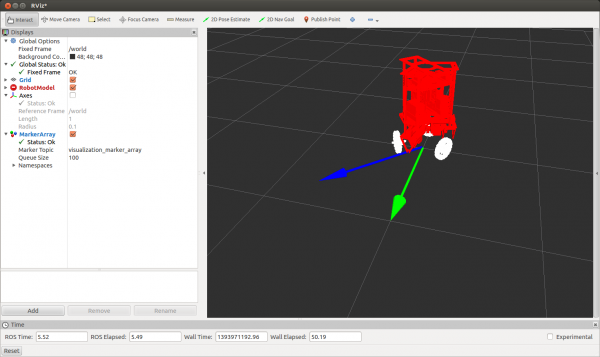}
\vspace{-.1in}
\caption{When integrated with Trikey, ControlIt! can be configured to publish
ROS \texttt{visualization\_msgs/MarkerArray} messages containing the current
and goal headings of the robot. These marker messages can be visualized in
RViz. The green arrow is the goal heading, whereas the blue arrow is the
current heading. In this screenshot, ControlIt! is in the process of rotating
Trikey counter clockwise when viewed from above.}
\label{fig:TrikeyHeading}
\end{figure}

\begin{figure}[tbh]
\centering
\subfigure[]{\includegraphics[width=.7\columnwidth]{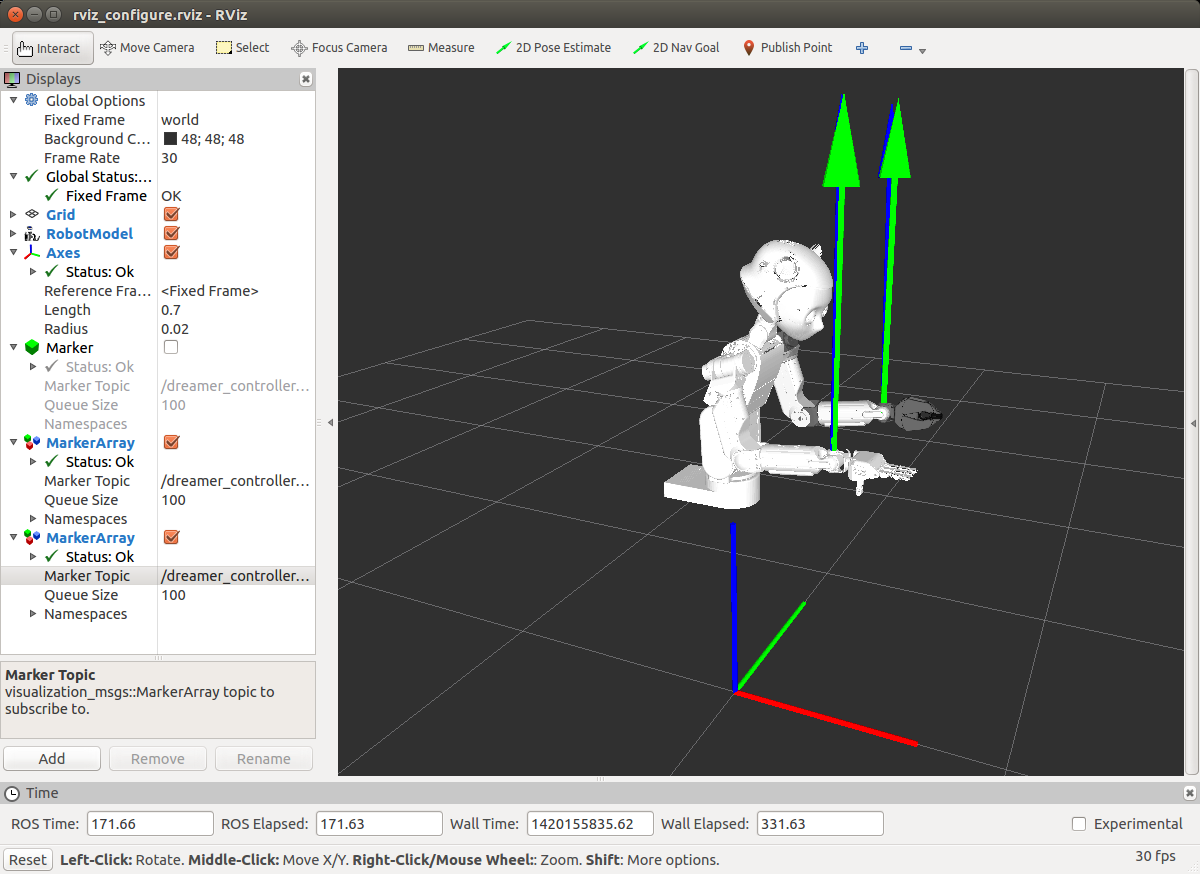}}
\qquad
\subfigure[]{\includegraphics[width=.7\columnwidth]{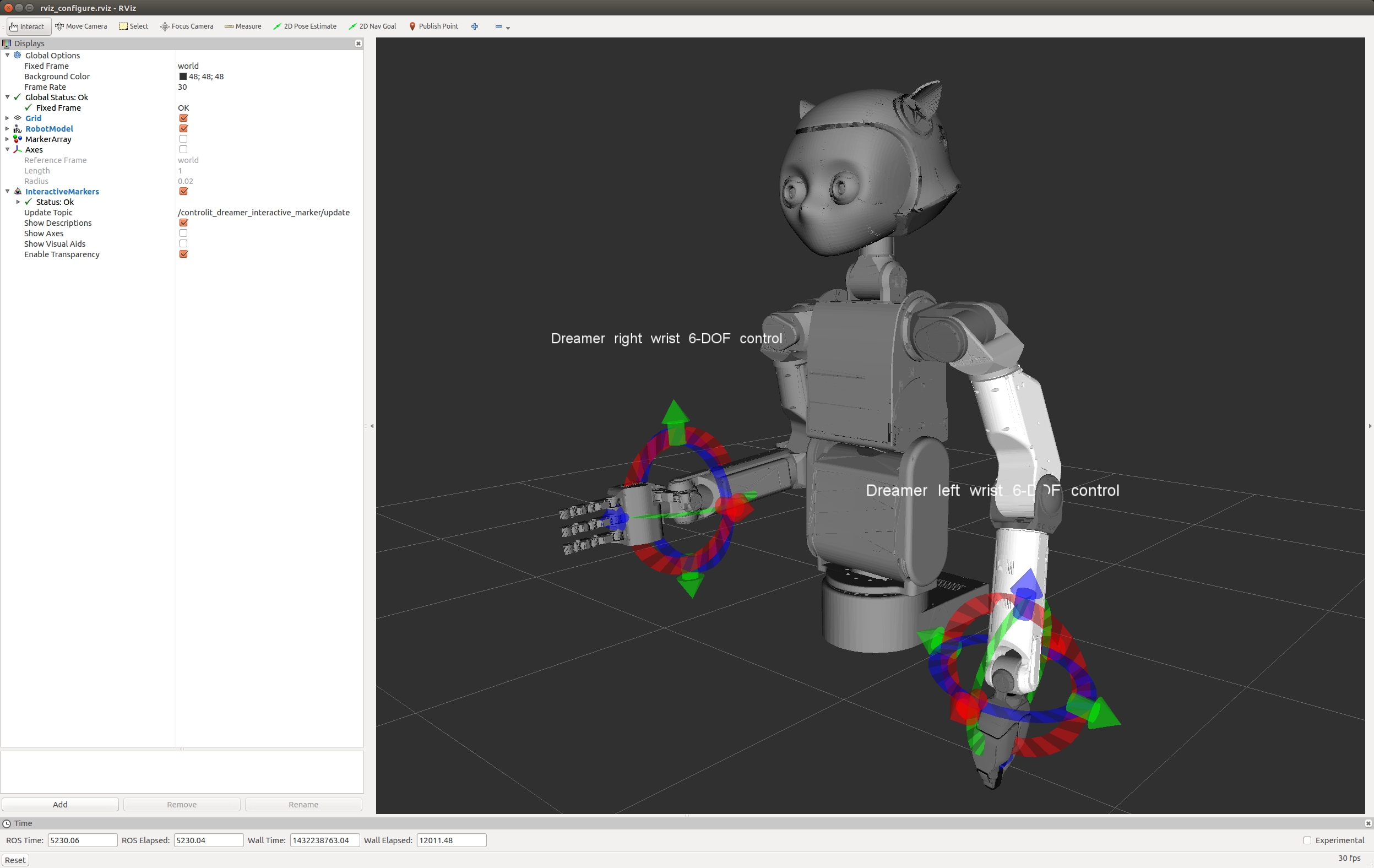}}
\vspace{-.1in}
\caption{Two Cartesian position tasks and two orientation tasks are used to
position and orient Dreamer's end effectors in the world. The orientation and
Cartesian position tasks are higher priority than a joint position task that
defines the robot's posture. (a) Shows the current and goal 2DOF orentations.
(b) Shows how ROS 6-DOF interactive markers denote the current position and
orientation of the wrists. The interactive markers can be dynamically and
visually changed by the user to update the goal positions and orientaions of
the robot's wrists.}
\label{fig:DreamerEndEffectorControl}
\end{figure}

\textbf{Task-based introspection capabilities.} Tasks can configured to
publish ROS \texttt{visualization\_msgs/MarkerArray} and
\texttt{visualization\_msgs/InteractiveMarkerUpdate} messages onto ROS topics
that show the current and goal states of the controller. These messages can be
visualized in RViz to understand what the task-level controller is trying to
achieve. For example, Figure~\ref{fig:TrikeyHeading} shows the marker array
messages published by a 2D orientation task. The green arrow shows the goal
heading whereas the blue arrow shows the current heading.
Figure~\ref{fig:DreamerEndEffectorControl} shows visualizations of 2D and 3D
orientation tasks and Cartesian position tasks.

\begin{figure}[]
\centering
\includegraphics[width=.9\columnwidth]{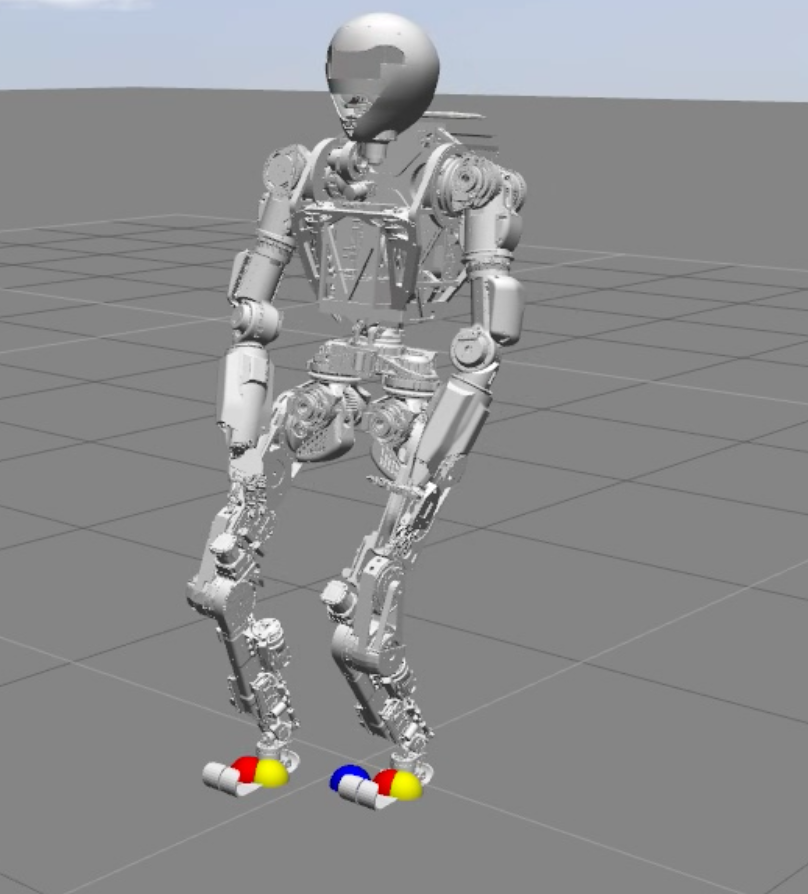}
\vspace{-.1in}
\caption{This is a screenshot from a Gazebo simulation where ControlIt! was
used to make an early prototype of Valkyrie walk six steps. ControlIt's
compound task consisted of a Center-Of-Mass (COM) task, posture task,
Cartesian position task for the hip height, prototype internal tensions task,
and, for each foot, a Cartesian position task, orientation task, and Center-
of-Pressure (COP) task. The red balls mark the goal COP locations of the feet,
yellow balls are the current COP locations, and the blue ball is the COM
projected onto the ground. Note that in this screenshot the yellow balls,
which represent the actual COPs, are on the left edge of the feet, meaning the
feet are very close to rolling clockwise when viewed from the front.}
\label{fig:ValkyrieWalkingSim}
\end{figure}

Figure~\ref{fig:ValkyrieWalkingSim} shows visualizations of the actual and
desired center of pressures and the current center of mass projected onto the
ground. This information is useful to visually determine the stability of the
current posture.

\textbf{ROS service-based introspection capabilities.}
Table~\ref{table:introspectionServices} lists the various service-based
controller introspection capabilities that are provided by ControlIt!. These
services can be called by external processes and are useful for integrating
ControlIt! into a larger system. All services are namespaced by the
controller's name enabling multiple instances of ControlIt! to simultaneously
exist.

\begin{table}[htb]
\centering
\begin{tabulary}{\columnwidth}{l|L}
\midrule
\textbf{Service} &  \textbf{Description} \\
\midrule
diagnostics/getActuableJointIndices & Provides the order of every actuable joint in the robot model (omits joints that are real but not actuable)\\
\midrule
diagnostics/getCmdJointIndices & Provides the order of the joints in the command issued by ControlIt! to the robot.\\
\midrule
diagnostics/getConstraintJacobianMatrices & Provides the Jacobian matrices belonging to the constraints in the constraint set.\\
\midrule
diagnostics/getConstraintParameters & Provides a list of every constraint parameter and its current value.\\
\midrule
diagnostics/getControlItParameters & Provides the current values of the ControlIt! parameters defined in Appendix A.2.\\
\midrule
diagnostics/getControllerConfiguration & Provides the current state of the compound task and constraint set.\\
\midrule
diagnostics/getRealJointIndices & Provides the order of every real joint in the robot model.\\
\midrule
diagnostics/getTaskParameters & Provides a list of every task parameter is its current value.\\
\midrule
\end{tabulary}
\caption{ControlIt!'s ROS service-based controller introspection capabilities.} \label{table:introspectionServices}
\end{table}

\textbf{ROS topic-based introspection capabilities.}
Table~\ref{table:introspectionTopics} lists the various topic-based controller
introspection capabilities that are provided by ControlIt!. These topics can
be subscribed to by external processes and are useful for integrating
ControlIt! into a larger system. All topics are namespaced by the controller's
name enabling multiple instances of ControlIt! to simultaneously exist.

\begin{table}[htb]
\centering
\begin{tabulary}{\columnwidth}{l|L}
\midrule
\textbf{Service} &  \textbf{Description} \\
\midrule
diagnostics/RTTCommLatency & Publishes the latest round-trip communication time between ControlIt! and the joint-level controllers. This is done by transmitting sequence numbers to the joint-level controllers, which are reflected back through the joint state data. ControlIt! monitors the time between transmitting a particular sequence number and receiving it back.\\
\midrule
diagnostics/command & Publishes the latest command issued by ControlIt! to the robot.\\
\midrule
diagnostics/errors & Publishes any run-time errors that are encountered. An example error is when the command includes NaN values.\\
\midrule
diagnostics/gravityVector & Publishes the current gravity compensation vector. \\
\midrule
diagnostics/jointState & Publishes the latest joint state information.\\
\midrule
diagnostics/modelLatency & Publishes the “staleness” of the currently active model. The model latency is the time since the model was last updated.\\
\midrule
diagnostics/servoComputeLatency & Publishes the amount of time it took to execute the computations within one cycle of the servo loop.\\
\midrule
diagnostics/servoFrequency & Publishes the instantaneous servo frequency.\\
\midrule
diagnostics/warnings & Publishes any run-time warnings that are encountered. An example warning is when the joint position or velocity exceeds expected limits.\\
\midrule
\end{tabulary}
\caption{ControlIt!'s ROS topic-based controller introspection capabilities.} \label{table:introspectionTopics}
\end{table}

\section{ControlIt! Configuration File} \label{sec:configuration}

ControlIt! enables user to specify the controller configuration using a YAML configuration file. The syntax of this file is shown below. By enabling YAML-based configuration, ControlIt! can be made to work with a wide variety of applications without modifying the source code and recompiling. \\

\noindent \textbf{Task specification:}

\begin{verbatim}
tasks:
  - name: [task name]  # user defined
    type: [task type]  # must match plugin name
    ... # task-specific parameters and their values
  ... # additional tasks
\end{verbatim}

\noindent \textbf{Constraint specification:}

\begin{verbatim}
constraints:
  - name: [constraint name]  # user defined
    type: [constraint type]  # must match plugin name
    ... # constraint-specific parameters and their values
  ... # additional constraints
\end{verbatim}

\noindent \textbf{Compound task specification:}

\begin{verbatim}
compound_task:
  - name: [task name]
    priority: [priority level]
    operational_state: [enable or disable]
  ... # additional tasks
\end{verbatim}

\noindent \textbf{Constraint set specification:}

\begin{verbatim}
constraint_set:
  - name: [constraint name]
    type: [constraint type]
    operational_state: [enable or disable]
  ... # additional constraints
\end{verbatim}

\noindent \textbf{Binding Specification:}

\begin{verbatim}
bindings:
  - parameter: [parameter name] # must match real parameter name
    direction: [input or output]
    topic: [topic name]
    transport_type: [transport type] # must match plugin name
    properties:
     - [transport-specific property]
     ... # additional transport-specific properties
  ... # additional bindings
\end{verbatim}

\noindent \textbf{Event Specification:}

\begin{verbatim}
 events:
   - name: [event name] # user defined
     expression: [logical expression over parameters]
   ... # additional events
\end{verbatim}

\section{Controller Gains} \label{sec:gains}

The following tables provide the gains used by the various controllers in the
product disassembly application using Dreamer. The negative joint position
controller gains are strange but were configured as such by Meka Robotics, the
robot's manufacturer (Meka Robotics has since been bought by Google). We don't
know for sure why some gains are negative since we are unable to access the
details of the joint-level controllers. It's possible that the direction of
the encoder is opposite of the motor resulting in the need for negative gains.
Regardless, these were the functioning settings used in the development and
testing of ControlIt! on Dreamer.

The reason why the left and right arms have different gains is because the left
arm is about three years newer than the right arm and internally the
mechatronics of the left arm are significantly different from that of the
right arm.

\begin{table}[htb]
\centering
\begin{tabulary}{\columnwidth}{l|L|L|L}
\midrule
\textbf{Controller} &  \textbf{Kp} &  \textbf{Ki} &  \textbf{Kd} \\
\midrule
torso\_lower\_pitch & -3 & 0 & 0\\
\midrule
left\_shoulder\_extensor & 10 & 1 & 0\\
\midrule
left\_shoulder\_abductor & 10 & 1 & 0\\
\midrule
left\_shoulder\_rotator & 10 & 1 & 0 \\
\midrule
left\_elbow & 10 & 1 & 0 \\
\midrule
left\_wrist\_rotator & 50 & 0 & 0 \\
\midrule
left\_wrist\_pitch & 15 & 0 & 1 \\
\midrule
left\_wrist\_yaw & 15 & 0 & 1 \\
\midrule
right\_shoulder\_extensor & 7 & 0 & 0 \\
\midrule
right\_shoulder\_abductor & 6 & 0 & 0 \\
\midrule
right\_shoulder\_rotator & 5 & 0 & 0 \\
\midrule
right\_elbow & 5 & 0 & 0 \\
\midrule
right\_wrist\_rotator & -3 & 0 & 1 \\
\midrule
right\_wrist\_pitch & -15 & 0 & -1 \\
\midrule
right\_wrist\_yaw & -15 & 0 & -1 \\
\midrule
\end{tabulary}
\caption{Dreamer joint torque controller gains.} \label{table:jointTorqueControllerGains}
\end{table}

\begin{table}[htb]
\centering
\begin{tabulary}{\columnwidth}{l|L|L|L}
\midrule
\textbf{Task} &  \textbf{Kp} &  \textbf{Ki} &  \textbf{Kd} \\
\midrule
Joint Position Task & 60 & 0 & 3 \\
\midrule 
Left Hand Orientation & 60 & 0 & 3 \\
\midrule 
Right Hand Orientation & 60 & 0 & 3 \\
\midrule 
Left Hand Position & 64 & 0 & 3 \\
\midrule 
Right Hand Position & 64 & 0 & 3 \\
\midrule 
\midrule
\end{tabulary}
\caption{ControlIt! Task-level controller gains used to control Dreamer.} \label{table:taskGains}
\end{table}

\section{Example Application Code} \label{sec:application_code}

\begin{figure}[tbh]
\centering
\includegraphics[width=.9\columnwidth]{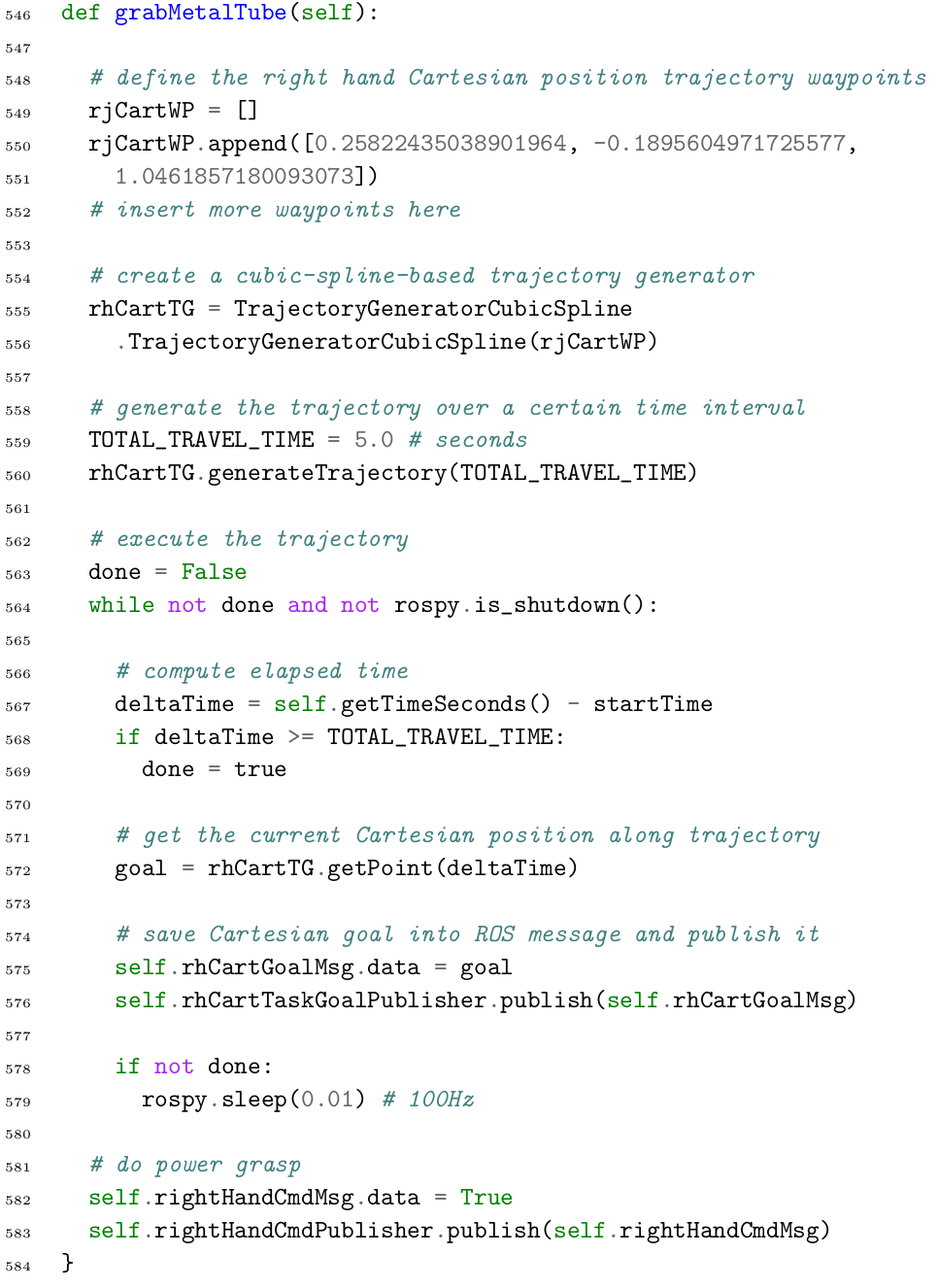}
 
 
 
 
 
       
 

 
\vspace{-.1in}
\caption{Code fragment from product disassembly application}
\label{fig:codeFragment}
\end{figure}

Figure~\ref{fig:codeFragment} contains an example code fragment from the
product disassembly . The application is written in the Python programming
language, though any programming language supported by ROS could be used
including C++. The code fragment shows how the Cartesian position trajectory is generated for
moving the right hand into a position where it can grab the metal tube. Lines
548-552 specify the Cartesian (x, y, z) waypoints that the hand is expected to
traverse. For brevity, only one waypoint is shown. Line 555 creates a cubic-spline interpolator, which is used on line 559 to generate the intermediate
points between the waypoints.  The while loop starting on line 564 obtains the
current goal Cartesian position based on the elapsed time (line 572) and
transmits this goal via a ROS topic (line 576). The goal parameter of the
right hand Cartesian position task within ControlIt! is bound to this ROS
topic enabling ControlIt! to follow the desired Cartesian trajectory. The
trajectory is transmitted at 100Hz, based on line 579. Once the trajectory is
done, line 583 issues a command to close the fingers in the right hand is
issued via another bound ROS topic.

\section{ControlIt! SMACH FSM Integration} \label{sec:smach}

The following screenshot is a visualization of the product disassembly finite state machine provided by ROS SMACH Visualizer. It is updated in real-time as the application in running. This particular screenshot shows that Dreamer is in the ``GrabValveState'' which is when her left gripper is being positioned to grab the valve.

\begin{figure}[b]
\centering
\includegraphics[width=.8\columnwidth]{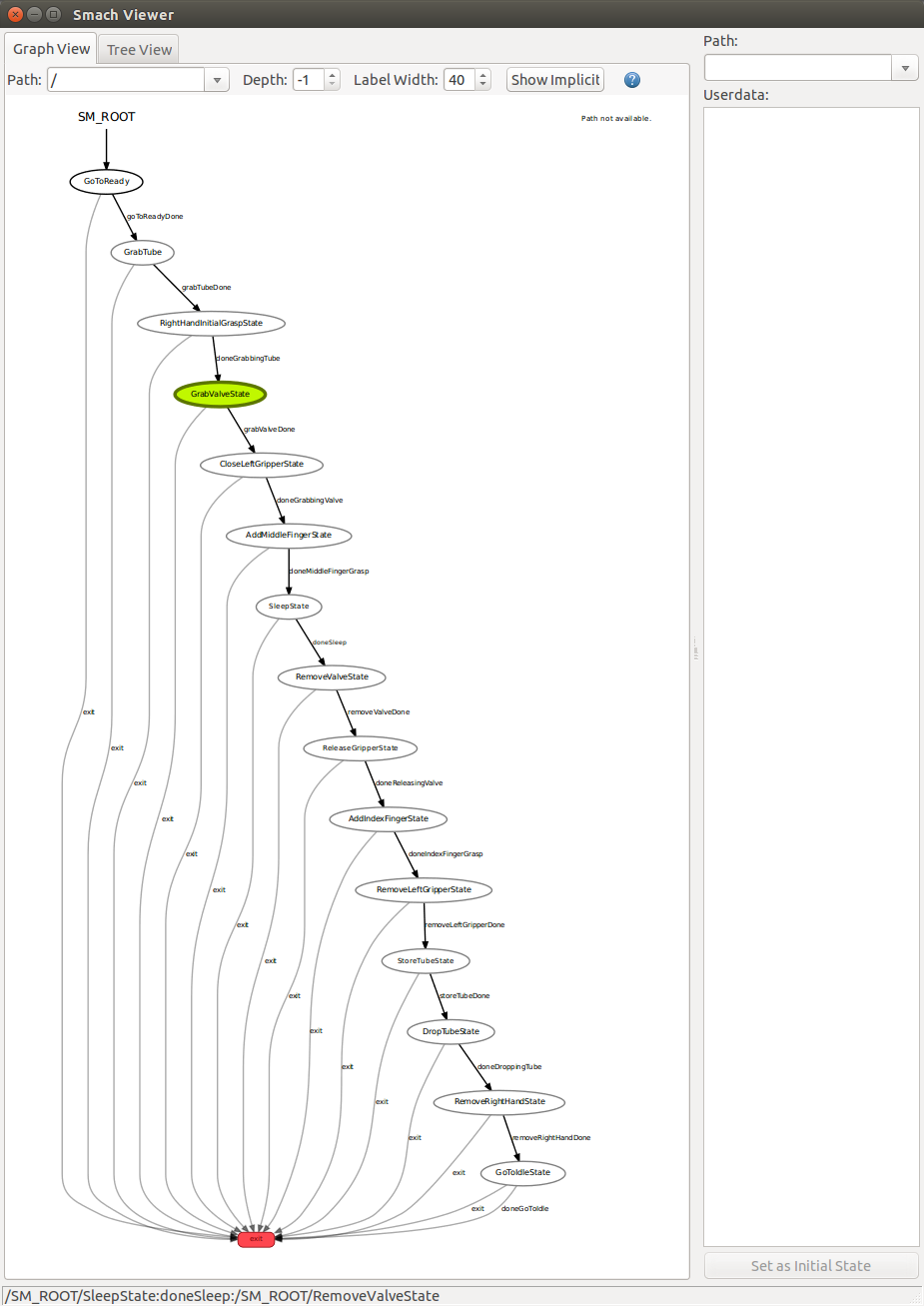}
\vspace{-.1in}
\caption{This figure shows a visualization of the FSM used by the product disassembly application. The ROS package SMACH is used to both implement the FSM logic and visualize its execution. The green arrow indicates the current state of the demo.}
\label{fig:smach}
\end{figure}


\bibliographystyle{IEEEtran}
\bibliography{main}

\begin{thebibliography}{100}
\providecommand{\url}[1]{#1}
\csname url@samestyle\endcsname
\providecommand{\newblock}{\relax}
\providecommand{\bibinfo}[2]{#2}
\providecommand{\BIBentrySTDinterwordspacing}{\spaceskip=0pt\relax}
\providecommand{\BIBentryALTinterwordstretchfactor}{4}
\providecommand{\BIBentryALTinterwordspacing}{\spaceskip=\fontdimen2\font plus
\BIBentryALTinterwordstretchfactor\fontdimen3\font minus
  \fontdimen4\font\relax}
\providecommand{\BIBforeignlanguage}[2]{{%
\expandafter\ifx\csname l@#1\endcsname\relax
\typeout{** WARNING: IEEEtran.bst: No hyphenation pattern has been}%
\typeout{** loaded for the language `#1'. Using the pattern for}%
\typeout{** the default language instead.}%
\else
\language=\csname l@#1\endcsname
\fi
#2}}
\providecommand{\BIBdecl}{\relax}
\BIBdecl

\bibitem{Sentis2005}
L.~Sentis and O.~Khatib, ``Synthesis of whole-body behaviors through
  hierarchical control of behavioral primitives,'' \emph{International Journal
  of Humanoid Robotics}, pp. 505--518, 2005.

\bibitem{Sentis2007-Thesis}
L.~Sentis, ``Synthesis and control of whole-body behaviors in humanoid
  systems,'' Ph.D. dissertation, Stanford University, 2007, supervised by
  Oussama Khatib,
  \url{http://www.me.utexas.edu/~lsentis/files/Thesis-Sentis-2007.pdf}.

\bibitem{Sentis2010}
L.~Sentis, J.~Park, and O.~Khatib, ``Compliant control of multicontact and
  center-of-mass behaviors in humanoid robots,'' \emph{IEEE Transactions on
  Robotics}, vol.~26, no.~4, pp. 483--501, 6 2010,
  \url{http://www.me.utexas.edu/~lsentis/files/tro-2010.pdf}.

\bibitem{Website-WBC_TC}
\BIBentryALTinterwordspacing
{IEEE Robotics and Automation Society}. (2015) Whole body control technical
  committee. [Online; accessed 13-February-2015]. [Online]. Available:
  \url{http://www.ieee-ras.org/whole-body-control}
\BIBentrySTDinterwordspacing

\bibitem{Website-MoveIt}
\BIBentryALTinterwordspacing
{Ioan A. Sucan and Sachin Chitta}. (2015) Moveit! [Online; accessed
  13-February-2015]. [Online]. Available: \url{http://moveit.ros.org/}
\BIBentrySTDinterwordspacing

\bibitem{Website_ROS_Control}
\BIBentryALTinterwordspacing
{Robot Operating System}. (2015) Ros control. [Online; accessed
  13-February-2015]. [Online]. Available: \url{http://wiki.ros.org/ros_control}
\BIBentrySTDinterwordspacing

\bibitem{Aghili2005}
F.~Aghili, ``A unified approach for inverse and direct dynamics of constrained
  multibody systems based on linear projection operator: applications to
  control and simulation,'' \emph{Robotics, IEEE Transactions on}, vol.~21,
  no.~5, pp. 834--849, Oct 2005.

\bibitem{Hyon2007}
S.-H. Hyon, J.~G. Hale, and G.~Cheng, ``Full-body compliant human -- humanoid
  interaction: Balancing in the presence of unknown external forces,''
  \emph{Robotics, IEEE Transactions on}, vol.~23, no.~5, pp. 884--898, Oct
  2007, \url{http://dx.doi.org/10.1109/TRO.2007.904896}.

\bibitem{Nakanishi2007}
J.~Nakanishi, M.~Mistry, and S.~Schaal, ``Inverse dynamics control with
  floating base and constraints,'' in \emph{Robotics and Automation, 2007 IEEE
  International Conference on}, April 2007, pp. 1942--1947,
  \url{http://dx.doi.org/10.1109/ROBOT.2007.363606}.

\bibitem{Mistry2010}
M.~Mistry, J.~Buchli, and S.~Schaal, ``Inverse dynamics control of floating
  base systems using orthogonal decomposition,'' in \emph{Robotics and
  Automation (ICRA), 2010 IEEE International Conference on}, May 2010, pp.
  3406--3412.

\bibitem{Nagasaka2010}
K.~Nagasaka, Y.~Kawanami, S.~Shimizu, T.~Kito, T.~Tsuboi, A.~Miyamoto,
  T.~Fukushima, and H.~Shimomura, ``Whole-body cooperative force control for a
  two-armed and two-wheeled mobile robot using generalized inverse dynamics and
  idealized joint units,'' in \emph{Robotics and Automation (ICRA), 2010 IEEE
  International Conference on}, May 2010, pp. 3377--3383.

\bibitem{Mistry2011}
M.~Mistry and L.~Righetti, ``Operational space control of constrained and
  underactuated systems,'' in \emph{Proceedings of Robotics: Science and
  Systems}, Los Angeles, CA, USA, June 2011.

\bibitem{Righetti2011}
L.~Righetti, J.~Buchli, M.~Mistry, and S.~Schaal, ``Inverse dynamics control of
  floating-base robots with external constraints: A unified view,'' in
  \emph{Robotics and Automation (ICRA), 2011 IEEE International Conference on},
  May 2011, pp. 1085--1090, \url{http://dx.doi.org/10.1109/ICRA.2011.5980156}.

\bibitem{Righetti2012}
L.~Righetti and S.~Schaal, ``Quadratic programming for inverse dynamics with
  optimal distribution of contact forces,'' in \emph{Humanoid Robots
  (Humanoids), 2012 12th IEEE-RAS International Conference on}, Nov 2012, pp.
  538--543, \url{http://dx.doi.org/10.1109/HUMANOIDS.2012.6651572}.

\bibitem{Wakita2011}
K.~Wakita, J.~Huang, P.~Di, K.~Sekiyama, and T.~Fukuda,
  ``Human-walking-intention-based motion control of an omnidirectional-type
  cane robot,'' \emph{Mechatronics, IEEE/ASME Transactions on}, vol.~18, no.~1,
  pp. 285--296, Feb 2013, \url{http://dx.doi.org/10.1109/TMECH.2011.2169980}.

\bibitem{Lee2012}
\BIBentryALTinterwordspacing
S.-H. Lee and A.~Goswami, ``\BIBforeignlanguage{English}{A momentum-based
  balance controller for humanoid robots on non-level and non-stationary
  ground},'' \emph{\BIBforeignlanguage{English}{Autonomous Robots}}, vol.~33,
  no.~4, pp. 399--414, 2012. [Online]. Available:
  \url{http://dx.doi.org/10.1007/s10514-012-9294-z}
\BIBentrySTDinterwordspacing

\bibitem{Salini2013_Thesis}
J.~Salini, ``Dynamic control for the task/posture coordination of humanoids:
  Towards synthesis of complex activities,'' Ph.D. dissertation, Universit
  Pierre et Marie Curie, 2013.

\bibitem{Moro2013}
F.~L. Moro, M.~Gienger, A.~Goswami, N.~G. Tsagarakis, and D.~G. Caldwell, ``An
  attractor-based whole-body motion control (wbmc) system for humanoid
  robots,'' in \emph{Humanoid Robots (Humanoids), 2013 13th IEEE-RAS
  International Conference on}, 2013, pp. 42--49.

\bibitem{Righetti2013}
\BIBentryALTinterwordspacing
L.~Righetti, J.~Buchli, M.~Mistry, M.~Kalakrishnan, and S.~Schaal,
  ``\BIBforeignlanguage{English}{Optimal distribution of contact forces with
  inverse-dynamics control},'' \emph{\BIBforeignlanguage{English}{The
  International Journal of Robotics Research}}, vol.~32, no.~3, pp. 280--298,
  2013. [Online]. Available: \url{http://ijr.sagepub.com/content/32/3/280}
\BIBentrySTDinterwordspacing

\bibitem{Saab2013}
L.~Saab, O.~Ramos, F.~Keith, N.~Mansard, P.~Soueres, and J.~Fourquet, ``Dynamic
  whole-body motion generation under rigid contacts and other unilateral
  constraints,'' \emph{Robotics, IEEE Transactions on}, vol.~29, no.~2, pp.
  346--362, April 2013.

\bibitem{Lengagne2013}
\BIBentryALTinterwordspacing
S.~Lengagne, J.~Vaillant, E.~Yoshida, and A.~Kheddar, ``Generation of
  whole-body optimal dynamic multi-contact motions,'' \emph{Int. J. Rob. Res.},
  vol.~32, no. 9-10, pp. 1104--1119, Aug. 2013. [Online]. Available:
  \url{http://dx.doi.org/10.1177/0278364913478990}
\BIBentrySTDinterwordspacing

\bibitem{Koolen2013-ICRA}
T.~Koolen, ``Force control for a humanoid robot using momentum and
  instantaneous capture point dynamics,'' in \emph{Robotics and Automation
  (ICRA), 2013 IEEE International Conference on}, 2013.

\bibitem{Henze2014}
B.~Henze, C.~Ott, and M.~Roa, ``Posture and balance control for humanoid robots
  in multi-contact scenarios based on model predictive control,'' in
  \emph{Intelligent Robots and Systems (IROS 2014), 2014 IEEE/RSJ International
  Conference on}, Sept 2014, pp. 3253--3258.

\bibitem{Righetti2014}
\BIBentryALTinterwordspacing
L.~Righetti, M.~Kalakrishnan, P.~Pastor, J.~Binney, J.~Kelly, R.~Voorhies,
  G.~Sukhatme, and S.~Schaal, ``\BIBforeignlanguage{English}{An autonomous
  manipulation system based on force control and optimization},''
  \emph{\BIBforeignlanguage{English}{Autonomous Robots}}, vol.~36, no. 1-2, pp.
  11--30, 2014. [Online]. Available:
  \url{http://dx.doi.org/10.1007/s10514-013-9365-9}
\BIBentrySTDinterwordspacing

\bibitem{Escande2014-IJRR}
A.~Escande, N.~Mansard, and P.-B. Wieber, ``Hierarchical quadratic programming:
  Fast online humanoid-robot motion generation,'' \emph{The International
  Journal of Robotics Research}, vol.~33, no.~7, pp. 1006--1028, 2014.

\bibitem{Hyon2009}
S.~Hyon, ``A motor control strategy with virtual musculoskeletal systems for
  compliant anthropomorphic robots,'' \emph{Mechatronics, IEEE/ASME
  Transactions on}, vol.~14, no.~6, pp. 677--688, Dec 2009,
  \url{http://dx.doi.org/10.1109/TMECH.2009.2033117}.

\bibitem{Sentis2013}
L.~Sentis, J.~Peterson, and R.~Philippsen, ``Implementation and stability
  analysis of prioritized whole-body compliant controllers on a wheeled
  humanoid robot in uneven terrains,'' \emph{Autonomous Robots}, vol.~35,
  no.~4, pp. 301--319, 2013,
  \url{http://www.me.utexas.edu/~lsentis/files/sentis-petersen-philippsen--auro-2013-2.pdf}.

\bibitem{Mizuuchi2007}
I.~Mizuuchi, Y.~Nakanishi, Y.~Sodeyama, Y.~Namiki, T.~Nishino, N.~Muramatsu,
  J.~Urata, K.~Hongo, T.~Yoshikai, and M.~Inaba, ``An advanced musculoskeletal
  humanoid kojiro,'' in \emph{Humanoid Robots, 2007 7th IEEE-RAS International
  Conference on}, Nov 2007, pp. 294--299.

\bibitem{Nakanishi2013}
Y.~Nakanishi, S.~Ohta, T.~Shirai, Y.~Asano, T.~Kozuki, Y.~Kakehashi,
  H.~Mizoguchi, T.~Kurotobi, Y.~Motegi, K.~Sasabuchi, J.~Urata, K.~Okada,
  I.~Mizuuchi, and M.~Inaba, ``Design approach of biologically-inspired
  musculoskeletal humanoids,'' \emph{Int. Journal of Advanced Robotic Systems},
  vol.~10, no. 216, 2013,
  \url{http://www.intechopen.com/journals/international_journal_of_advanced_robotic_systems/design-approach-of-biologically-inspired-musculoskeletal-humanoids}.

\bibitem{Dietrich2013}
A.~Dietrich, C.~Ott, and A.~Albu-Schaffer, ``Multi-objective compliance control
  of redundant manipulators: Hierarchy, control, and stability,'' in
  \emph{Intelligent Robots and Systems (IROS), 2013 IEEE/RSJ International
  Conference on}, Nov 2013, pp. 3043--3050.

\bibitem{Ott2011}
C.~Ott, M.~Roa, and G.~Hirzinger, ``Posture and balance control for biped
  robots based on contact force optimization,'' in \emph{Humanoid Robots
  (Humanoids), 2011 11th IEEE-RAS International Conference on}, Oct 2011, pp.
  26--33.

\bibitem{Englsberger2013}
J.~Englsberger, C.~Ott, and A.~Albu-Schaffer, ``Three-dimensional bipedal
  walking control using divergent component of motion,'' in \emph{Intelligent
  Robots and Systems (IROS), 2013 IEEE/RSJ International Conference on}, Nov
  2013, pp. 2600--2607.

\bibitem{Moro2014}
\BIBentryALTinterwordspacing
F.~Moro, N.~Tsagarakis, and D.~Caldwell, ``\BIBforeignlanguage{English}{Walking
  in the resonance with the {COMAN} robot with trajectories based on human
  kinematic motion primitives ({kMPs})},''
  \emph{\BIBforeignlanguage{English}{Autonomous Robots}}, vol.~36, no.~4, pp.
  331--347, 2014. [Online]. Available:
  \url{http://dx.doi.org/10.1007/s10514-013-9357-9}
\BIBentrySTDinterwordspacing

\bibitem{Whiteman2010}
E.~Whitman and C.~Atkeson, ``Control of instantaneously coupled systems applied
  to humanoid walking,'' in \emph{Humanoid Robots (Humanoids), 2010 10th
  IEEE-RAS International Conference on}, Dec 2010, pp. 210--217.

\bibitem{Hutter2013}
M.~Hutter, M.~Bloesch, J.~Buchli, C.~Semini, S.~Bazeille, L.~Righetti, and
  J.~Bohg, ``{AGILITY} - dynamic full body locomotion and manipulation with
  autonomous legged robots,'' in \emph{Safety, Security, and Rescue Robotics
  (SSRR), 2013 IEEE International Symposium on}, Oct 2013, pp. 1--4.

\bibitem{Hirai1998}
K.~Hirai, M.~Hirose, Y.~Haikawa, and T.~Takenaka, ``The development of honda
  humanoid robot,'' in \emph{Robotics and Automation, 1998. Proceedings. 1998
  IEEE International Conference on}, vol.~2, May 1998, pp. 1321--1326 vol.2.

\bibitem{Kajita2003}
S.~Kajita, F.~Kanehiro, K.~Kaneko, K.~Fujiwara, K.~Harada, K.~Yokoi, and
  H.~Hirukawa, ``Resolved momentum control: humanoid motion planning based on
  the linear and angular momentum,'' in \emph{Intelligent Robots and Systems,
  2003. (IROS 2003). Proceedings. 2003 IEEE/RSJ International Conference on},
  vol.~2, Oct 2003, pp. 1644--1650 vol.2.

\bibitem{Bouyarmane2012}
K.~Bouyarmane and A.~Kheddar, ``On the dynamics modeling of free-floating-base
  articulated mechanisms and applications to humanoid whole-body dynamics and
  control,'' in \emph{Humanoid Robots (Humanoids), 2012 12th IEEE-RAS
  International Conference on}, Nov 2012, pp. 36--42.

\bibitem{Ohmichi1985}
T.~Ohmichi, S.~Hosaka, M.~Nishihara, T.~Ibe, A.~Okino, J.~Nakayama, T.~Miida,
  and M.~Ishida, ``Development of the multi-function robot for the containment
  vessel of the nuclear plant,'' in \emph{International conference on advanced
  robotics}, vol.~19, no.~20, 1985.

\bibitem{Hirose1991}
S.~Hirose, A.~Morishima, S.~Tukagosi, T.~Tsumaki, and H.~Monobe, ``Design of
  practical snake vehicle: articulated body mobile robot {KR-II},'' in
  \emph{Advanced Robotics, 1991. 'Robots in Unstructured Environments', 91
  ICAR., Fifth International Conference on}, June 1991, pp. 833--838 vol.1.

\bibitem{Eiji1993}
N.~Eiji and N.~Sei, ``Leg-wheel robot: a futuristic mobile platform for
  forestry industry,'' in \emph{Advanced Robotics, 1993. Can Robots Contribute
  to Preventing Environmental Deterioration? Proceedings, 1993 IEEE/Tsukuba
  International Workshop on}, Nov 1993, pp. 109--112.

\bibitem{Matsumoto1995}
O.~Matsumoto, S.~Kajita, K.~Tani, and M.~Oooto, ``A four-wheeled robot to pass
  over steps by changing running control modes,'' in \emph{Robotics and
  Automation, 1995. Proceedings., 1995 IEEE International Conference on},
  vol.~2, May 1995, pp. 1700--1706 vol.2.

\bibitem{Asfour2000}
T.~Asfour, K.~Berns, and R.~Dillmann, ``The humanoid robot {ARMAR}: Design and
  control,'' in \emph{IN IEEE/APS INTL CONFERENCE ON HUMANOID ROBOTS}, 2000,
  pp. 7--8.

\bibitem{Katz2006}
D.~Katz, E.~Horrell, Y.~Yang, B.~Burns, T.~Buckley, A.~Grishkan,
  V.~Zhylkovskyy, O.~Brock, and E.~Learned-Miller, ``The umass mobile
  manipulator uman: An experimental platform for autonomous mobile
  manipulation.'' in \emph{Workshop on Manipulation in Human Environments at
  Robotics: Science and systems.}, 2006.

\bibitem{Albu-Schaeffer2007}
\BIBentryALTinterwordspacing
C.~Loughlin, A.~Albu‐Schäffer, S.~Haddadin, C.~Ott, A.~Stemmer, T.~Wimböck,
  and G.~Hirzinger, ``The {DLR} lightweight robot: design and control concepts
  for robots in human environments,'' \emph{Industrial Robot: An International
  Journal}, vol.~34, no.~5, pp. 376--385, 2007. [Online]. Available:
  \url{http://dx.doi.org/10.1108/01439910710774386}
\BIBentrySTDinterwordspacing

\bibitem{Borst2007}
C.~Borst, C.~Ott, T.~Wimbock, B.~Brunner, F.~Zacharias, B.~Bauml,
  U.~Hillenbrand, S.~Haddadin, A.~Albu-Schäffer, and G.~Hirzinger, ``A
  humanoid upper body system for two-handed manipulation,'' in \emph{Robotics
  and Automation, 2007 IEEE International Conference on}, April 2007, pp.
  2766--2767.

\bibitem{Theobold2008}
D.~Theobold, J.~Ornstein, J.~G. Nichol, and S.~E. Kullberg, ``Mobile robot
  platform google patents,'' US Patent, 7 2008, 7,348,747.

\bibitem{Freitas2009}
G.~Freitas, F.~Lizarralde, L.~Hsu, and N.~Reis, ``Kinematic reconfigurability
  of mobile robots on irregular terrains,'' in \emph{Robotics and Automation,
  2009. ICRA '09. IEEE International Conference on}, May 2009, pp. 1340--1345.

\bibitem{Beetz2010}
M.~Beetz, L.~Mosenlechner, and M.~Tenorth, ``{CRAM} -- a cognitive robot
  abstract machine for everyday manipulation in human environments,'' in
  \emph{Intelligent Robots and Systems (IROS), 2010 IEEE/RSJ International
  Conference on}, Oct 2010, pp. 1012--1017.

\bibitem{Iwata2009}
H.~Iwata and S.~Sugano, ``Design of human symbiotic robot {TWENDY-ONE},
  year={2009}, month={May}, pages={580-586}, keywords={control system
  synthesis;human-robot interaction;humanoid robots;manipulators;mobile
  robots;service robots;TWENDY-ONE robot;anthropomorphic dual hand;compact
  passive mechanism;dexterity function;elderly person physical support;human
  symbiotic robot design;kitchen support robot;mobility function;omni-wheeled
  vehicle;physical
  support;Anthropomorphism;Humans;Layout;Manipulators;Robots;Safety;Senior
  citizens;Skin;Symbiosis;Vehicles}, doi={10.1109/ROBOT.2009.5152702},
  issn={1050-4729},'' in \emph{Robotics and Automation, 2009. ICRA '09. IEEE
  International Conference on}.

\bibitem{Reiser2009}
U.~Reiser, C.~Connette, J.~Fischer, J.~Kubacki, A.~Bubeck, F.~Weisshardt,
  T.~Jacobs, C.~Parlitz, M.~Hagele, and A.~Verl, ``Care-o-bot 3 - creating a
  product vision for service robot applications by integrating design and
  technology,'' in \emph{Intelligent Robots and Systems, 2009. IROS 2009.
  IEEE/RSJ International Conference on}, Oct 2009, pp. 1992--1998.

\bibitem{King2010}
\BIBentryALTinterwordspacing
C.-H. King, T.~L. Chen, A.~Jain, and C.~C. Kemp, ``Towards an assistive robot
  that autonomously performs bed baths for patient hygiene.'' in
  \emph{IROS}.\hskip 1em plus 0.5em minus 0.4em\relax IEEE, 2010, pp. 319--324.
  [Online]. Available:
  \url{http://dblp.uni-trier.de/db/conf/iros/iros2010.html#KingCJK10}
\BIBentrySTDinterwordspacing

\bibitem{Stephens2010}
B.~Stephens and C.~Atkeson, ``Dynamic balance force control for compliant
  humanoid robots,'' in \emph{Intelligent Robots and Systems (IROS), 2010
  IEEE/RSJ International Conference on}, Oct 2010, pp. 1248--1255.

\bibitem{Stilman2010}
M.~Stilman, J.~Olson, and W.~Gloss, ``Golem krang: Dynamically stable humanoid
  robot for mobile manipulation,'' in \emph{Robotics and Automation (ICRA),
  2010 IEEE International Conference on}, May 2010, pp. 3304--3309.

\bibitem{Meeussen2010}
W.~Meeussen, M.~Wise, S.~Glaser, S.~Chitta, C.~McGann, P.~Mihelich,
  E.~Marder-Eppstein, M.~Muja, V.~Eruhimov, T.~Foote, J.~Hsu, R.~Rusu,
  B.~Marthi, G.~Bradski, K.~Konolige, B.~Gerkey, and E.~Berger, ``Autonomous
  door opening and plugging in with a personal robot,'' in \emph{Robotics and
  Automation (ICRA), 2010 IEEE International Conference on}, May 2010, pp.
  729--736.

\bibitem{Hart2011_R2}
S.~Hart, J.~Yamokoski, and M.~Diftler, ``Robonaut 2: A new platform for
  human-centered robot learning,'' \emph{Robotics Science and Systems}, 2011.

\bibitem{Stephens2011}
B.~J. Stephens, ``Push recovery control for force-controlled humanoid robots,''
  Ph.D. dissertation, Carnegie Mellon University, 2011.

\bibitem{Moro2011}
F.~Moro, N.~Tsagarakis, and D.~Caldwell, ``A human-like walking for the
  {COmpliant huMANoid COMAN} based on {CoM} trajectory reconstruction from
  kinematic motion primitives,'' in \emph{Humanoid Robots (Humanoids), 2011
  11th IEEE-RAS International Conference on}, Oct 2011, pp. 364--370.

\bibitem{Tsagarakis2011}
N.~Tsagarakis, Z.~Li, J.~Saglia, and D.~Caldwell, ``The design of the lower
  body of the compliant humanoid robot ``{cCub}'','' in \emph{Robotics and
  Automation (ICRA), 2011 IEEE International Conference on}, May 2011, pp.
  2035--2040.

\bibitem{Bertrand2014}
S.~Bertrand and J.~Pratt, ``Momentum-based control framework: application to
  the humanoid robots atlas and valkyrie,'' in \emph{IROS 2014 Workshop on
  Whole-Body Control for Robots in the Real World}, 2014.

\bibitem{Herzog2014_IROS}
A.~Herzog, L.~Righetti, F.~Grimminger, P.~Pastor, and S.~Schaal, ``Balancing
  experiments on a torque-controlled humanoid with hierarchical inverse
  dynamics,'' in \emph{Proceeedings of 2014 IEEE/RSJ International Conference
  on Intelligent Robots and Systems}, 2014.

\bibitem{Herzog2014_Arxiv}
\BIBentryALTinterwordspacing
------, ``Momentum-based balance control for torque-controlled humanoids,''
  \emph{CoRR}, vol. abs/1305.2042, 2013. [Online]. Available:
  \url{http://arxiv.org/abs/1305.2042}
\BIBentrySTDinterwordspacing

\bibitem{hutter2012}
M.~Hutter, C.~Gehring, M.~Bloesch, M.~A. Hoepflinger, C.~D. Remy, and
  R.~Siegwart, ``Starl{ETH}: {A} compliant quadrupedal robot for fast,
  efficient, and versatile locomotion,'' in \emph{15{t}h {I}nternational
  {C}onference on {C}limbing and {W}alking {R}obot - {CLAWAR} 2012}, 2012.

\bibitem{Hutter2014}
M.~Hutter, H.~Sommer, C.~Gehring, M.~Hoepflinger, M.~Bloesch, and R.~Siegwart,
  ``Quadrupedal locomotion using hierarchical operational space control,''
  \emph{The International Journal of Robotics Research}, 2014.

\bibitem{Fuchs2009}
M.~Fuchs, C.~Borst, P.~Giordano, A.~Baumann, E.~Kraemer, J.~Langwald,
  R.~Gruber, N.~Seitz, G.~Plank, K.~Kunze, R.~Burger, F.~Schmidt, T.~Wimboeck,
  and G.~Hirzinger, ``Rollin' justin - design considerations and realization of
  a mobile platform for a humanoid upper body,'' in \emph{Robotics and
  Automation, 2009. ICRA '09. IEEE International Conference on}, May 2009, pp.
  4131--4137.

\bibitem{Semini2011}
C.~Semini, N.~G. Tsagarakis, E.~Guglielmino, M.~Focchi, F.~Cannella, and D.~G.
  Caldwell, ``Design of {HyQ} -- a hydraulically and electrically actuated
  quadruped robot,'' in \emph{Proceedings of the Institution of Mechanical
  Engineers, Part I: Journal of Systems and Control Engineering}, vol. 225,
  no.~6, August 2011, pp. 831--849.

\bibitem{Tellez2008}
R.~Tellez, F.~Ferro, S.~Garcia, E.~Gomez, E.~Jorge, D.~Mora, D.~Pinyol,
  J.~Oliver, O.~Torres, J.~Velazquez, and D.~Faconti, ``{Reem-B}: An autonomous
  lightweight human-size humanoid robot,'' in \emph{Humanoid Robots, 2008.
  Humanoids 2008. 8th IEEE-RAS International Conference on}, Dec 2008, pp.
  462--468.

\bibitem{Phlippsen2011}
\emph{An Open Source Extensible Software Package to Create Whole-Body Compliant
  Skills in Personal Mobile Manipulators}, 2011,
  \url{http://www.me.utexas.edu/~lsentis/files/iros-wbc-2011.pdf}.

\bibitem{Website-UTA-WBC}
\BIBentryALTinterwordspacing
{Human Centered Robotics Laboratory at the University of Texas at Austin}.
  (2015) Uta-wbc. [Online; accessed 13-February-2015]. [Online]. Available:
  \url{https://github.com/lsentis/uta-wbc-dreamer}
\BIBentrySTDinterwordspacing

\bibitem{Heineman2001}
G.~T. Heineman and W.~T. Councill, Eds., \emph{Component-based Software
  Engineering: Putting the Pieces Together}.\hskip 1em plus 0.5em minus
  0.4em\relax Boston, MA, USA: Addison-Wesley Longman Publishing Co., Inc.,
  2001.

\bibitem{Szyperski2002}
C.~Szyperski, \emph{Component Software: Beyond Object-Oriented Programming},
  2nd~ed.\hskip 1em plus 0.5em minus 0.4em\relax Boston, MA, USA:
  Addison-Wesley Longman Publishing Co., Inc., 2002.

\bibitem{Kanehiro2002}
F.~Kanehiro, K.~Fujiwara, S.~Kajita, K.~Yokoi, K.~Kaneko, H.~Hirukawa,
  Y.~Nakamura, and K.~Yamane, ``Open architecture humanoid robotics platform,''
  in \emph{Robotics and Automation, 2002. Proceedings. ICRA '02. IEEE
  International Conference on}, vol.~1, 2002, pp. 24--30 vol.1.

\bibitem{Hirukawa2004}
\BIBentryALTinterwordspacing
H.~Hirukawa, F.~Kanehiro, K.~Kaneko, S.~Kajita, K.~Fujiwara, Y.~Kawai,
  F.~Tomita, S.~Hirai, K.~Tanie, T.~Isozumi, K.~Akachi, T.~Kawasaki, S.~Ota,
  K.~Yokoyama, H.~Handa, Y.~Fukase, J.~ichiro Maeda, Y.~Nakamura, S.~Tachi, and
  H.~Inoue, ``Humanoid robotics platforms developed in \{HRP\},''
  \emph{Robotics and Autonomous Systems}, vol.~48, no.~4, pp. 165 -- 175, 2004,
  humanoids 2003. [Online]. Available:
  \url{http://www.sciencedirect.com/science/article/pii/S0921889004000946}
\BIBentrySTDinterwordspacing

\bibitem{Ando2005}
N.~Ando, T.~Suehiro, K.~Kitagaki, T.~Kotoku, and W.-K. Yoon, ``{RT}-middleware:
  distributed component middleware for rt (robot technology),'' in
  \emph{Intelligent Robots and Systems, 2005. (IROS 2005). 2005 IEEE/RSJ
  International Conference on}, Aug 2005, pp. 3933--3938.

\bibitem{Website-Orocos-Toolchain}
\BIBentryALTinterwordspacing
{Orocos}. (2015) Orocos toolchain. [Online; accessed 13-February-2015].
  [Online]. Available: \url{http://www.orocos.org/toolchain}
\BIBentrySTDinterwordspacing

\bibitem{Metta2006_YARP}
G.~Metta, P.~Fitzpatrick, and L.~Natale, ``{YARP}: Yet another robot
  platform,'' \emph{International Journal on Advanced Robotics Systems},
  vol.~3, no.~1, 2006.

\bibitem{Website-ROS}
\BIBentryALTinterwordspacing
{Robot Operating System}. (2015) Ros. [Online; accessed 13-February-2015].
  [Online]. Available: \url{http://www.ros.org/}
\BIBentrySTDinterwordspacing

\bibitem{Quigley2009_ROS}
M.~Quigley, K.~Conley, B.~Gerkey, J.~Faust, T.~Foote, J.~Leibs, R.~Wheeler, and
  A.~Y. Ng, ``{ROS}: an open-source robot operating system,'' \emph{ICRA
  workshop on open source software}, vol.~3, no. 3.2, p.~5, 2009.

\bibitem{Nesnas2006}
I.~A. Nesnas, R.~Simmons, D.~Gaines, C.~Kunz, A.~Dias-Caldron, T.~Estlin,
  R.~Madison, J.~Guineau, M.~McHenry, I.~Shu, and D.~Apfelbaum, ``{CLARAty}:
  Challenges and steps toward reusable robotic software,'' \emph{Int. Journal
  of Advanced Robotic Systems}, vol.~3, no.~1, 2006.

\bibitem{Nesnas2007}
I.~A. Nesnas, ``{CLARAty}: A collaborative software for advancing robotic
  technologies,'' in \emph{Proc. of NASA Science and Technology Conference},
  June 2007.

\bibitem{Hirzinger2006}
G.~Hirzinger and B.~Bauml, ``Agile robot development ({aRD}): A pragmatic
  approach to robotic software,'' in \emph{Intelligent Robots and Systems, 2006
  IEEE/RSJ International Conference on}, Oct 2006, pp. 3741--3748.

\bibitem{Website-Microblx}
\BIBentryALTinterwordspacing
{Microblx}. (2014) Microblx - a lightweight, dynamic, reflective, hard
  real-time safe function block framework. [Online; accessed 13-February-2015].
  [Online]. Available: \url{http://www.microblx.org/}
\BIBentrySTDinterwordspacing

\bibitem{Klotzbuecher2013}
M.~Klotzbuecher and H.~Bruyninckx, ``microblx: a reflective, real-time safe,
  embedded function block framework,'' in \emph{15th Real Time Linux Workshop},
  October 2013.

\bibitem{Website-OpenRDK}
\BIBentryALTinterwordspacing
{RoCoCo Laboratory}. (2015) Open robot development kit. [Online; accessed
  13-February-2015]. [Online]. Available: \url{http://openrdk.sourceforge.net/}
\BIBentrySTDinterwordspacing

\bibitem{Calisi2008}
D.~Calisi, A.~Censi, L.~Iocchi, and D.~Nardi, ``{OpenRDK}: A modular framework
  for robotic software development,'' in \emph{Intelligent Robots and Systems,
  2008. IROS 2008. IEEE/RSJ International Conference on}, Sept 2008, pp.
  1872--1877.

\bibitem{Calisi2012}
D.~Calisi, A.~Censi, L.~Locchi, and D.~Nardi, ``Design choices for modular and
  flexible robotic software development: the {OpenRDK} viewpoint,''
  \emph{Journal of Software Engineering for Robotics}, vol.~3, no.~1, 2012,
  \url{http://joser.unibg.it/index.php?journal=joser&page=article&op=view&path%5B%5D=48}.

\bibitem{Munich2005}
M.~Munich, J.~Ostrowski, and P.~Pirjanian, ``{ERSP}: a software platform and
  architecture for the service robotics industry,'' in \emph{Intelligent Robots
  and Systems, 2005. (IROS 2005). 2005 IEEE/RSJ International Conference on},
  Aug 2005, pp. 460--467.

\bibitem{Website-RTAI}
\BIBentryALTinterwordspacing
{Dipartimento Di Scienze e Tecnologie Aerospaziali del Politecnico di Milano}.
  (2015) Real-time application interface. [Online; accessed 13-February-2015].
  [Online]. Available: \url{https://www.rtai.org/}
\BIBentrySTDinterwordspacing

\bibitem{Tsouroukdissian2014}
A.~R. Tsouroukdissian, ``Ros control, an overview,'' in \emph{ROSCon 2014},
  September 2014.

\bibitem{Website-Conman}
\BIBentryALTinterwordspacing
J.~Bohren. (2015) Conman - a robot state estimator and controller manager for
  use in orocos rtt and ros. [Online; accessed 13-February-2015]. [Online].
  Available: \url{https://github.com/jbohren/conman}
\BIBentrySTDinterwordspacing

\bibitem{Website-iTaSC}
\BIBentryALTinterwordspacing
{Orocos Wiki}. (2015) itasc (instantaneous task specification using
  constraints) website. [Online; accessed 13-February-2015]. [Online].
  Available: \url{http://www.orocos.org/wiki/orocos/itasc-wiki}
\BIBentrySTDinterwordspacing

\bibitem{Schutter2007}
\BIBentryALTinterwordspacing
J.~De~Schutter, T.~De~Laet, J.~Rutgeerts, W.~Decré, R.~Smits, E.~Aertbeliën,
  K.~Claes, and H.~Bruyninckx, ``Constraint-based task specification and
  estimation for sensor-based robot systems in the presence of geometric
  uncertainty,'' \emph{The International Journal of Robotics Research},
  vol.~26, no.~5, pp. 433--455, 2007. [Online]. Available:
  \url{http://ijr.sagepub.com/content/26/5/433.abstract}
\BIBentrySTDinterwordspacing

\bibitem{Decre2009}
W.~Decre, R.~Smits, H.~Bruyninckx, and J.~De~Schutter, ``Extending {iTaSC} to
  support inequality constraints and non-instantaneous task specification,'' in
  \emph{Robotics and Automation, 2009. ICRA '09. IEEE International Conference
  on}, May 2009, pp. 964--971.

\bibitem{Decre2013}
W.~Decre, H.~Bruyninckx, and J.~De~Schutter, ``Extending the {iTaSC}
  constraint-based robot task specification framework to time-independent
  trajectories and user-configurable task horizons,'' in \emph{Robotics and
  Automation (ICRA), 2013 IEEE International Conference on}, May 2013, pp.
  1941--1948.

\bibitem{Website-ROCK}
\BIBentryALTinterwordspacing
{Rock Robotics}. (2015) Rock - the robot construction kit. [Online; accessed
  13-February-2015]. [Online]. Available:
  \url{http://rock-robotics.org/stable/}
\BIBentrySTDinterwordspacing

\bibitem{Brunner1999}
B.~Brunner, K.~Landzettel, G.~Schreiber, B.~Stinmetz, and G.~Girzinger, ``A
  universal task level ground control and programming system for space robot
  applications - the {MARCO} concept and its application to the ets vii
  project,'' in \emph{Proc. of the 5th iSAIRAS Int. Symp. on Artifical
  Intelligence, Robotics, and Automation in Space}, 1999.

\bibitem{Fleury1997}
S.~Fleury, M.~Herrb, and R.~Chatila, ``{GenoM}: a tool for the specification
  and the implementation of operating modules in a distributed robot
  architecture,'' in \emph{Intelligent Robots and Systems, 1997. IROS '97.,
  Proceedings of the 1997 IEEE/RSJ International Conference on}, vol.~2, Sep
  1997, pp. 842--849 vol.2.

\bibitem{Website-Ecto}
\BIBentryALTinterwordspacing
{Willow Garage}. (2015) Ecto - a c++/python computation graph framework.
  [Online; accessed 13-February-2015]. [Online]. Available:
  \url{http://plasmodic.github.io/ecto/}
\BIBentrySTDinterwordspacing

\bibitem{Hart2014}
S.~Hart, P.~Dinh, J.~Yamokoski, B.~Wightman, and N.~Radford, ``Robot task
  commander: A framework and {IDE} for robot application development,'' in
  \emph{Intelligent Robots and Systems (IROS 2014), 2014 IEEE/RSJ International
  Conference on}, Sept 2014, pp. 1547--1554.

\bibitem{Arkin1990}
R.~Arkin and R.~Murphy, ``Autonomous navigation in a manufacturing
  environment,'' \emph{Robotics and Automation, IEEE Transactions on}, vol.~6,
  no.~4, pp. 445--454, Aug 1990.

\bibitem{Alami1998}
R.~Alami, R.~Chatila, S.~Fleury, M.~Ghallab, and F.~Ingrand, ``An architecture
  for autonomy,'' \emph{INTERNATIONAL JOURNAL OF ROBOTICS RESEARCH}, vol.~17,
  pp. 315--337, 1998.

\bibitem{Jenkins2004}
\BIBentryALTinterwordspacing
O.~C. JENKINS and M.~J. MATARIĆ, ``Performance-derived behavior vocabularies:
  Data-driven acquisition of skills from motion,'' \emph{International Journal
  of Humanoid Robotics}, vol.~01, no.~02, pp. 237--288, 2004. [Online].
  Available:
  \url{http://www.worldscientific.com/doi/abs/10.1142/S0219843604000186}
\BIBentrySTDinterwordspacing

\bibitem{Pastor2009}
P.~Pastor, H.~Hoffmann, T.~Asfour, and S.~Schaal, ``Learning and generalization
  of motor skills by learning from demonstration,'' in \emph{Robotics and
  Automation, 2009. ICRA '09. IEEE International Conference on}, May 2009, pp.
  763--768.

\bibitem{Kim2010}
K.~Kim, J.-Y. Lee, D.~Choi, J.-M. Park, and B.-J. You, ``Autonomous task
  execution of a humanoid robot using a cognitive model,'' in \emph{Robotics
  and Biomimetics (ROBIO), 2010 IEEE International Conference on}, Dec 2010,
  pp. 405--410.

\bibitem{Ott2013}
C.~Ott, B.~Henze, and D.~Lee, ``Kinesthetic teaching of humanoid motion based
  on whole-body compliance control with interaction-aware balancing,'' in
  \emph{Intelligent Robots and Systems (IROS), 2013 IEEE/RSJ International
  Conference on}, Nov 2013, pp. 4615--4621.

\bibitem{Simmons1998}
R.~Simmons and D.~Apfelbaum, ``A task description language for robot control,''
  in \emph{Intelligent Robots and Systems, 1998. Proceedings., 1998 IEEE/RSJ
  International Conference on}, vol.~3, Oct 1998, pp. 1931--1937 vol.3.

\bibitem{Kortenkamp1999}
D.~Kortenkamp, R.~Burridge, R.~P. Bonasso, D.~Schreckenghost, and M.~B. Hudson,
  ``An intelligent software architecture for semiautonomous robot control,'' in
  \emph{In Autonomy Control Software Workshop, Autonomous Agents 99}, 1999, pp.
  36--43.

\bibitem{Website-RBDL}
\BIBentryALTinterwordspacing
{Martin Felis}. (2015) Rigid body dynamics library. [Online; accessed
  13-February-2015]. [Online]. Available: \url{http://rbdl.bitbucket.org/}
\BIBentrySTDinterwordspacing

\bibitem{Website-ROS-Pluginlib}
\BIBentryALTinterwordspacing
{Robot Operating System}. (2015) Ros pluginlib. [Online; accessed
  13-February-2015]. [Online]. Available: \url{http://wiki.ros.org/pluginlib}
\BIBentrySTDinterwordspacing

\bibitem{Website-ROS-SHM}
\BIBentryALTinterwordspacing
J.~James. (2015) Ros shared memory interface. [Online; accessed
  13-February-2015]. [Online]. Available:
  \url{https://bitbucket.org/jraipxg/ros_shared_memory_interface}
\BIBentrySTDinterwordspacing

\bibitem{Website-muParser}
\BIBentryALTinterwordspacing
{Ingo Berg}. (2015) muparser - a fast math library. [Online; accessed
  13-February-2015]. [Online]. Available: \url{http://muparser.beltoforion.de/}
\BIBentrySTDinterwordspacing

\bibitem{Website-ROS-Launch}
\BIBentryALTinterwordspacing
{Robot Operating System}. (2014) Ros launch. [Online; accessed
  13-February-2015]. [Online]. Available: \url{http://wiki.ros.org/roslaunch}
\BIBentrySTDinterwordspacing

\bibitem{Website-ROS-Bag}
\BIBentryALTinterwordspacing
------. (2015) Ros bag. [Online; accessed 13-February-2015]. [Online].
  Available: \url{http://wiki.ros.org/rosbag}
\BIBentrySTDinterwordspacing

\bibitem{Website_Gazebo}
\BIBentryALTinterwordspacing
{Open Source Robotics Foundation}. (2015) Gazebo simulator website. [Online;
  accessed 13-February-2015]. [Online]. Available: \url{http://gazebosim.org/}
\BIBentrySTDinterwordspacing

\bibitem{Website-ROS-Client}
\BIBentryALTinterwordspacing
{Robot Operating System}. (2015) Ros client libraries. [Online; accessed
  13-February-2015]. [Online]. Available:
  \url{http://wiki.ros.org/Client%20Libraries}
\BIBentrySTDinterwordspacing

\bibitem{Radford2014}
N.~A. Radford, P.~Strawser, K.~Hambuchen, J.~S. Mehling, W.~K. Verdeyen,
  S.~Donnan, J.~Holley, J.~Sanchez, V.~Nguyen, L.~Bridgwater, R.~Berka,
  R.~Ambrose, C.~McQuin, J.~D. Yamokoski, S.~Hart, R.~Guo, A.~Parsons,
  B.~Wightman, P.~Dinh, B.~Ames, C.~Blakely, C.~Edmonson, B.~Sommers, R.~Rea,
  C.~Tobler, H.~Bibby, B.~Howard, L.~Nui, A.~Lee, M.~Conover, L.~Truong,
  D.~Chesney, R.~P. Jr., G.~Johnson, C.-L. Fok, N.~Paine, L.~Sentis,
  E.~Cousineau, R.~Sinnet, J.~Lack, M.~Powell, B.~Morris, and A.~Ames,
  ``Valkyrie: {NASA's} first bipedal humanoid robot,'' \emph{Journal of Field
  Robotics}, 10 2014,
  \url{http://www.me.utexas.edu/~hcrl/publications/JFR-NASA-HCRL-Final.pdf}.

\bibitem{Kim2015_Arxiv}
\BIBentryALTinterwordspacing
D.~Kim, Y.~Zhao, G.~Thomas, and L.~Sentis, ``Accessing whole-body operational
  space control in a point-foot series elastic biped: Balance on split terrain
  and undirected walking,'' \emph{Arxive preprint}, 2015. [Online]. Available:
  \url{http://arxiv.org/abs/1501.02855}
\BIBentrySTDinterwordspacing

\bibitem{Zhao2015}
Y.~Zhao, N.~Paine, K.~Kim, and L.~Sentis, ``Stability and performance limits of
  latency-prone distributed feedback controllers,'' 2015,
  http://arxiv.org/pdf/1501.02854v1.pdf.

\bibitem{Johnson2015_IHMC}
\BIBentryALTinterwordspacing
M.~Johnson, B.~Shrewsbury, S.~Bertrand, T.~Wu, D.~Duran, M.~Floyd, P.~Abeles,
  D.~Stephen, N.~Mertins, A.~Lesman, J.~Carff, W.~Rifenburgh, P.~Kaveti,
  W.~Straatman, J.~Smith, M.~Griffioen, B.~Layton, T.~de~Boer, T.~Koolen,
  P.~Neuhaus, and J.~Pratt, ``Team {IHMC}'s lessons learned from the {DARPA}
  robotics challenge trials,'' \emph{Journal of Field Robotics}, vol.~32,
  no.~2, pp. 192--208, 2015. [Online]. Available:
  \url{http://onlinelibrary.wiley.com/doi/10.1002/rob.21571/abstract}
\BIBentrySTDinterwordspacing

\bibitem{Khatib1987}
O.~Khatib, ``A unified approach for motion and force control of robot
  manipulators: The operational space formulation,'' vol. RA-3, no.~1, pp.
  43--53, February 1987.

\end{thebibliography}

\end{document}